\documentclass[11pt]{article}
\usepackage[a4paper, top = 1.25in, bottom = 1.25in, left = 1.25in, right = 1.25in]{geometry}

\usepackage[numbers]{natbib}
\usepackage[hidelinks]{hyperref}
\usepackage[utf8]{inputenc}
\usepackage{graphicx}
\usepackage{float}
\usepackage{subcaption}
\usepackage{amsfonts}
\usepackage{amsmath}
\usepackage{amsthm}
\usepackage{amssymb}
\usepackage{enumerate}
\usepackage{enumitem}
\usepackage{xcolor}
\usepackage{diagbox}
\usepackage{soul}
\usepackage{multirow}
\usepackage{verbatim}
\usepackage{xurl}
\usepackage{longtable}
\usepackage{ulem}
\usepackage{titlesec}
\usepackage{lmodern}
\usepackage[nameinlink,noabbrev]{cleveref}

\Crefname{figure}{Fig.}{Figs.}

\setlist[itemize]{itemsep=0.5em,topsep=0.5em}

\titleformat{\subsubsection}[hang]{\normalsize\bfseries\itshape}{}{0em}{}

\begin{document}


\begin{center}
\LARGE{Empirical analysis of binding precedent efficiency in Brazilian Supreme Court via case classification}
\vspace{.5cm}

\Large{
Raphaël Tinarrage$^{1,2}$,
Henrique Ennes$^{1,3}$,
Lucas Resck$^{1}$, \\
Lucas T. Gomes$^{4}$,
Jean R. Ponciano$^{5}$,
Jorge Poco$^{1}$
}
\vspace{.5cm}

\begin{minipage}[t]{0.8\linewidth}
\begin{flushleft}
\large{
$^{1}$EMAp, Fundação Getulio Vargas, Rio de Janeiro, Brazil\\
$^{2}$IST Austria, Klosterneuburg, Austria\\
$^{3}$INRIA Université Côte d'Azur, Valbonne, France\\
$^{4}$Direito Rio, Fundação Getulio Vargas, Rio de Janeiro, Brazil\\
$^{5}$ICMC, University of São Paulo, São Carlos, Brazil\\
}
\end{flushleft}
\end{minipage}
\vspace{.5cm}

\large{\today}\normalsize
\end{center}

\paragraph{Abstract}
Binding precedents (\textit{súmulas vinculantes}) constitute a juridical instrument unique to the Brazilian legal system and whose objectives include the protection of the Federal Supreme Court against repetitive demands. Studies of the effectiveness of these instruments in decreasing the Court's exposure to similar cases, however, indicate that they tend to fail in such a direction, with some of the binding precedents seemingly creating new demands. We empirically assess the legal impact of five binding precedents, 11, 14, 17, 26, and 37, at the highest Court level through their effects on the legal subjects they address. This analysis is only possible through the comparison of the Court's ruling about the precedents' themes before they are created, which means that these decisions should be detected through techniques of Similar Case Retrieval, which we tackle from the angle of Case Classification. 
The contributions of this article are therefore twofold: on the mathematical side, we compare the use of different methods of Natural Language Processing --- TF-IDF, LSTM, Longformer, and regex --- for Case Classification, whereas on the legal side, we contrast the inefficiency of these binding precedents with a set of hypotheses that may justify their repeated usage. 
We observe that the TF-IDF models performed slightly better than LSTM and Longformer when compared through common metrics; however, the deep learning models were able to detect certain important legal events that TF-IDF missed.
On the legal side, we argue that the reasons for binding precedents to fail in responding to repetitive demand are heterogeneous and case-dependent, making it impossible to single out a specific cause.
We identify five main hypotheses, which are found in different combinations in each of the precedents studied.



\paragraph{Acknowledgments}
This work is the product of a collaboration between mathematicians and Law specialists. As far as the latter field is concerned, in addition to the fourth author, we have received invaluable contributions from Beatriz Sabdin Chagas, Carla Marcondes Damian, Ana Clara Macedo Jaccoud, and Pedro Burlini de Oliveira.
As part of an introduction to research for undergraduate students, we have received support in reading documents from Livia Cales, Victoria Cury, Samanta Duarte Clara Lopes, Eduardo Portol, João Meirelles, Ana Rosenburg, and Helena Torres.
We are also particularly grateful to José Luiz Nunes for his help in accessing and navigating the dataset, as well as his sound methodological recommendations.
Finally, we are indebted to Dr.~Rafael Ramia Muneratti (Public Defender’s Office of the State of São Paulo) and Dr.~Wallace Paiva Martins Junior (Public Prosecutor's Office of the State of São Paulo), who kindly agreed to be interviewed about Binding Precedent 26.
The researcher Jorge Poco is partially financed by the National Council for Scientific and Technological Development (CNPq) under Grants \#311144/2022-5, the Carlos Chagas Filho Foundation for Research Support of Rio de Janeiro State (FAPERJ) under Grant \#E-26/201.424/2021, and the Fundação Getulio Vargas.
The researcher Jean R. Ponciano is partially financed by the University of São Paulo (PRPI Ordinance No. 1032, Process 22.1.09345.01.2).
Lucas Resck acknowledges financial support from Fundação Getulio Vargas and partial support from the Coordination for the Improvement of Higher Education Personnel (CAPES).

\section{Introduction}\label{sec:introduction}

In recent years, the progress in Natural Language Processing (NLP) sparked significant interest in its application within the legal domain. 
Brazil, renowned for having the world's highest volume of legal cases, is no exception and has already witnessed several implementations of Machine Learning methods in its judiciary system \citep{falcao2012relatorio,supremo_numeros,pereira2020viii,10.1145/3342558.3345416,CORREIA2022102794,fernandes2020appellate,bertalan2020predicting,resck2022legalvis,Nunes2022,bertalan2022using,salomao2022artificial}.
Notably, these algorithms have the potential to assist legal professionals in finding cases similar to a given one,  or a given legal topic --- a task commonly referred to as \textit{Similar Case Retrieval}. 
While these methods are supported by numerous mathematical studies and proofs of concept, one could argue that some of them lack empirical validation.
This article seeks to bridge this gap by comparing various document retrieval methods and evaluating their results from a legal perspective.

In particular, we explore the legal instrument of binding precedent (\textit{súmula vinculante}, abbreviated BP), that emerged in Brazil in the 2004 judicial reform.
Its purpose was to address the issue of an overwhelming number of cases with repetitive demands inundating the Brazilian Federal Supreme Court (\textit{Supremo Tribunal Federal}, STF), which ideally should handle only a limited number of cases. 
This situation was partly due to the Brazilian legal system's civil law approach, where Supreme Court decisions do not serve as authority for lower courts, leading to continued case congestion.
To mitigate this, Constitutional Amendment 45 introduced instruments inspired by the common law system, including binding precedents. 
They aimed at standardizing jurisprudence, providing normative force over lower instances and the broader public administration.

\begin{figure}[!ht]\centering
\footnotesize{Number of documents citing a binding precedent}
\includegraphics[width=.99\linewidth]{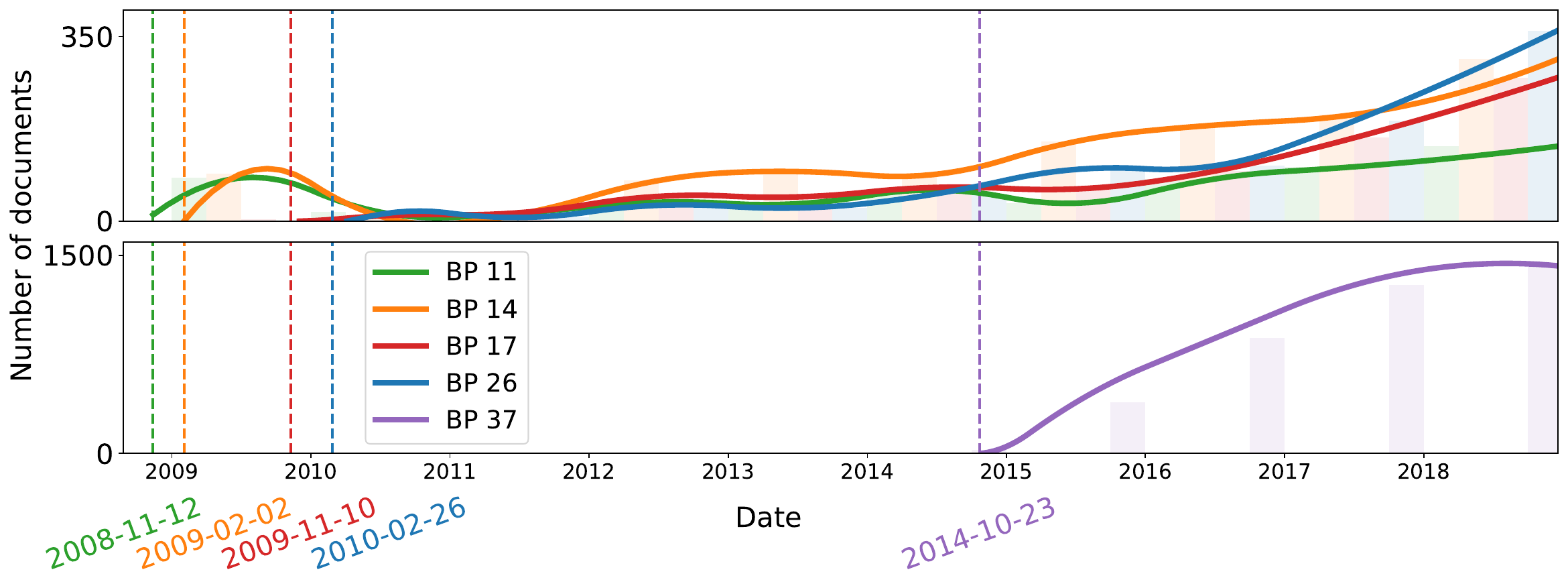}
\caption{Histograms of the number of cases judged by the Federal Supreme Court citing Binding Precedents 11, 14, 17, 26, or 37, in our collection (Dataset~\#1). The bins have a length of one year, and the curves are obtained via quadratic spline interpolation. 
The dashed vertical lines represent the date of publication of each BP. 
We draw the reader's attention to the fact that they all exhibit an increasing trend.
}
\label{fig:CitationCurve_AfterPublication_allSVs}
\end{figure}

After its publication, one expects the binding precedent to be cited frequently, until the legal understanding of the courts on the subject settles down, resulting in a significant decrease in the number of citations.
However, among the most cited binding precedents, this trend is by no means observed. On the contrary, they show steady growth, as illustrated in \Cref{fig:CitationCurve_AfterPublication_allSVs} for the five precedents of interest in this paper: 11, 14, 17, 26, and 37.
They were chosen for their high number of citations, as well as the variety of legal topics they cover, from administrative to criminal law.

This article aims to shed light on the reasons why these precedents have not led, as expected, to a reduction in repeated demands.
To this end, we employ Similar Case Retrieval and Case Classification methods to trace the history of these precedents and quantify certain trends.
This information is subsequently used to provide a legal analysis.
In summary, this article includes two main mathematical and two juridical contributions:
\begin{itemize}
\item[1.] The application and comparative analysis of classical algorithms for Case Classification on a database of Brazilian legal documents (including TF-IDF-based models, LSTM, Longformer, and regex);
\item[2.] The outline of a methodology for assessing the impact of a law on jurisprudence, through time series of similar cases and features' correlations;
\item[3.] Application of the mentioned methodology to five binding precedents emitted by the Brazilian Federal Supreme Court, enabling an empirical study of the juridical mechanisms behind their inefficiency; 
\item[4.] The identification of five main hypotheses explaining the large number of cases reaching the Supreme Court.
\end{itemize}

\subsubsection{Dataset and reproducibility}
For our analysis, we used a set of decisions of the Supreme Court produced between 1989 and 2018, collected and annotated by the project \textit{Supremo em Números} \cite{falcao2012relatorio}.
In fact, these documents were scraped from the official STF website, where they can be found one by one.
The collection of documents gathered by the project is, however, not publicly available, and has been kindly provided to us.
More precisely, we will consider two subsets of this collection, detailed further in \Cref{subsection:dataset}: Dataset~\#1, consisting of all decisions citing a binding precedent (29,743 documents), and Dataset~\#2, gathering all decisions belonging to the topics ``Administrative Law'', ``Criminal Law'' or ``Criminal Procedure Law'' (634,068 documents).
We point out that the data gathered by \textit{Supremo em Números} have already been used in several works \cite{10.1145/3342558.3345416,CORREIA2022102794,resck2022legalvis}.
In addition, other similar databases have been reported, such as the one maintained by the project \textit{Victor} \cite{de2020victor}.

\subsubsection{Overview}
The remainder of this article is organized as follows.
In \Cref{sec:related_works}, we set out the legal context behind the Brazilian Supreme Court's binding precedents, with a particular focus on the five precedents specifically studied in this article, and we give an overview of the mathematical literature surrounding Similar Case Retrieval.
Datasets \#1 and \#2 are introduced in \Cref{sec:methods}, as well as the models used throughout this article (TF-IDF, LSTM, and Longformer), that we train and test on the former dataset.
In \Cref{sec:generalization}, we apply these models to the latter dataset and evaluate the quality of their predictions.
The legal analysis of our results is the focus of \Cref{sec:legal_analysis}, where we first describe our methodology (in \Cref{subsec:methodology}, summarized in \Cref{fig:overview} below), apply it to the five precedents under consideration (\Cref{subsec:BP_11,subsec:BP_14,subsec:BP_17,subsec:BP_26,subsec:BP_37}), and gather our findings in a juridical discussion (\Cref{subsec:legal_conclusions}).
We conclude and address future works in \Cref{sec:conclusion}.

\begin{figure}[!ht]\centering
\includegraphics[width=.99\linewidth]{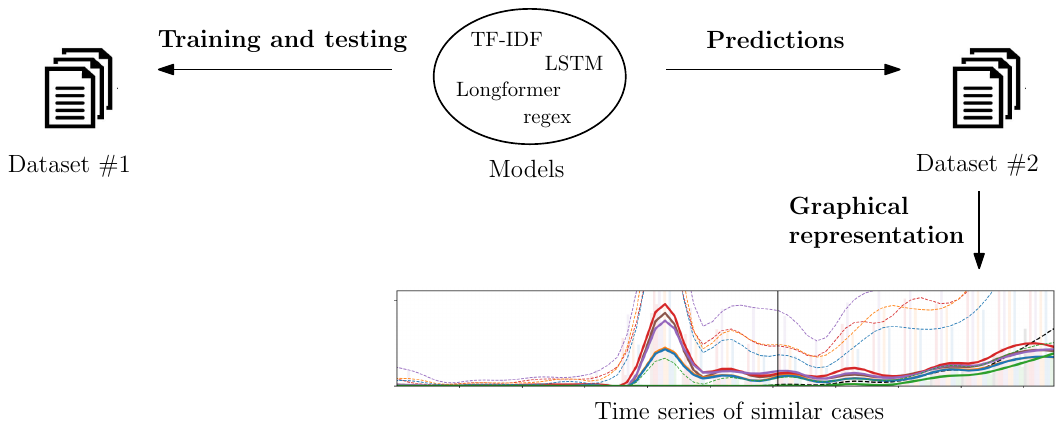}
\vspace{.1cm}
\caption{Schematic overview of the article. 
To understand the dynamics behind the use of a precedent, we train the models on an initial set of labeled documents, then apply these models to a larger set of data, and represent the results as a time series.}
\label{fig:overview}
\end{figure}

\subsubsection{Abbreviations}
Throughout this article, expressions in Portuguese will be written in italics and may be followed by a translation in English in parenthesis. 
In addition, the following list contains the abbreviations that will be used, and also introduced throughout the text.

\begin{itemize}
\itemsep0.1cm
\item[ADI] \hspace{15px}\textit{Ação Direta de Inconstitucionalidade} (Direct Action for Unconstitutionality)
\item[ARE] \hspace{15px}\textit{Recurso Extraordinário com Agravo} (Aggravated Extraordinary Appeal)
\item[BP] \hspace{15px}\textit{Súmula Vinculante} (Binding Precedent)
\item[HC] \hspace{15px}\textit{Habeas Corpus} (Habeas Corpus)
\item[Inq] \hspace{15px}\textit{Inquérito} (Inquiry/Investigation)
\item[LIME] \hspace{15px}Local
Interpretable Model-agnostic Explanations
\item[LSTM] \hspace{15px}Long Short Term Memory
\item[NLP] \hspace{15px}Natural Language Processing
\item[Rcl] \hspace{15px}\textit{Reclamação} (Complaint/Appeal)
\item[RE] \hspace{15px}\textit{Recurso Extraordinário} (Extraordinary appeal)
\item[regex] \hspace{15px}Regular Expression
\item[STF] \hspace{15px}\textit{Supremo Tribunal Federal} (Brazilian Federal Supreme Court)
\item[STJ] \hspace{15px}\textit{Superior Tribunal de Justiça} (Brazilian Superior Court of Justice)
\item[SVM] \hspace{15px}Support
Vector Machine
\item[TF-IDF] \hspace{15px}Term Frequency-Inverse Document Frequency
\end{itemize}

\section{Related works}\label{sec:related_works}

\subsection{Legal context}\label{subsec:legal_literature}
\subsubsection{The BPs as a juridical tool}
The type of precedent known in Brazil as \textit{súmula vinculante} (binding precedent) emerged in the judiciary reform of 2004, through Constitutional Amendment 45 (EC 45/2004, Article 103-A\footnote{Article 103-A \url{https://www.planalto.gov.br/ccivil_03/constituicao/Constituicao.htm\#art103a}}), to standardize decisions, based on the importation of the jurisdiction logic usual in common law \citep[p.~827]{book1702fbed}. The introduction of these instruments can be understood as an attempt to unify decision-making, aiming to achieve equality and legal certainty, i.e., to avoid identical cases being decided in different ways, violating the constitutional guarantee of equality before the law \citep[p.~20]{camara2022manual}. 
In fact, the new Federal Constitution article requires, for the creation of a BP, the fulfillment of three conditions: (i) dealing with a matter under current controversy among judicial bodies or between them and the public administration; (ii) representing a risk of serious legal uncertainty; and (iii) being the subject of significant multiplication of processes on an identical legal issue. 
Thus, as a legal instrument, the BP seeks not only to guarantee equality in judicial decisions but also to increase the efficiency of the Judiciary, by avoiding delays in jurisdictional intervention.

To date, 58 BPs have been published\footnote{List of BPs \url{https://portal.stf.jus.br/textos/verTexto.asp?servico=jurisprudenciaSumulaVinculante}}.
It should be noted that, in the literature, there is considerable criticism of the BPs, whether in terms of their institution, their formulation, or their consequences.
These issues will be explored in detail throughout the article. We refer the reader to the work of \cite{amaral2016habeas} and \cite{pereira2020viii} for a general analysis of BPs.

\subsubsection{The precedents 11, 14, 17, 26, and 37}
Once a BP is published, it is expected that the subject ceases to be a matter of controversial interpretation.
Consequently, the issue should also cease to generate processes with identical demands and, primarily, prevent these from continuing to be brought before the Higher Courts. 
It is therefore important to establish, empirically, the degree of efficiency of this type of instrument, understood here as the degree of reduction in similar cases reaching the final stage of appeal. 
In this article, we analyze the impact of the creation of five binding precedents, chosen as those generating many cases, but also for the diversity of legal topics they cover.
Three of them belong to criminal law (11, 14, and 26), and the others to administrative law (17 and 37).
We give in the list below a brief summary of their content and refer the reader to the corresponding section for a deeper explanation, as well as the results of our analysis.
\begin{itemize}[leftmargin=3mm]
\item[BP~11,] 2008-11-12 (see \Cref{subsec:BP_11}): Determines the use of handcuffs acceptable only when some risk is anticipated, predicting disciplinary punishments to the responsible public agents and/or nullity of the penal process in the case of unjustified use.
\item[BP~14,] 2009-02-02 (see \Cref{subsec:BP_14}): Grants to the investigated individual and their attorneys full access to all documented evidence in ongoing criminal investigations.
\item[BP~17,] 2009-11-10 (see \Cref{subsec:BP_17}): In Brazil, public administration at all levels, when sentenced to pay debts resulting from juridical decisions, incorporate these amounts into the public budget by the system of \textit{precatórios} (court order payments), in which orders of payment are issued to beneficiaries. The BP determines that no late payment interests should be applied to the potential delay between the issuing of orders of payment and the actual payment taking place. 
\item[BP~26,] 2010-02-26 (see \Cref{subsec:BP_26}): Determines the unconstitutionality of prohibiting convicts of ``heinous crimes" (such as murder, rape, and drug dealing) to receive a reduction of the sentence from a fully closed to semi-open or open carceral system. Moreover, the BP authorizes judges to employ ``criminological examinations'' to determine the feasibility of this reduction of the sentence.
\item[BP~37,] 2014-10-23 (see \Cref{subsec:BP_37}): 
Establishes that the Judiciary cannot increase salaries of public servants based on the principle of isonomy (equality) without prior legislation to authorize such adjustments.
\end{itemize}

\subsection{NLP for legal documents}\label{subsec:nlp_literature}

\subsubsection{Text embeddings}
The literature surrounding NLP for legal document analysis is rich, reflecting the growing interest in leveraging computational techniques to navigate the complexities of legal texts.
Well-studied topics in this area include the analysis of the precedent network \citep{Leibon2018,10.1145/3342558.3345416,Nunes2022}, Named Entity Recognition \citep{CORREIA2022102794}, summarization of legal texts \citep{hachey2006extractive,galgani2012citation}, and prediction of judicial outcome \citep{aletras2016predicting,bertalan2020predicting,bertalan2022using}.
In addition, of particular interest to us is the detection of similar documents, reviewed further in the next paragraph.

To tackle these problems, a standard technique consists of embedding (i.e., vectorizing) the documents, the most classical methods being bag-of-words, TF-IDF, and $n$-grams.
Among recent techniques of words and documents embeddings, we can cite GLoVE \citep{pennington2014glove}, contextualized word representation \citep{peters-etal-2018-deep}, word2vec \citep{church2017word2vec}, doc2vec \citep{le2014distributed}, Latent Dirichlet Allocation (LDA) \citep{blei2003latent}, entities and relations-based embedding \citep{yang2014embedding}, Universal Sentence Encoder (USE) \citep{cer2018universal}, and TextCNN \citep{chen2015convolutional}.
Another alternative is the pre-trained language models, such as BERT \citep{devlin2018bert} and its variants \citep{zhang2019ernie,peters2019knowledge,hayashi2020latent,Beltagy2020Longformer,vuong2022sm}.

In particular, word2vec is used by \cite{fernandes2020appellate} to study appellate court modifications in Brazilian legal documents, that is, modifications by the Supreme Court of the lower Court judge’s decision.
On the other hand, TF-IDF has been already used for the classification of documents citing binding precedents and has been reported to outperform other embeddings (such as doc2vec, USE, and Longformer) \cite{resck2022legalvis}.
In this article, we will consider TF-IDF embeddings coupled with several classifiers, as well as a recurrent neural network (LSTM) and a Large Language Model (Longformer).

\subsubsection{Similar Case Matching, Retrieval, and Case Classification}
At their most elementary level, although having a specific binding property, BPs are precedent and, as such, the task of searching for documents similar to the object of a BP, in content and language, may be modeled through textual proximity and/or classification techniques. 
The problem of identifying juridical decisions similar in a corpus of documents, known as automatic \textit{Similar Case Matching}, has received significant attention in the literature. For example, a multitude of NLP embedding models have been compared in the CAIL2019-SCM dataset, a corpus of thousands of decisions  published by the
Supreme People’s Court of China, in which the task is to determine, for a triple of cases $(A,B,C)$, whether $A$ is more similar to $B$ or to $C$ \citep{xiao2019cail2019scm}. Because of the problem structure, different NLP techniques have been applied to solving the task, including pre-trained models \citep{xiao2019cail2019scm}, context-based multi-learning \citep{10.1007/978-3-031-44696-2_43}, ideas from optimal transport \citep{Yu_2022}, causal inference \citep{10.1145/3539618.3591709}, and even techniques combining regex and neural networks \citep{9207528}. Further techniques for Similar Case Matching were developed and applied to datasets other than CAIL \citep{kumar2011similarity, mignone2021augmented}, with the work of \cite{BHATTACHARYA2022103069} especially significant for mixing both NLP and citation-based approaches. 

Although related in spirit, Similar Case Matching is not exactly the task we are tackling. Instead of computing similarity indexes between pairs of documents or telling which of two documents is more similar to a third one, we need to find in a huge unlabeled corpus of decisions \textit{all} decisions that are similar in content to a particular document. This alternative problem, arguably harder, is known as \textit{Similar Case Retrieval} and unsurprisingly, has been tackled through different artificial intelligence models \citep{10.1007/978-3-030-01177-2_12, raghav2016analyzing, 1488823, turtle1995text}, including some that we will be using in this work \citep{resck2022legalvis}. Curiously, though, none of these works seemed to use these techniques to describe the temporal behavior of specific precedents as we do here, restraining themselves to tools for aiding the work of juridical agents.

In order to retrieve documents similar to the thematic content of a binding precedent, we will employ a trick that reduces the problem to that of \textit{Case Classification}.
Starting from the collection of all documents (Dataset~\#2), we will consider the subset of those citing a BP (Dataset~\#1). 
Based on this second dataset, which is labeled (each document is associated with a BP), we can simply train models for this classification task, using any of the various methods popular in the literature.
Subsequently, we will apply these models to the larger dataset, thus addressing the problem of Similar Case Retrieval. 
In short, our approach is hybrid, and we will use the term ``Case Classification'' to describe it in the remainder of this article.

To close this section, we emphasize that the reason why we claim the task of case retrieval to be harder than case matching goes beyond the natural computational problem of the sheer size of inputs in the former. 
Indeed, in this context, when going from the training set (Dataset~\#1), containing actual citations of BPs, to the complete set (Dataset~\#2), where no citations to BPs are to be found, some overfitting/underfitting of the trained models is expected. 
However, this bias cannot be directly measured, as the documents of this second dataset are not labeled. 
Notice that the difficulty is twofold: not only there will be different proportions of positive and negative classes between the two sets (the positive class representing documents citing a specific BP), but also the very textual elements from the positive and negative classes are expected to be distinct, even when regex is used to remove the citations themselves.
In conclusion, the problem of Similar Case Retrieval tackled in this article goes beyond other analyses of classification problems found in the literature, where one can, a posteriori, quantify the validity of models through accuracy, recall, or $F_1$ score.
To validate our results, two techniques will be used.
First, we will employ the labels in Dataset~\#1 (which accounts for 4.7\% of Dataset~\#2) to compute common metrics such as those cited above.
This will be performed in \Cref{sec:methods}.
Secondly, we will propose a thorough manual validation, based in particular on the reading of documents and the analysis of prediction curves (in \Cref{sec:generalization}).

\subsubsection{Network analysis}
Although there is a multitude of NLP models that solve both the similarity matching and retrieval problems, to our knowledge, none of these have yet been applied to the actual \textit{juridical} study of the creation, the evolution, and the authority of precedents. On the other hand, empirical legal literature has often investigated these questions using ideas from graph theory. That is, given the natural referential structure inherited from the use of precedents, a substantial part of the empirical study of courts is dedicated to modeling them through directed graphs, with vertices the set of all decisions, and with an arrow from a decision $x$ towards a decision $y$ if and only if $y$ cites $x$ as a precedent. Drawing from the general theory of citation networks \citep{10.1145/324133.324140}, authors have developed metrics to measure the authority of a specific decision as a precedent in a determined court through the decision's (outward) degree in the induced graph, emulating the hypothesis that the more important a precedent, the more cited it is. 

These metrics have then been applied to identify the most important precedents in different settings and the evolution of their authority through time. In a fundamental empirical work on the United States Supreme Court \citep{FOWLER200816}, for example, the authors used the (normalized) degree of the decisions to identify historical features of the most important precedents to the court, concluding that for most cases, when an important decision is taken by the court, the number of documents citing it follows a curve which tends to increase with time, reach a peak, and then decrease exponentially. Interestingly, the authors indicate that although this seems to be a general trend, internal forces within the court, such as the reversion of previous decisions, may significantly impact these curves. 
The author of \cite{Whalen2013} points out that even external sources, such as the court's political composition impact the citation curves of precedents in the Supreme Court and, in a study considering all instances in the American justice, \cite{Smith2005-sl} identifies the probability distributions generating the precedent graphs. Similar conclusions regarding the change of authority over time are found in the International Criminal Court \citep{tarissan:hal-01217997}, the Canadian legal system \citep{neale2013citation}, and the Court of Justice of the European Union \citep{7752308}. In this last study, the authors had access to a list of the most important precedents identified by the court's legal experts and attested that, in practice, there exist important cases, called symbolic, which will not have high authority scores, even considering their doctrinal impact. 

In an attempt to replicate these studies to the Dutch legal system, \cite{winkels2011determining} discusses whether the use of citation networks is appropriate to a civil law setting, in which precedents play a minor role compared to international courts or common law systems. Interestingly, the authors conclude in favor of the usefulness of network analysis also in this alternative setting, although results should be differently interpreted.

\subsubsection{Network analysis of the Brazilian system}
Another application of network analysis to a civil law system can be found in \cite{10.1145/3342558.3345416}, who applied graph techniques to study precedents of the Brazilian Federal Supreme Court. The authors argue that, given the recent reformations of the Brazilian system aimed toward a higher observance of precedents, graph-based methods become even more useful to this setting. Moreover, historical analysis of authority scores over time indicates citation curves of precedent similar to what was found by \citep{FOWLER200816}, although here, the topic of the decision --- i.e., constitutional, criminal, etc.~--- has a strong influence on the shape of the curve. Nonetheless, the authors point out that even in the more ``mixed" Brazilian system, some complications arise from the non-exclusivity of the precedent paradigm: the lack of a common law culture makes Justices' citations to other documents missing standardization, which not only complicates the whole construction of the networks but also ``contaminate" the documents with only marginal references to precedent. By marginal preferences to precedent, we mean citations to documents that do not impact the merit of the decision, only referring to some procedural aspect of previous decisions.

Moreover, this non-existence of a standardized citation system for the STF allows for what the authors call indirect citations, that is, references to documents that cannot be captured through parsing techniques based on regex. In practice, this means that the interpretation of the authority of a precedent as the number of citations to it at a certain interval of time is insufficient to capture their impact on the legal system.

Finally, the situation is aggravated by the defensive posture assumed by the court to cope with its excessively high demands, with many of the most cited precedents being only procedural decisions of the STF denying to arbitrate on the merit of cases and directing them to other courts. This distinction between decisions where the precedent use has a significant juridical value in the judge's juridical reasoning, which we shall call \textit{citations by merit}, versus those in which the citation is only a procedural justification for dictating the next formal steps of the case, which we call \textit{procedural citations}, impose a noteworthy constraint in the interpretation of the number of citations of a document as metric for its authority, for a particularly juridically insignificant decision may be frequently cited due to some incidental procedural extract. These constraints are specially challenging to deal with in the study of the STF: although the vast majority of court decisions cite precedent, most of them lie in the procedural category \citep{supremo_numeros}, limiting the ordinary interpretation of authority scores.

One is then led to conclude that the infant precedent citation culture in Brazil and the predominance of procedural-only citations discourage the use of graph techniques in the study of BPs. Additionally, the very use of these precedents as a tool to \textit{decrease} the frequency of repetitive cases arriving at the STF indicates that degree-based scores are insufficient to describe the effectiveness of the BPs as defense devices, suggesting the use of embedding-based methods.

\section{Training and testing on Dataset~\#1}\label{sec:methods}

\subsection{Datasets}\label{subsection:dataset}

\subsubsection{Documents}
Cases decided at the STF are consolidated in a document, containing the decision's text, as well as various metadata.
The research project \textit{Supremo em Números} gathered, in text format, more than 2,500,000 of these documents, published between 1989 and 2018 \citep{falcao2012relatorio}.
They were obtained by scraping the official STF website\footnote{STF search engine \url{https://portal.stf.jus.br/}}, where all the documents are publicly accessible.

It is worth noting that the documents follow a typical structure:
\begin{enumerate}[style=multiline,leftmargin=1.9cm,labelwidth=3.5cm,itemsep=0.5em,topsep=0.5em]
\item[\textit{Cabeçalho} (Header)] 
Contains a number of metadata, such as the case identification number, the involved parties, the relator (Justice in charge), the origin of the case (if appealed from a lower court), and its type (\textit{classe processual}, e.g., Rcl).

\item[\textit{Ementa} (Abstract)]
A summary of the key points of the decision, highlighting the legal thesis, the constitutional provision at issue, and the most relevant reasoning.

\item[\textit{Relatório} (Report)]
Description of the procedural history of the case. It summarizes the main facts, arguments, lower court decisions (if any), and other relevant filings.

\item[\textit{Fundamentação} (Reasoning)]
Detailed legal and argumentative analysis.
It presents the interpretation of the relevant legal and constitutional norms, and how precedents apply.

\item[\textit{Dispositivo} (Ruling)]
The final part of the decision, stating the outcome in clear terms.
In Plenary or Panel decisions, the ruling reflects the voting result.
\end{enumerate}
Although one could analyze documents by distinguishing the different parts that compose them, in this article, we have chosen to treat them as a whole. In particular, and as described in \Cref{subsec:models_employed}, two embedding methods will be considered: bag-of-words models (with TF-IDF) and more advanced NLP methods (LSTM and Longformer).

In addition, each document of the collection is associated with a category (\textit{ramo do direito}, ``branch of law''), such as ``Administrative Law'' or ``Criminal Law''.
Although there is no official list published by the STF, these categories align with the well-known branches of Brazilian law.
They are used internally, on the official website, to classify laws and documents\footnote{One can search for a binding precedent based on its branch at \url{https://portal.stf.jus.br/jurisprudencia/aplicacaosumula.asp}, or, similarly, for a document at \url{https://jurisprudencia.stf.jus.br/pages/search}}.

We draw the reader's attention to the fact that the metadata presented so far, associated with each document --- e.g., the relator, document type, or its branch --- is non-ambiguous and does not depend on annotation by an expert.
However, we will need to extract other types of information from the documents. 
This includes, for instance, the state of provenance of the document, the outcome of the decision, the use of a specific law, or more generally the presence of certain words. This will be achieved by a simple regular expression search in the documents.
Furthermore, in order to verify the content of certain documents (whether to check that the models have correctly detected a document or to better understand its content), we will have to read them. In this case, the information is potentially ambiguous and we will make this explicit each time.

\subsubsection{Construction of our datasets}
From the collection of documents gathered by \textit{Supremo em Números}, we have devised two datasets that will allow us to perform our analysis of binding precedent efficiency.
For the first, we have collected, through regular expression searches, all documents that cite a binding precedent, among the 58 precedents edited by the STF until 2018.
Then, we have chosen to consider only the ten most cited precedents and to discard the others. 
The final dataset amounts to 29,743 documents, citing a precedent among BPs 3, 4, 10, 11, 14, 17, 20, 26, 33, and 37.
We will refer to it as Dataset~\#1.

We remind the reader that this article focuses specifically on the precedents 11, 14, 17, 26, and 37.
As already presented in \Cref{subsec:legal_literature}, they were chosen for the diversity of the subjects they cover, and the high number of cases they produce.
The number of documents corresponding to each of these BPs is given in \Cref{tab:datasets_1_and_2}.
We draw the reader's attention to the fact that some documents in our collection cite several BPs.
More precisely, out of the 29,743 documents, 1116 cite exactly two BPs, and 93 cite three or more.
As a consequence, in the table for Dataset~\#1, the total number of documents is not equal to the sum of the values in the rows above.

In order to take a more comprehensive look at the issues covered by these precedents, we shall also consider the so-called Dataset~\#2, consisting of those documents, in the whole collection, whose branch is ``Administrative Law'', ``Criminal Law'' or ``Criminal Procedure Law''.
These are the categories of the five specific BPs under study.
In particular, Dataset~\#1 is a subset of Dataset~\#2.
The number of documents contained in Dataset~\#2 is 634,068, and we display the number per branch in \Cref{tab:datasets_1_and_2}.

\begin{figure}[!ht]\centering
\footnotesize{Time series of the publication date of documents in Dataset~\#1}
\includegraphics[width=.99\linewidth]{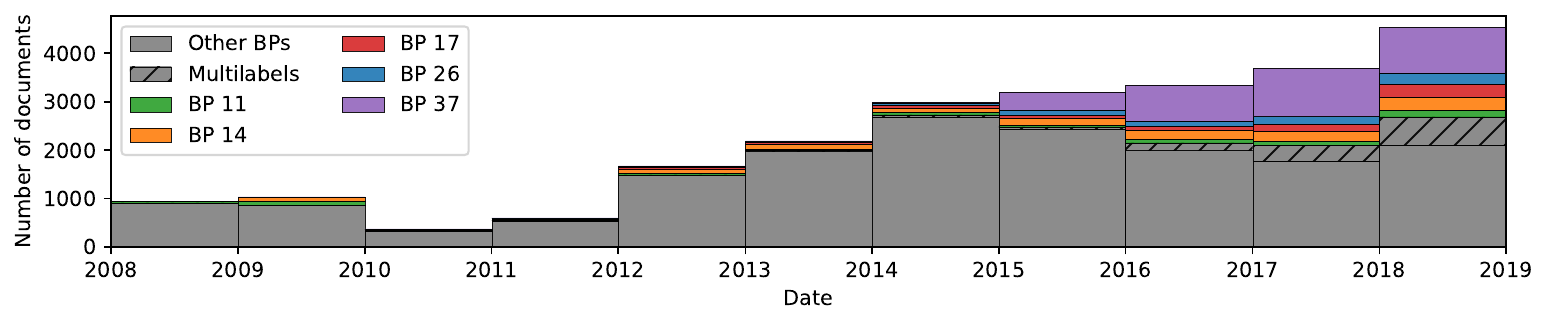}
\footnotesize{Time series of the publication date of documents in Dataset~\#2}
\includegraphics[width=.99\linewidth]{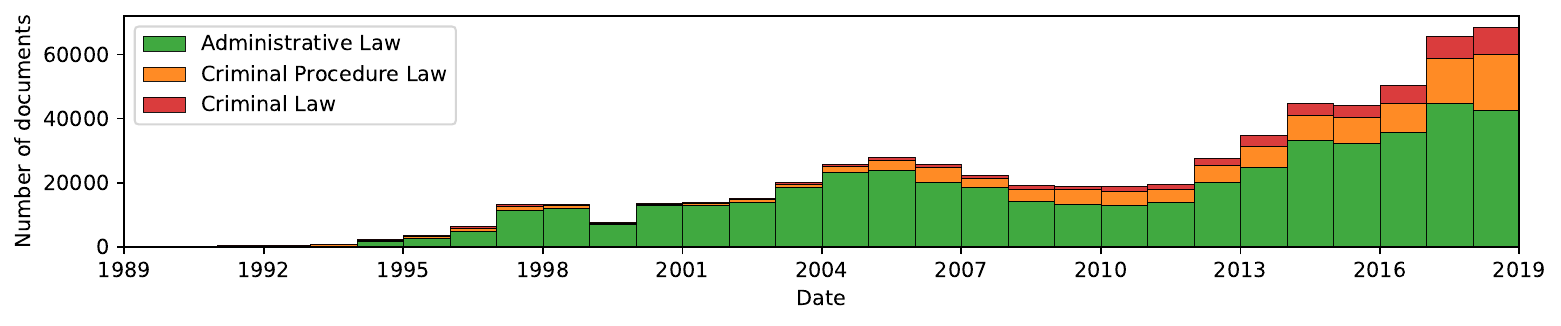}
\caption{Histograms of the number of cases in Dataset~\#1 (top) and Dataset~\#2 (bottom). The bins have a length of one year.}
\label{fig:TimeSeries_Dataset1_and_Dataset2}
\end{figure}

To provide a last insight into these datasets, \Cref{fig:TimeSeries_Dataset1_and_Dataset2} shows the time series of document publication dates. We can see that, in both cases, the trend is increasing, reflecting the congestion of the Supreme Court over time.

\begin{table}[!ht]
    \begin{subtable}[t]{.3\linewidth}
    \centering
    \begin{tabular}[t]{r|rr}
    Dataset~\#1 & \begin{tabular}{@{}c@{}}Number of \\ documents\end{tabular} & \begin{tabular}{@{}c@{}}Of which cite \\ another BP\end{tabular} \\\hline
    BP~11 & 830 & 42\\
    BP~14 & 1601 & 62\\
    BP~17 & 856 & 16\\
    BP~26 & 954 & 188\\
    BP~37 & 3984 & 895\\
    Other BPs & 22,650 & 1138\\
    Total & 29,743 & 1209
    \end{tabular}
    \end{subtable}%
    ~~~~~~~~~~~~~~~~~~~~~~~~~~~~
    \begin{subtable}[t]{.3\linewidth}
    \begin{tabular}[t]{r|r}
    Dataset~\#2 & \begin{tabular}{@{}c@{}}Number of \\ documents\end{tabular} \\\hline
    Administrative Law & 480,903 \\
    Criminal Procedure Law & 106,537 \\
    Criminal Law & 46,628 \\
    Total & 634,068
    \end{tabular}
    \end{subtable}%
\caption{Composition of the datasets involved in our study.}
\label{tab:datasets_1_and_2}
\end{table}

\subsubsection{Case Classification and labels}
In this article, we seek to analyze the influence of binding precedents on Brazilian case law. 
To do so, our methodology consists of detecting all documents that relate to the theme of the studied BP and analyzing them. These documents may be those that cite the BP, but also those that could use it without citing it explicitly (potential citations), or those that were published before the BP, but could have cited it if the BP already existed.
In other words, we are dealing with a \textit{Similar Document Retrieval} problem, where the cases are those that fall within the scope of the BP.
We therefore need to train models to detect potential citations of the precedent.

In this context, Dataset~\#1 is particularly convenient.
Indeed, since the various binding precedents cover different themes, one decides whether a document deals with a specific theme by reading the corresponding BP(s) it cites.
In opposition, in Dataset~\#2, apart from the branch to which the document belongs, no information is available regarding its juridical content.
For this reason, we will employ the former to train and test our models.
This way, the problem is reduced to that of \textit{Case Classification}, in the form of a mere binary classification task.
More specifically, since we want to train the models not to detect \textit{citations} but \textit{potential applications} of the BP, we will mask explicit mentions of the precedent in the training (and testing) data. 
This approach simulates a situation where the BP does not yet exist.
In the rest of this section, only Dataset~\#1 will be considered.

However, restricting our study to Dataset~\#1 will not solve our problem, since we would then only obtain documents that \textit{explicitly cite} some BP. To carry out our program, we, therefore, apply these models to Dataset~\#2, which contains the document categories in which we expect to detect potential cases of application of the precedents.
The application of our models to Dataset~\#2 will be studied from a Machine Learning perspective in \Cref{sec:generalization} and a juridical perspective in \Cref{sec:legal_analysis}. More precisely, in the former section, we will examine the generalization problem we face when passing from Dataset~\#1 to Dataset~\#2, and in the latter, we will use our findings to draw juridical conclusions regarding the BPs considered.

\subsubsection{Length of the documents}
When it comes to implementing automatic document analysis, an important quantity is the length of the documents.
While some methods accept documents with an arbitrary length --- for instance, bag-of-words vectorizations ---, others accept only a limited input.
As an example, the popular transformer-based model BERT treats, in its native form, documents of 512 tokens at most \citep{devlin2018bert}.
As we will describe more thoroughly in \Cref{subsec:models_employed}, we have decided to consider three distinct models: TF-IDF, LSTM, and Longformer.
The first embedding accepts documents of any length, but the last two have a limitation, which must be chosen a priori. 
We have chosen, for both models, a maximal length of 4096 words or tokens.

To get an idea of the impact of this limitation, \Cref{subfig:length_of_documents_raw} shows the distribution of document lengths in Dataset~\#1.
In particular, the median length is 1019 words, and very few documents are longer than 4096 words (only $3\%$).
That said, the models considered expect a specific preprocessing of the data.
For TF-IDF and LSTM, this encompasses the lemmatization of the texts and the removal of Portuguese-specific stop words.
For Longformer, a specific tokenizer is used.
The distribution of the length of the documents, after the application of these preprocessings, is shown in \Cref{subfig:length_of_documents_lemmatization,subfig:length_of_documents_tokenization}, respectively.
Clearly, after lemmatization and stop words removal, the length of the documents is reduced. 
In particular, the choice of a maximum of 4096 words for LSTM means that almost all documents can be processed in their entirety (only 0.8\% exceed the limit).
On the other hand, after Longformer's tokenization, the documents increase in length. Indeed, long or rare words are typically broken into 2 or 3 tokens.
Fortunately, and as visualized in \Cref{subfig:length_of_documents_tokenization}, only $6\%$ of the tokenized documents have a length greater than 4096.
We conclude that our choice of maximum length should not significantly impair the analysis.

\begin{figure}[H]\centering
\begin{subfigure}{1\linewidth}\centering
\footnotesize{Distribution of the number of words in raw documents (no preprocessing)}
\includegraphics[width=.99\linewidth]{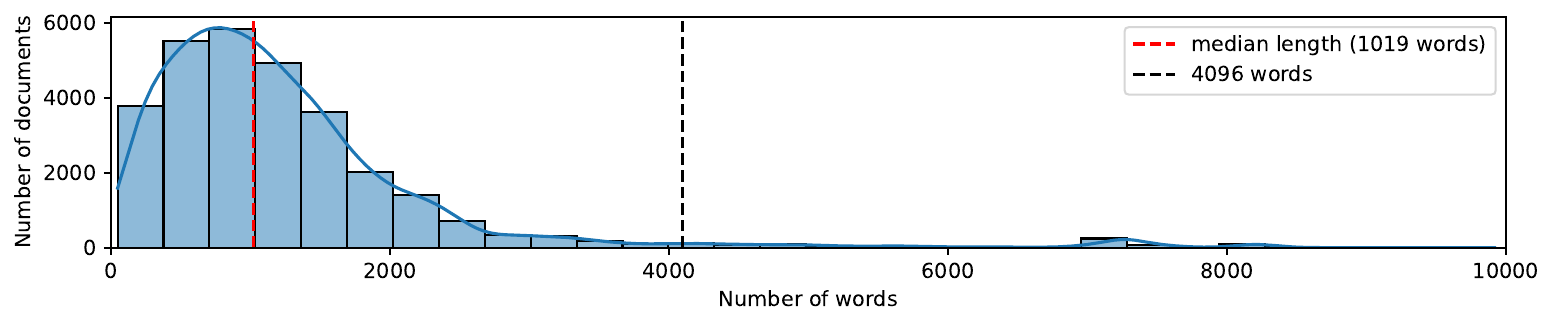}
\vspace{-.25cm}
\caption{Raw documents}
\label{subfig:length_of_documents_raw}
\end{subfigure}
\end{figure}

\begin{figure}[H]\ContinuedFloat\centering
\vspace{-1.25cm}
\begin{subfigure}{1\linewidth}\centering
\footnotesize{Distribution of the number of words after lemmatization and stop words removal}
\includegraphics[width=.99\linewidth]{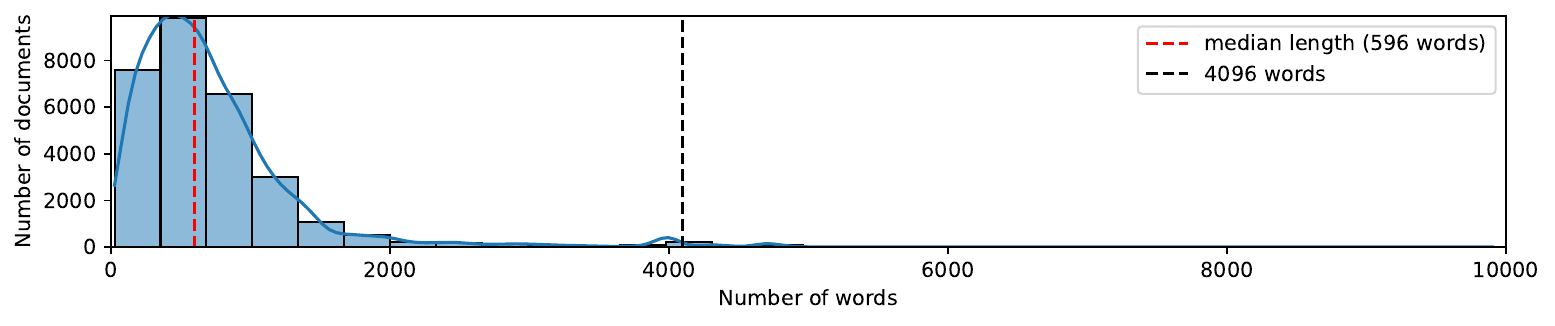}
\vspace{-.25cm}
\caption{Preprocessing for TF-IDF and LSTM}
\label{subfig:length_of_documents_lemmatization}
\end{subfigure}
\end{figure}

\begin{figure}[H]\ContinuedFloat\centering
\vspace{-1.25cm}
\begin{subfigure}{1\linewidth}\centering
\footnotesize{Distribution of the number of tokens after tokenization for Longformer}
\includegraphics[width=.99\linewidth]{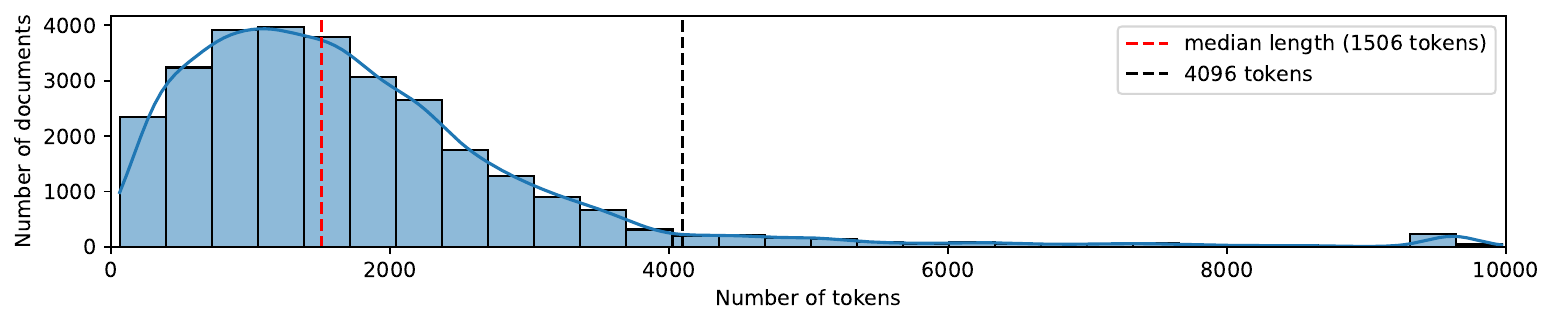}
\vspace{-.25cm}
\caption{Preprocessing for Longformer}
\label{subfig:length_of_documents_tokenization}
\end{subfigure}
\caption{Distribution of length of documents in Dataset~\#1, for different preprocessings.}
\label{fig:length_of_documents}
\end{figure}

\subsection{Models for Case Classification}\label{subsec:models_employed}

In this section, we delve into the methodologies employed for analyzing our datasets, focusing on four distinct approaches: TF-IDF, LSTM, Longformer, and regex. 
Together, they faithfully represent some of the most common NLP methods.

\subsubsection{TF-IDF} Term Frequency-Inverse Document Frequency (TF-IDF) is a vector embedding technique divided into two parts. In the ``term-frequency'' part, each document $d$ is embedded in the space $\mathbb{R}^{n}$, where $n$ is the number of different words on the corpora, and where the $i$-th component of the vector is the frequency that the $i$-th word of the corpora appears in the document. We denote it by $\text{TF}(i,d)$. The ``inverse document frequency" part is another vector of $\mathbb{R}^n$, with $i$-th coordinate given by the logarithm
\begin{equation*}
    \text{IDF}(i,d)=\log \frac{\#\text{ documents in the corpora}}{\#\text{ documents in the corpora that } i \text{ appears}}
\end{equation*}
Finally, the TF-IDF vector has the $i$-th coordinate given by the product
\begin{equation*}
    \text{TF-IDF}(i,d) = \text{TF}(i,d)\times \text{IDF}(i,d).
\end{equation*}

By itself, TF-IDF is only an embedding, so it must be coupled with some classification algorithm for us to perform Case Classification. In this article, we will use three kinds of classification models for the embedded documents: linear Support Vector Machine (SVM), logistic regression, and random forests. We will not explicitly describe these classifiers here, but refer the interested reader to \cite{geron2022hands}.

In practice, we trained the models on a binary task on Dataset~\#1 (described in \Cref{subsec:validation}), where 90\% of the data was used for training and 10\% for testing.
Based on a validation set, the models were trained to maximize the $F_1$ score, from grid search on specific ranges of parameters ($\{1,5,10,50\}$ for SVM, $\{10,10^2,10^3,10^4\}$ for logistic, and $\{10,10^2,10^3\}$ trees in the forest), through cross-validation with 5 folds.

\subsubsection{LSTM}
A recurrent network architecture was also implemented for this task, given the well-known capacity of these models to work with textual data, especially for longer texts \citep{geron2022hands, sak2014long}. 
The implemented neural networks were composed of three layers: one embedding layer, followed by two layers consisting of 128 LSTM (Long Short-Term Memory) neurons each, all using \texttt{relu} activation functions. For the embedding layer, \texttt{tensorflow} was used, where the vocabulary (i.e., set of unique words in the corpora) was restricted to the 10,000 most common terms, using 1,000 out-of-bucket embedding for the rest of them. Moreover, because the architecture expects a constant input size of documents, only the first 4096 tokens of each document were considered, using additional masking for shorter documents. Although naive, this approach is common when dealing with long text analysis in neural networks. 
Finally, the output consisted of a single neuron followed by a sigmoid non-linearity. We stress that no pre-training of the network's weights
was used.

Models with this LSTM architecture were trained for each BP using binary cross-entropy for loss  (weighted in accordance with the imbalance between the positive and negative classes). As we will describe in \Cref{subsec:validation}, 10$\%$ of the data was dedicated for testing, and a further 10$\%$ for validation. We tested with three hyperparameters on the validation set: the learning rate, which we took from the set $\{5\times 10^{-5},5\times 10^{-4}, 5\times 10^{-3}\}$ and which was fixed throughout training; the batch size, which we took from the set $\{16,32,64\}$, and which was implemented via gradient accumulation to avoid overloading the GPU; and the number of epochs, in the set $\{5,20,40\}$. Our best results were obtained using the same values for learning rate and batch size ($5\times 10^{-4}$ and 32, respectively) for all BPs, but different numbers of epochs.

\subsubsection{Longformer}
It is a transformer-based architecture designed to efficiently handle long sequences by incorporating a modified self-attention mechanism that scales linearly with sequence length \citep{Beltagy2020Longformer}. 
For sequence classification tasks, Longformer uses a global attention token that attends to all parts of the sequence, while each token within the sequence only attends to a local window around itself. 
This approach effectively reduces the computational overhead compared to standard full-attention transformers  \citep{vaswani_attention_2017}, enabling the model to process longer documents without sacrificing critical contextual information. 
As with other transformer models, Longformer includes layers of multi-head self-attention, feed-forward networks, and layer normalization.

In this article, we utilized the publicly available \texttt{longformer-base-4096} model released by Allen Institute for AI (allenai)\footnote{Pretrained model available at \url{https://huggingface.co/allenai/longformer-base-4096}}. 
This model was pre-trained on a large corpus of English texts and fine-tuned in our experiments for sequence classification on Dataset~\#1. 
Even though its pre-training data is in English, and our data in Portuguese, the underlying learned representations of language structure, context, and semantic relationships are expected to transfer well to other languages.

More specifically, for each BP, the model was fine-tuned for 5 epochs (after grid search with 1, 3, and 5 epochs) through weighted binary cross-entropy loss (as LSTM), with a learning rate of $5\times 10^{-5}$ (the best in the range $\{1,5\}\times \{10^{-3}, 10^{-4}, 10^{-5}\}$), and a batch size of 64 (chosen among powers of two between 2 and 512) via gradient accumulation.
The token limit we chose is 4096, the maximum for \texttt{longformer-base-4096}.

\subsubsection{regex} 
Regular expression (regex) stands out as a commonly used tool among professionals for retrieving similar cases. 
In particular, it constitutes the search engine of the official STF website.
We designed a regex search for the topics encompassed by the five BPs by selecting, with the help of legal expertise, the most important words in their statements (refer to \Cref{subsec:BP_11,subsec:BP_14,subsec:BP_17,subsec:BP_26,subsec:BP_37} for the statement and juridical context of each BP, or to \Cref{subsec:legal_literature} for an overview). 
We draw the reader's attention to the fact that, as our models are intended to detect, in a subsequent section, the documents \textit{before} the creation of the BPs, we do not use the words \textit{Súmula Vinculante} or any specific mention of a law.
More precisely, we consider:

\begin{itemize}
    \item[BP~11:] the words \textit{algemas} or \textit{algemado} (handcuffs, handcuffed),
    \item[BP~14:] the expression \textit{acesso aos elementos/autos/documentos} (access to elements, records or documents),
    \item[BP~17:] the word \textit{precatório} and the expression \textit{juros de mora} (court orders, late payment interest),
    \item[BP~26:] the expressions \textit{exame criminológico} and \textit{progressão de regime} (criminological examination, regime progression),
    \item[BP~37:] the expressions \textit{isonomia}, \textit{vencimentos} and \textit{servidores públicos} (isonomy, salaries, public servants).
\end{itemize}

We emphasize that these regular expression searches are fairly elementary, and could potentially be improved by Named-entity recognition (NER) techniques. However, they are quick to perform, and highly explicable, allowing direct interpretation of the results. 
In addition, as we will see during the validation, they achieved surprisingly correct scores (see \Cref{tab:similar_case_matching}).

\subsubsection{Running times}

Training the TF-IDF models and applying them to Datasets~\#1 and \#2 did not require significant computational resources and were performed on a CPU.
In contrast, the LSTM and Longformer models required a GPU for both training and prediction. Among the models, Longformer was the most resource-intensive. For each BP, the training process took approximately 14 hours\footnote{On a system equipped with an Intel(R) Xeon(R) Gold 6230 CPU, 192GB of RAM, and an Nvidia(R) Dual Quadro RTX6000 GPU}, using the parameters detailed in the corresponding section above. The primary bottleneck in training stemmed from the document lengths (4096 tokens).
On the other hand, prediction on Dataset~\#2 (containing 634,068 documents), if carried out without tricks, would take around 31 hours (approximately 175 milliseconds per document), using the maximum batch size supported by the GPU, which is 16.
To reduce this time, we employed two optimization strategies.
First, we utilized the fast vectorization capability natively included with \texttt{longformer-base-4096} and enabled FP16 computational precision, which reduced the processing time by 55\% (i.e., 78 milliseconds per document). Second, we implemented dynamic batching: instead of grouping documents into fixed-size batches of 64 and padding (or truncating) them to 4096 tokens, we ordered documents by length and dynamically created the largest possible batches. 
More precisely, we ensured that, after padding to the length of the largest document in the batch, the total token count did not exceed $16\times4096$. This further reduced the processing time by 33\% (or 53 milliseconds per document).
With these optimizations, the prediction of Dataset~\#2 using a Longformer model was completed in approximately 9 hours.

\subsection{Validation}\label{subsec:validation}

Our first analysis focuses on Dataset~\#1, a collection of documents labeled by the binding precedent they cite. 
We are, therefore, dealing with a binary classification task.
As we will see in this section, and as is expected from the NLP literature, all the models perform well on this task. 
That said, we remind the reader that the main objective of this article lies in the analysis of Dataset~\#2 (in \Cref{sec:generalization,sec:legal_analysis}), which is unlabeled and presents a more challenging scenario with different outcomes.

\subsubsection{Preparation of the data}
Some preprocessing was applied to the documents before training. 
For both the TF-IDF and LSTM models, this encompassed the lemmatization of the texts, the removal of Portuguese-specific accents, and of usual stop words (e.g., \textit{a}, \textit{o}, \textit{um}, \textit{mas}). 
Because Longformer does not expect this sort of preprocessing, raw texts were given.

Additionally, we want to avoid the models from learning textual cues that mark direct citations to BPs, such as the very term \textit{Súmula Vinculante} or similar. 
Therefore, regular expressions were used to substitute these terms with empty strings. Additionally, as our interest consists in detecting possible uses of BPs before their publications, dates were removed. 
Once more, Longformer models receive a slightly different masking process than the others: each BP citation is identified using regular expressions and is replaced by \texttt{<mask>} tokens, keeping the same length as the original text.

Our models are trained on Dataset~\#1 (described in \Cref{subsection:dataset}).
We randomly split the dataset into training and test data, while preserving the distribution of the BPs among the documents. 
For such, 10$\%$ was dedicated to testing, and from the training data, a further 10$\%$ was used for validation. 
Although the pre-processing of input texts was not the same for each class of models, we ensured that all shared the same documents for training, testing, and validation.
Lastly, given a BP, we attributed to the documents the labels ``1'' or ``0'', depending on whether the document cites or not the BP, and trained the models for this binary classification task.
Ultimately, for each of our five model types (TF-IDF+SVM, TF-IDF+logistic, TF-IDF+random forest, LSTM, and Longformer), we obtain five different trained models (one per BP).

A word of caution is warranted regarding how the data are split into training and test sets.
As opposed to a multiclass dataset, where each document would be associated with one label, our dataset is \textit{multilabeled}, in the sense that certain documents can be associated with several labels. To split such a dataset, it would be ideal to reproduce the same distribution of combinations of labels in the training and the test set.
To achieve this, we use the iterative stratification algorithm\footnote{Multilabel data stratification is implemented in \texttt{skmultilearn} \url{http://scikit.ml/stratification.html}} presented in \citep{sechidis2011stratification,pmlr-v74-szymański17a}.

\subsubsection{Classification task}

From a methodological point of view, we emphasize that, in this section, our approach consists of training models for the recognition of citations to the binding precedents. 
Furthermore, the models are trained on Dataset~\#1, where explicit references to BPs have been removed; this will ensure that, when later applied to Dataset~\#2, the models will not simply be able to detect documents that cite one of the five BPs, but more generally documents that, from a legal point of view, fall within the scope of the BP.

In order to detect documents citing one of the five BPs considered, we will train five independent models, one for each BP, addressing the binary task related to the BP.
In other words, we have transformed the multiclass classification task into five binary classification tasks. However, another approach, common in such problems, is worth mentioning: training only one multiclass classifier, capable of outputting the BP detected among the five, or else a null class if no citation is detected.
That said, this latter point of view is not well suited to our problem, for two reasons. Firstly, our problem is inherently multilabel: as we saw in \Cref{tab:datasets_1_and_2}, 4\% of the documents in Dataset~\#1 cite two BPs or more. By training separate binary classifiers, we enable the possibility of potentially assigning several BPs to a single document.

Secondly, and perhaps more importantly, we do not want the detections of one BP to influence another. Indeed, although the BPs deal with distinct legal themes, the limits of their scope in practice can be blurred, sometimes overlapping with one another. 
A unified multilabel model risks narrowing their scope, which is not the case with isolated binary classifiers.
This phenomenon will be observed, for example, in the analyses of BPs 17 and 37 (in \Cref{subsec:BP_17,subsec:BP_37}). 
While one addresses late payment penalties and the other deals with salary adjustments, there are cases that intertwine these two themes, or that rely on common legal arguments (such as isonomy).

\subsubsection{Scores}

To assess the performance of our binary classifiers, we use a handful of score metrics.
By denoting P (number of class 1 examples), N (class 0), TP (class 1 predicted class 1), FP (class 0 predicted class 1), TN (class 0 predicted class 0), and FN (class 1 predicted class 0), we consider the following metrics: 
\begin{align*}
\mathrm{Accuracy}&=(\mathrm{TP}+\mathrm{TN})/(\mathrm{P}+\mathrm{N}),\\ \mathrm{Precision}&=\mathrm{TP}/(\mathrm{TP}+\mathrm{FP}),\\
\mathrm{Recall}&=\mathrm{TP}/\mathrm{P},\\
F_1&=2~\mathrm{Precision}\cdot\mathrm{Recall}/(\mathrm{Precision}+\mathrm{Recall}).
\end{align*}
Intuitively, accuracy measures the ratio of correct predictions, while precision measures the accuracy of pointing to class 1, and recall measures how much of class 1 is being correctly predicted. The $F_1$ score is the harmonic mean of recall and precision.
We point out that because our classes are highly imbalanced (with many more documents in class~0 than in class~1, as seen in \Cref{tab:datasets_1_and_2}), the accuracy is of limited value. 
Our reference is the $F_1$ score, which specifically measures how well class~1 is predicted.

Although our models ultimately produce a binary decision, some internally generate a continuous output, that can be interpreted as a probability of belonging to class 1. For instance, LSTM’s native output is a value $ p $ between 0 and 1, and one classifies a document as class 1 if $ p \geq 0.5 $.
In contrast, Longformer's raw output is a logit, represented as a pair $ (p, q) $ that can be greater than 1 or less than 0. 
One typically applies a softmax activation to this logit, producing a new pair $ (p', q') $ in the interval $ [0,1] $. 
The second value, $ q' $, is then interpreted as the probability of belonging to class 1, and one predicts the class by checking if $ q' \geq 0.5 $.
For the classifiers paired with TF-IDF, only logistic regression natively uses a probability value, for it is inherently a probabilistic model. 
Each document results in a value $ p $ and is classified as class 1 if $ p \geq 0.5 $. 
We can thus directly request this probability output from the model.
By contrast, SVM and random forest do not have a native notion of probability, though one can force them to return a continuous output. 
In the case of SVM, a natural continuous quantity is the signed distance $ d $ from the margin; a document is said to belong to class 1 if $ d \geq 0 $. 
One can transform $ d $ into a probability via Platt scaling, obtaining a new value $ d' $, and then classify according to $ d' \geq 0.5 $. 
Note that the conditions $ d \geq 0 $ (``pure'' SVM prediction) and $ d' \geq 0.5 $ (``calibrated probability'') are not equivalent: they may disagree slightly on near-boundary samples.
However, the results are generally very similar.
Last, random forest can also straightforwardly provide a probability: instead of returning the raw majority-vote class label from the ensemble of decision trees, it averages the fraction of training samples in each leaf that belongs to the positive class. 
The final decision is again made by checking if $ p \geq 0.5 $.

Based on these probability outputs, one could fine-tune the thresholds --- rather than always using 0.5 --- to potentially achieve better predictions.
The \textit{Area Under the Precision-Recall Curve} (AUPRC) swaps a classification threshold between extreme values, measuring precision and recall for each threshold, and creates a precision vs.~recall curve. 
The area under this curve establishes a common performance metric.

The results of all models are presented in \Cref{tab:similar_case_matching}.
We draw the reader's attention to the fact that regex has an undefined AUPRC score since there is no natural way of associating its predictions with probabilities.
Once again, we emphasize that the main goal of this article is to analyze binding precedent efficiency through case classification on Dataset~\#2.
This entire section, devoted to comparing the models on Dataset~\#1 (our reference dataset for validation, since it is labeled), is only one step in our overall approach.
In \Cref{sec:generalization}, we will examine how the models perform on Dataset~\#2, where the outcomes may not align perfectly with the scores reported below.

\begin{table}[!ht]\centering
\begin{tabular}{lllllll}
        \hline
        \textbf{BP}         & \textbf{Model} & \textbf{$F_1$} & \textbf{Precision} & \textbf{Recall} & \textbf{AUPRC} \\
        \hline
        \multirow{6}{*}{11} 
& TF-IDF+SVM  & \textbf{98.2} & 98.8 & \textbf{97.6} & \textbf{98.8} \\
& TF-IDF+logistic  & 96.9 & \textbf{100} & 94.0 & \textbf{98.8} \\
& TF-IDF+forest  &  91.5 & \textbf{100} & 84.3 & 98.6 \\
& LSTM &  91.1 & 89.5 & 92.8 & 94.1 \\
& Longformer &  95.1 & 97.5 & 92.8 & 96.2 \\
& regex  &  91.6 & 98.6 & 85.5 & $\times$ \\
\hline
        \multirow{6}{*}{14} 
& TF-IDF+SVM & \textbf{99.1} & \textbf{100} & \textbf{98.1} & \textbf{99.9} \\ 
& TF-IDF+logistic & 98.7 & \textbf{100} & 97.5 & \textbf{99.9} \\ 
& TF-IDF+forest & 96.2 & 98.7 & 93.8 & 99.8 \\
& LSTM & 94.9 & 91.8 & \textbf{98.1} & 99.6 \\
& Longformer & 98.1 & 98.7 & 97.5 & 99.4 \\
& regex & 78.7 & 98.1 & 65.6 & $\times$ \\
\hline
        \multirow{6}{*}{17} 
& TF-IDF+SVM & \textbf{98.8} & \textbf{100} & \textbf{97.7} & \textbf{100} \\
& TF-IDF+logistic & \textbf{98.8} & \textbf{100} & \textbf{97.7} & \textbf{100} \\ 
& TF-IDF+forest & 95.8 & \textbf{100} & 91.9 & \textbf{100} \\
& LSTM & 95.2 & 98.8 & 91.9 & 97.9 \\
& Longformer & \textbf{98.8} & \textbf{100} & \textbf{97.7} & 98.6 \\
& regex & 92.9 & 94.0 & 91.9 & $\times$ \\
\hline 
        \multirow{6}{*}{26} 
& TF-IDF+SVM & \textbf{98.9} & \textbf{100} & \textbf{97.9} & \textbf{99.1} \\
& TF-IDF+logistic & 97.3 & \textbf{100} & 94.7 & 98.9 \\ 
& TF-IDF+forest & 85.5 & \textbf{100} & 74.7 & 98.9 \\ 
& LSTM & 97.8 & \textbf{100} & 95.8 & 98.7 \\
& Longformer & 94.6 & 97.8 & 91.6 & 97.0 \\ 
& regex & 76.9 & 98.4 & 63.2 & $\times$ \\
\hline
        \multirow{6}{*}{37}
& TF-IDF+SVM & 96.9 & \textbf{100} & 94.0 & 99.6 \\
& TF-IDF+logistic & 96.1 & \textbf{100} & 92.5 & 99.6 \\ 
& TF-IDF+forest & 91.4 & \textbf{100} & 84.2 & 99.5 \\ 
& LSTM & 96.5 & 99.5 & 93.7 & 98.2 \\
& Longformer & \textbf{97.2} & 99.2 & \textbf{95.2} & \textbf{99.9} \\
& regex & 86.7 & 93.3 & 80.9 & $\times$ \\
\hline
\end{tabular}
    \caption{Test scores for the task of binary classification in Dataset~\#1.
    For each BP and each model, we report the metrics $F_1$, precision, recall, and AUPRC (in percentage). 
    The binary scores ($F_1$, precision and recall) are computed from the models' raw outputs, whereas the AUPRC is computed from their probability outputs. 
    As described in the text, the models' binary outputs are obtained by applying a threshold of 0.5 to their probability outputs.
    For each BP, the best score in each column is shown in bold.}
    \label{tab:similar_case_matching}
\end{table}

\subsubsection{Discussion}

One sees from \Cref{tab:similar_case_matching} that the TF-IDF model, equipped with the classifier SVM, consistently achieves the best performance across all BPs, measured by the $F_1$ score, with the exception of BP~37, where it swaps places with Longformer.
TF-IDF vectorization also performs well with logistic regression, reaching second place (or tied for first) in BPs~11, 14, and 17.
In the case of BP~26, LSTM does better, and for BP~37, both LSTM and Longformer have higher scores.
On the other hand, the deep learning models LSTM and Longformer show slightly lower scores overall, but they are still competitive.
The strong performance of TF-IDF in this task has already been reported in \cite{resck2022legalvis}, using, as we do, Brazilian legal documents (more precisely, data from \textit{Supremo em Números}).
Namely, it has been observed that the TF-IDF-based models outperform certain more modern embeddings (such as Doc2vec, Universal Sentence Encoder, and Longformer). 
In this context, TF-IDF's superior performance is not simply attributed to the fact that deep learning models are limited by the number of tokens --- we saw in \Cref{subsection:dataset} that the limitation of 4096 tokens allows almost all documents to be read in their entirety (see \Cref{subfig:length_of_documents_lemmatization,subfig:length_of_documents_tokenization}) ---, but its ability to leverage specific words that allow it to identify the BP.
This is supported by the regex search: although being a rather simple way of answering the problem of Case Classification, its performance is not too bad, even outperforming LSTM for BP~11.

The case of TF-IDF coupled with random forest is interesting: it achieves perfect precision across all BPs.
That is to say, it is highly reliable when it comes to positive predictions.
On the other hand, its recall is the lowest in every case, not including regex.
In other words, this model is exposed to the problem of missing potentially relevant documents. 
This will be observed in \Cref{sec:generalization} when applied to Dataset~\#2, where its usefulness will then be limited.

\subsection{Explainability}\label{subsec:explainability}

To understand further what the models have learned, we move on to study their most important features.
For the TF-IDF models, common measures of importance can be directly computed from the models.
For Longformer, however, there is no direct way of identifying the most important features, thus we will use the explainability algorithm LIME \cite{ribeiro_why_2016}.
We point out that, although not presented here, a similar analysis could be made with LSTM. 

\subsubsection{Explainability with TF-IDF}
For the classifiers SVM, logistic regression, and random forest, based on the TF-IDF vectorization, common measures of importance of features are respectively the weights of the linear kernel, the coefficients in the decision function, and the standard deviation of impurity decrease in the trees.
We compute these quantities via the native functions of \texttt{scipy}. 
We inspect, for each model, the top features, and gather those common to all models.
\Cref{tab:importance_features} presents the importance of the words selected this way. 

\begin{table}[!ht]\centering
\begin{tabular}{p{0.2cm}llllll}
\hline
\textbf{BP} & \textbf{Model} & \textbf{Features} &&&&\\*
\hline
\multirow{5}{*}{11}
&& \textit{algemado} & \textit{algemas} & \textit{audiência} & \textit{nulidade} & \textit{uso} \\*
&&  \footnotesize{(handcuffed)} & \footnotesize{(handcuffs)} & \footnotesize{(court hearing)} & \footnotesize{(nullity)} & \footnotesize{(use)}\\*
& SVM & 0.41 & 1 & 0.28 & 0.29 & 0.4 \\*
& logistic & 0.38 & 1 & 0.3 & 0.3 & 0.41 \\*
& forest & 0.39 & 0.96 & 0.37 & 0.27 & 0.26 \\
\hline
\multirow{5}{*}{14}
&& \textit{acesso} & \textit{criminal} & \textit{documentados} & \textit{inquérito} & \textit{investigação} \\*
&&\footnotesize{(access)} & \footnotesize{(criminal)} & \footnotesize{(documented)} & \footnotesize{(inquiry)} & \footnotesize{(investigation)}\\*
& SVM & 1 & 0.52 & 0.47 & 0.45 & 0.59 \\*
& logistic & 1 & 0.49 & 0.43 & 0.49 & 0.59 \\*
& forest & 1 & 0.27 & 0.54 & 0.65 & 0.48 \\
\hline
\multirow{5}{*}{17}
&& \textit{cálculos} & \textit{juros} & \textit{mora} & \textit{moratórios} & \textit{precatórios} \\*
&& \footnotesize{(calculations)} & \footnotesize{(interest)} & \footnotesize{(late)} & \footnotesize{(moratoriums)} & \footnotesize{(court order)} \\*
& SVM & 0.55 & 0.88 & 0.68 & 0.48 & 1 \\*
& logistic & 0.57 & 0.98 & 0.72 & 0.52 & 1 \\*
& forest & 0.18 & 0.9 & 0.62 & 0.68 & 1 \\
\hline
\multirow{5}{*}{26}
&& \textit{criminológico} & \textit{exame} & \textit{execuções} & \textit{progressão} & \textit{regime} \\*
&&\footnotesize{(criminological)} & \footnotesize{(examination)} & \footnotesize{(executions)} & \footnotesize{(progression)} & \footnotesize{(regime)}\\*
& SVM & 1 & 0.48 & 0.58 & 0.84 & 0.58 \\*
& logistic & 1 & 0.5 & 0.61 & 0.87 & 0.58 \\*
& forest & 1 & 0.57 & 0.38 & 0.75 & 0.32 \\
\hline
\multirow{5}{*}{37}
&& \textit{aumentar} & \textit{isonomia} & \textit{servidores} & \textit{simetria} & \textit{vencimentos} \\*
&&\footnotesize{(increase)} & \footnotesize{(isonomy)} & \footnotesize{(servants)} & \footnotesize{(symmetry)} & \footnotesize{(salaries)}\\*
& SVM & 0.66 & 0.95 & 0.16 & 0.89 & 0.46 \\*
& logistic & 0.58 & 1 & 0.2 & 0.67 & 0.57 \\*
& forest & 1 & 0.9 & 0.41 & 0.19 & 0.6 \\
\hline
\end{tabular}
\caption{For each BP and each TF-IDF model, we computed the importances of the features, and selected the five highest ones.
The values are normalized so that the largest importance is equal to one.}
\label{tab:importance_features}
\end{table}

The results of \Cref{tab:importance_features} are not surprising: the most important words are already contained in the wording of the BPs or are related grammatically, except five (\textit{audiência}, \textit{criminal}, \textit{inquérito}, \textit{cálculos}, and \textit{simetria}, studied below).
We refer the reader to \Cref{subsec:BP_11,subsec:BP_14,subsec:BP_17,subsec:BP_26,subsec:BP_37} for the wordings, as well as a more detailed analysis of the results.
It is remarkable that the three classifiers display similar importance for almost each of the words. 
This shows that they have all faithfully learned the statements of the BPs.

The first feature to stand out is \textit{audiência} (audience/trial), undirectly connected to BP~11. 
As we will explain in \Cref{subsec:BP_11}, this precedent was born out of the controversy surrounding the use of handcuffs (\textit{algemas}) in trials. 
It is therefore consistent that this word has been recognized as important by the models.

The second word, \textit{criminal}, is naturally associated with BP~14, as its field of action is criminal law. In this context, the word is often used in expressions such as \textit{persecução criminal} (criminal prosecution), \textit{instrução criminal} (criminal investigation), or \textit{responsabilidade criminal} (criminal liability).
More precisely, 75, 148, and 35 documents in Dataset~\#1 cite these expressions, respectively.
This amounts to 225 documents, 67.5\% of which cite BP~14.
On the other hand, it is also not surprising to find the word \textit{inquérito} (inquiry/investigation) in BP~14's features, since it can be viewed, in this context, as a synonym for \textit{procedimento investigatório} (investigative procedure), which is at the heart of the binding precedent's formulation.

For BP~17, a word not present in its wording is detected as important by the models: \textit{cálculos} (calculations). 
In \Cref{subsec:BP_17}, we will present the context of the creation and application of this binding precedent, which arose from a debate about late payment interests.
The use of this word is natural in this context, as in the expression ``calculation of interests''. 
For an example of its usage, Justice Marco Aurélio published a clarification of this precedent under Thesis 96\footnote{Thesis 96: ``Late payment interest applies to the period between the date of the calculations and the date of the requisition or the precatory order''  \url{https://portal.stf.jus.br/jurisprudenciaRepercussao/verAndamentoProcesso.asp?incidente=2598262&numeroProcesso=579431&classeProcesso=RE&numeroTema=96}}, explicitly using the word \textit{cálculos}. 
This thesis is reproduced in 79 documents within the dataset.

Finally, the word \textit{simetria} appears in the expression \textit{princípio da simetria} (principle of symmetry), a legal concept that was used in conjunction with BP~37 in the context of a controversy regarding the salaries of the Judiciary (\textit{magistratura}). This is studied in detail in \Cref{subsec:BP_37}.

In conclusion, these few examples demonstrate that, as was hoped when Dataset~\#1 was designed, the models are capable of learning not only the statements of the precedents but also some of their context and the themes they cover.
Remarkably, these features are highly interpretable: they are words that cover general themes specific to each BP, rather than purely syntactic peculiarities, which would be characteristic of overfitting.

\subsubsection{Explainability with LIME}

In the context of explainability, instead of directly assessing the weights of an interpretable model (e.g., a logistic regression), we can treat the model as a black box and verify its behavior when we disturb its input. For instance, one could be interested in how much the output probability changes if one removes certain words from the text. If it changes significantly, these words are deemed important for the model decision.

This is precisely the underlying mechanism of LIME (Local Interpretable Model-agnostic Explanations) \citep{ribeiro_why_2016}, a popular Machine Learning explainability method. 
More precisely, to explain the model decision for a specific text sample, LIME perturbs the input of the classifier by randomly removing words from the text and measures the model output probability. Then, after thousands of disturbances, LIME fits a linear regression to predict the model probability from the word presence. From the linear regression coefficients, one assigns an importance score for each word in the input text regarding their contribution to the model decision.

In practice, we use LIME as follows: for each document in the collection, we calculate with LIME the $N$ most important words (i.e., words which contribute most to the positive class), where $N$ is a hyperparameter, chosen as $N=5$. 
We then search, among all the documents belonging to a BP, for the most frequent features, and the number of times they appear. The $N$ most important features are represented in \Cref{tab:LIME_features}.

\begin{table}[!ht]\centering
\begin{tabular}{lllllll}
\hline
\textbf{BP} & \textbf{Features} &&&& \\
\hline
\multirow{3}{*}{11}   
& \textit{algemas} & \textit{algemado} & \textit{algemados} & \textit{despacho} & \textit{reclamação} \\
& \footnotesize{(handcuffs)} & \footnotesize{(handcuffed)} & \footnotesize{(handcuffed, plural)} & \footnotesize{(procedural ruling)} & \footnotesize{(complaint)} \\
& 66.1 & 22.3 & 4.6 & 4.0 & 1.9 \\
\hline
\multirow{3}{*}{14}   
& \textit{acesso} & \textit{autos} & \textit{criminal} & \textit{inquérito} & \textit{defesa} \\
& \footnotesize{(access)} & \footnotesize{(elements)} & \footnotesize{(criminal)} & \footnotesize{(inquiry)} & \footnotesize{(defense)} \\
& 26.3 & 19.9 & 17.0 & 16.4 & 13.4 \\
\hline
\multirow{3}{*}{17}   
& \textit{juros} & \textit{precatório} & \textit{moratórios} & \textit{precatórios} & \textit{mora} \\
& \footnotesize{(interests)} & \footnotesize{(court order)} & \footnotesize{(moratoriums)} & \footnotesize{(court orders)} & \footnotesize{(late)} \\
& 53.5 & 41.8 & 19.2 & 18.9 & 17.5 \\
\hline
\multirow{3}{*}{26}   
& \textit{criminológico} & \textit{regime} & \textit{SP} & \textit{exame} & \textit{geral} \\
& \footnotesize{(criminological)} & \footnotesize{(regime)} & \footnotesize{(State of São Paulo)} & \footnotesize{(examination)} & \footnotesize{(general)} \\
 & 44.5 & 36.2 & 19.4 & 15.4 & 11.3 \\
\hline
\multirow{3}{*}{37}   
& \textit{isonomia} & \textit{vencimentos} & \textit{2015} & \textit{magistratura} & \textit{aumentar} \\
& \footnotesize{(isonomy)} & \footnotesize{(salaries)} & \footnotesize{(2015)} & \footnotesize{(Judiciary members)} & \footnotesize{(to increase)} \\
& 32.7 & 13.2 & 9.0 & 8.6 & 8.5 \\ 
\hline
\end{tabular}
\caption{For each document, LIME detects the most important features. For each BP, we then select the five most frequent features and indicate the percentage of documents in which LIME detected them.}
\label{tab:LIME_features}
\end{table}

The table offers several interesting observations. 
First, and as expected, one sees that many words are shared with the TF-IDF models (presented in \Cref{tab:importance_features}).
However, in the latter models, we noticed that the important features were almost exclusively words already present in the BP statement. As a matter of fact, among the five words that were not, two were also detected by LIME (\textit{criminal} and \textit{inquérito}).
In the case of Longformer, LIME brings to light features that, although highly relevant to the identification of the legal theme covered by the BP, do not appear in its wording. 

This is the case for BP~11: the words \textit{despacho} (procedural ruling) and \textit{reclamação} (appeal/complaint) indicate that the model Longformer tends to favor documents whose type is \textit{Reclamação}, and which have not been judged on their merits, but deferred for procedural reasons.
This is precisely what will be observed in the dedicated legal section, and explains the recent use of this BP (see \Cref{subsec:BP_11}).

BP~26, presenting the words \textit{SP} and \textit{geral}, is also noteworthy.
The first refers to the State of São Paulo, which, as studied in \Cref{subsec:BP_26}, is precisely where most of the debate surrounding the BP is taking place.
Equally relevant, the word \textit{geral} is found in \textit{Procurador-Geral} or \textit{Procuradoria-Geral} (General Prosecutor, General Prosecutor's Office), present in 58.9\% of documents citing BP~26. 
In fact, prosecutors play a particular role in the context of BP~26, since it is they who may request a criminological examination, an issue that has become central to the abovementioned debate.

Finally, LIME detects two words in the last BP that are not part of its statement: \textit{2015} and \textit{magistratura} (Judiciary). As we will see in \Cref{subsec:BP_37}, BP~37 is paved with requests by different groups of litigants; in particular, major peaks include one in 2015 (from temporary professors) and another in 2017 (from Judiciary members).

\section{Generalization on Dataset~\#2}\label{sec:generalization}

In the previous section, we trained models on the first dataset (documents citing a BP).
Now we'd like to assess the quality of their predictions when applied to the second dataset (the whole collection of documents emitted by the Supreme Court in the selected branches).
That is to say, we seek to verify their ability to generalize their results when applied to a larger dataset.
First of all, a fine-tuning step is required, as detailed in \Cref{subsec:evaluating_generalization}. 
Then, we will check the quality of the predicted documents by three means: the study of the time series of predicted documents, the correlation between the words of these documents, and a manual reading of a few samples.
This analysis will be carried out precedent by precedent, in \Cref{subsec:manual_evaluation}.

\subsection{Evaluating the generalization of the models}\label{subsec:evaluating_generalization}

\subsubsection{Fine-tuning}
When applying the models, trained on Dataset~\#1 (29,743 documents) to Dataset~\#2 (634,068 documents), we face a significant problem: the models have been tuned to the first dataset, and can potentially be not adapted to the second.
In fact, we observed that they tend to predict way more documents than intended, i.e., they are underfitted.

To remedy this problem, we will not consider the binary output of the models, but rather their ``probabilities''.
As described in \Cref{subsec:validation}, as a native property of the models we considered, each document is associated with a likeliness of belonging to the positive class, through a value between 0 and 1.
Internally, a document is assigned the positive class if this value exceeds the threshold of $0.5$.
To adapt the models to the new dataset, one should increase this threshold, to discard the false positives.

However, as already pointed out in \Cref{subsection:dataset}, Dataset~\#2 does not come with indications of whether a document falls within the scope of a certain BP.
Thus, this dataset cannot be used directly to tune the thresholds.
Instead, we will take the following observation as a reference: if a document, issued \textit{after} the publication of the BP, is detected by the models, then it probably cites the BP.
In other words, we can take the labels of Dataset~\#1 as groundtruth, valid only after the publication of the BP.
A well-tuned model is expected to follow faithfully the curve of citations post-BP.

In practice, we chose the new thresholds as the largest value such that the recall of the predictions, after the publication of the BP, is greater or equal to a fixed value $p\in[0,1]$.
That is, we impose that at least a proportion $p$ of the groundtruth documents are detected.
Note that the larger $p$ is, the more documents are detected.
Specifically, we chose the values $p=0.9$ for BPs 11, 17, and 37, and $p=0.7$ for BPs 14 and 26.
Although arbitrary, these values have been chosen in accordance with the models' performance in \Cref{tab:similar_case_matching}, and because they yield satisfactory results.
We indicate in \Cref{tab:adjusted_probabilities}, for each model and each BP, the tuned value of the thresholds.
For all cases, the tuned threshold is higher than $0.5$, the native threshold, implying a model with fewer positive predictions.

\begin{table}[!ht]\centering
\begin{tabular}{llllll}
        \hline
        \textbf{Model}      & \textbf{BP~11}      & \textbf{BP~14}    & \textbf{BP~17}     & \textbf{BP~26}      & \textbf{BP~37}     \\\hline 
TF-IDF+SVM & 0.999 & 0.995 & 0.999 & 0.999 & 0.987 \\
TF-IDF+logistic & 0.998 & 0.999 & 0.998 & 0.999 & 0.951 \\
TF-IDF+forest & 0.7 & 0.869 & 0.697 & 0.893 & 0.74 \\
LSTM & 0.998 & 0.998 & 0.997 & 0.999 & 0.995 \\
Longformer & 0.615 & 0.999 & 0.999 & 1 & 0.997 \\
\hline
\end{tabular}
\caption{For each model and each BP, the threshold over which a document is assigned the positive class.
The thresholds refer to those tuned on Dataset~\#2 and are given with three decimal places.}
\label{tab:adjusted_probabilities}
\end{table}

\subsubsection{Time series of similar cases}

Our main tool for visualizing model predictions on Dataset~\#2 will be time series, displayed as curves. More precisely, we will represent the quantity of documents predicted by each model as a function of their publication date, in the form of a histogram, which is then interpolated by a curve.
This representation has the advantage of revealing considerable information. 
Firstly, from a purely methodological point of view, we can assess the predictive quality of the models: as explained above, we expect a fine-tuned model to follow closely the curve of documents citing the BP considered (we call it the groundtruth curve), at least after the date of publication of this BP.
Before the publication date, other phenomena can be observed. We expect certain legal events to induce \textit{peaks} in these curves, which can then be identified visually, and investigated further by reading the respective documents, or by adding additional metadata.
These curves also show the overall trends in BP citations, which we will use for our legal analysis in \Cref{sec:legal_analysis}.
Lastly, by plotting the prediction curves of several models on the same graph, we obtain greater confidence in the legal events detected. 
Indeed, if several models present a peak at the same time, it may well be a quantity of documents effectively linked to the content of the BP.
On the other hand, if a trend is exhibited by only one model, we might suspect that it is the result of poor training.

We specify that not all the documents detected by a model will be considered, but only those belonging to the branch of the BP under consideration --- ``Administrative Law'' for BPs 17 and 37, and ``Criminal Law'' or ``Criminal Procedure Law'' for BPs 11, 14, and 26. 
Indeed, since the scope of a BP is reduced to a precise branch, as described in \Cref{subsection:dataset}, documents predicted outside this category would be false positives, and therefore irrelevant. We have observed that, in practice, this adjustment brings only a minor change in the prediction curves.

Analyzing these curves, and assessing whether the predicted documents are relevant, will be the content of \Cref{subsec:manual_evaluation}.
In particular, the time series will be given in \Cref{fig:predictions_allmodels_SV11,fig:predictions_allmodels_SV14,fig:predictions_allmodels_SV17,fig:predictions_allmodels_SV26,fig:predictions_allmodels_SV37} for BP~11, 14, 17, 26 and 37, respectively.

\subsubsection{Correlations of words}

Besides, to evaluate the quality of the predictions, another tool will be employed: the correlations between the words.
Given a specific word (for instance, a relevant word associated with the BP or from its wording), the presence or absence of it in a document can be seen as a binary variable. 
From this point of view, the correlation between two words indicates how often they are found combined.
Since there is little difference in the way two documents on a similar subject are written --- it is often observed that some decisions are merely copies of another ---, the collection of documents predicted by a model is expected to show similar correlations compared to the collection of documents citing the BP.
Comparing these measurements will allow us to check the models' accuracy and to find out why some do not perform well.

\subsection{Manual evaluation}\label{subsec:manual_evaluation}

In this section, we will analyze the predictions of our models precedent by precedent, in order to check that they are correct, i.e., that the predicted documents are indeed related to the content of the precedent.
As we will see, some sets of predictions will need to be discarded. 
This section is therefore intended as a validation phase for the models, enabling the final legal analysis of the precedents in \Cref{sec:legal_analysis}. 
To reserve legal considerations for the final section, we will aim to minimize here references to the legal content of the documents and instead focus primarily on analyzing them through the lens of time series, word correlations, and regular expressions.

\subsubsection{Binding Precedent 11}

\begin{figure}[!ht]\centering
\footnotesize{Models' predictions for BP~11}
\includegraphics[width=.99\linewidth]{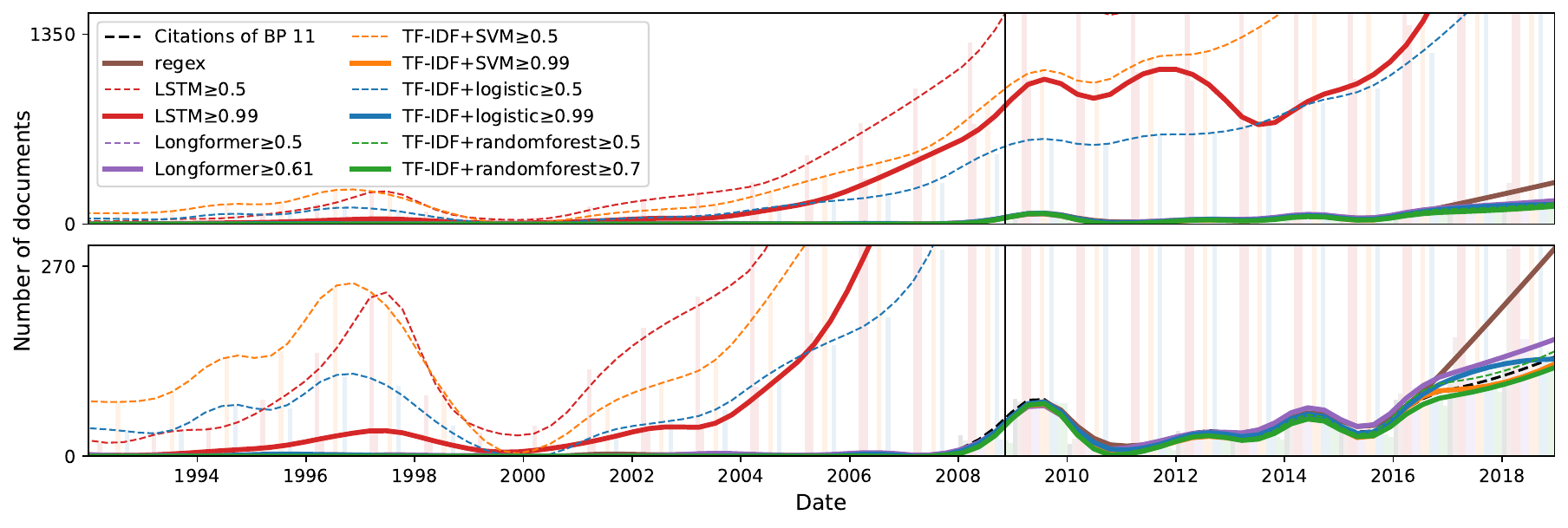}
\caption{Number of documents predicted by each model for BP~11 in Dataset~\#2, represented as a histogram (window length of one year) and interpolated via quadratic spline. 
We give the predictions for the thresholds adapted to Dataset~\#1 (dashed) and Dataset~\#2 (solid), as given in \Cref{tab:adjusted_probabilities}.
Two views are given, the second one zooming in on the ordinate axis.}
\label{fig:predictions_allmodels_SV11}
\end{figure}

We show in \Cref{fig:predictions_allmodels_SV11} the predictions of the five models on our whole collection (Dataset~\#2), as well as a regex search for documents containing the words \textit{algemas} or \textit{algemado} (handcuffs or handcuffed).
As explained in \Cref{subsec:evaluating_generalization}, the thresholds of the models, trained on Dataset~\#1, are not adapted for Dataset~\#2: way too many documents are predicted.
Therefore, we choose new thresholds with the following rule: the value such that 90\% of the documents citing BP~11 are predicted.
In the figure, the initial and the fine-tuned prediction curves are visualized as dashed or solid, respectively.

As it turns out, there are almost no documents, before the publication of BP~11, containing the words of the regex query.
This phenomenon is also illustrated by the models TF-IDF+SVM, TF-IDF+logistic, TF-IDF+random forest, and Longformer, which closely follow the groundtruth curve.
As a matter of fact, among the 1056 documents of Dataset~\#2 containing the words \textit{algemas} or \textit{algemado}, respectively 656, 679, 656, and 814 were predicted by the models, accounting for $88.3\%$, $82.9\%$, $93.6\%$, and $91.2\%$ of their predictions.
In these documents, these words are often used in conjunction with \textit{uso} and \textit{nulidade} (use and nullity), also present in the official wording of BP~11.
These observations illustrate the fact that the documents processed by STF have only begun to mention ``handcuffs'' since the publication of BP~11. 
This is an interesting legal fact, which we will explore in \Cref{subsec:BP_11} dedicated to legal analysis.

The model LSTM, however, estimates thousands of documents.
Only 4.4\% of its predictions contain the words \textit{algemas} or \textit{algemado}.
A manual inspection shows that this model tends to detect documents merely containing the words \textit{uso} or \textit{nulidade}, independently of the surrounding context.
For instance, among the 52,465 documents containing the word \textit{uso}, 9586 are associated with the positive class, accounting for 60.4\% of the predicted documents.
This suggests that LSTM is biased toward detecting this word, regardless of its role in the conjunction \textit{uso de algemas} (use of handcuffs).

\begin{figure}[!ht]\centering
\footnotesize{Correlations between words among the models' predictions for BP~11}
\includegraphics[width=.90\linewidth]{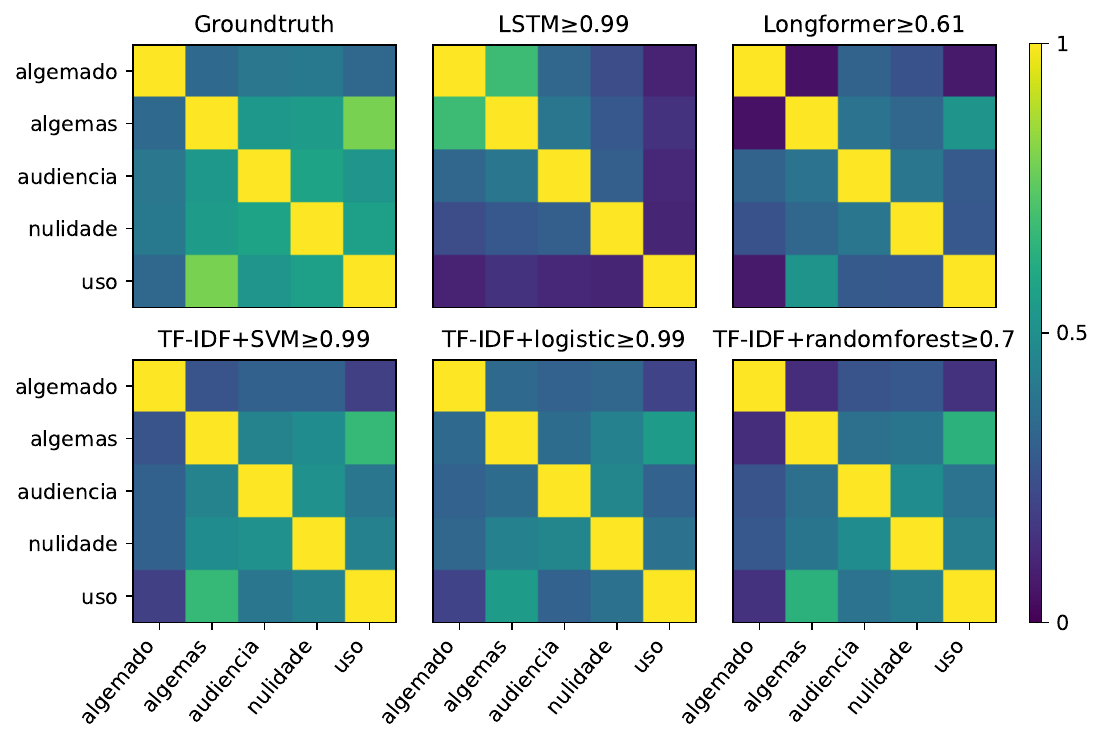}
\caption{Correlations between the selected relevant words, when restricted to documents citing BP~11 (``Groundtruth'') or documents predicted by one of our models.
Note that, as discussed in \Cref{{subsec:evaluating_generalization}}, correlations are calculated by seeing the words as binary random variables, hence correlation values lie between 0 and 1.
}\label{fig:correlations_SV11}
\end{figure}

To analyze this phenomenon further, let us consider the TF-IDF's most relevant words, obtained in \Cref{tab:importance_features}: \textit{algemado, algemas, audiência, nulidade}, and \textit{uso} (handcuffed, handcuffs, court hearing, nullity, use).
We give in \Cref{fig:correlations_SV11} the correlations between these five words when restricting the dataset to the documents explicitly citing BP~11, as well as the documents estimated by the models.
As explained in \Cref{{subsec:evaluating_generalization}}, a well-fitted model is expected to show similar correlations as the groundtruth.
The first image shows a high correlation between \textit{uso} and \textit{algemas}, indicating that they mainly belong to the set \textit{uso de algemas}. 
Besides, only LSTM does not reflect this behavior. 
This observation accounts for the fact that this model, by not taking into account the correlation between these words, shows poor results when used on the larger dataset.

Finally, we turn our attention to the jump in regex predictions, after 2016, which is not part of the groundtruth (i.e., the documents do not cite BP 11), seen in \Cref{fig:predictions_allmodels_SV11}.
A manual inspection of these documents reveals that, although they contain the words \textit{algemas} or \textit{algemado}, are not directly related to the topic of BP~11.
As a matter of fact, these documents mention the (non-binding) Precedent~279 \footnote{Precedent~279: ``For the mere reexamination of evidence, an extraordinary appeal (ARE) is not admissible'' \url{https://portal.stf.jus.br/jurisprudencia/sumariosumulas.asp?base=30&sumula=2174}} (\textit{Súmula 279}).
In addition to Precedent~279, the documents cite prior decisions based on it, and in particular ARE 965.920\footnote{ARE 965.920 \url{https://redir.stf.jus.br/paginadorpub/paginador.jsp?docTP=TP&docID=11766746}}, which contains the word \textit{algemas}.
Consequently, the regex search returned these documents.
However, this detection is merely coincidental: the question of handcuffs is not at the core of the documents. 
This illustrates the limitations of a simple regex search, as well as the importance of verifying, by reading, the predictions obtained.
In particular, we will not use these predictions to conduct our legal analysis.
To illustrate the issue more specifically, we highlight that ARE 965.920 is a case where BP~11 and Precedent~279 intersect. 
The decision’s rationale indicates that demonstrating the improper use of handcuffs (the subject of BP~11) would require reanalyzing evidence, which falls under the scope of Precedent~279. Consequently, the non-binding precedent was invoked to justify denying the appeal.

\subsubsection{Binding Precedent 14}

\begin{figure}[!ht]\centering
\footnotesize{Models' predictions for BP~14}
\includegraphics[width=.99\linewidth]{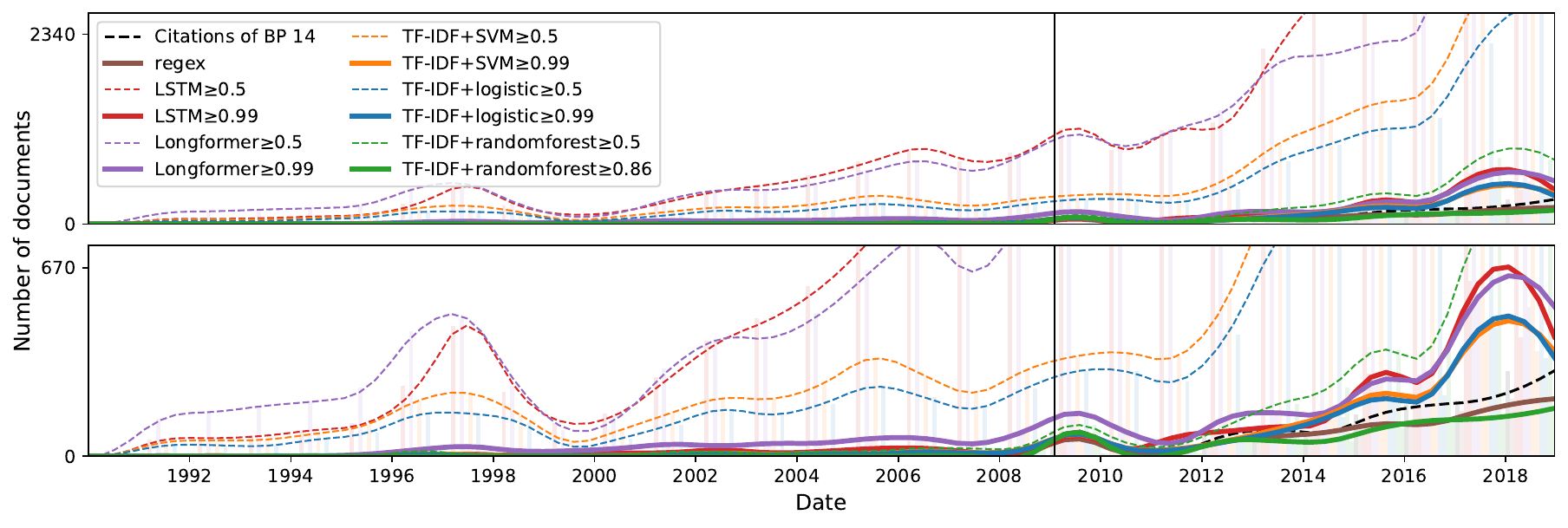}
\caption{Number of documents predicted by each model for BP~14 in Dataset~\#2, represented as a histogram (window length of one year) and interpolated via quadratic spline. 
We give the predictions for the thresholds adapted to Dataset~\#1 (dashed) and Dataset~\#2 (solid), as given in \Cref{tab:adjusted_probabilities}.
Two views are given, the second one zooming in on the ordinate axis.}
\label{fig:predictions_allmodels_SV14}
\end{figure}

The prediction curves for BP~14 are displayed in \Cref{fig:predictions_allmodels_SV14}.
We remind the reader that, for this precedent, we tuned the models' thresholds as those such that 70\% of the documents citing BP~14 are predicted, and not 90\%, as is the case for BPs 11, 17, and 37 (see \Cref{tab:adjusted_probabilities}). 
A quick analysis of the predicted documents indicates that, among the five models considered, only TF-IDF+random forest achieved the expected result, by faithfully following the groundtruth curve.
This is also the case for the regex search of \textit{acesso aos elementos/autos/documentos} (access to elements, records, or documents).

\begin{figure}[!ht]\centering
\footnotesize{Frequency of words among the models' predictions for BP~14}
\includegraphics[width=.55\linewidth]{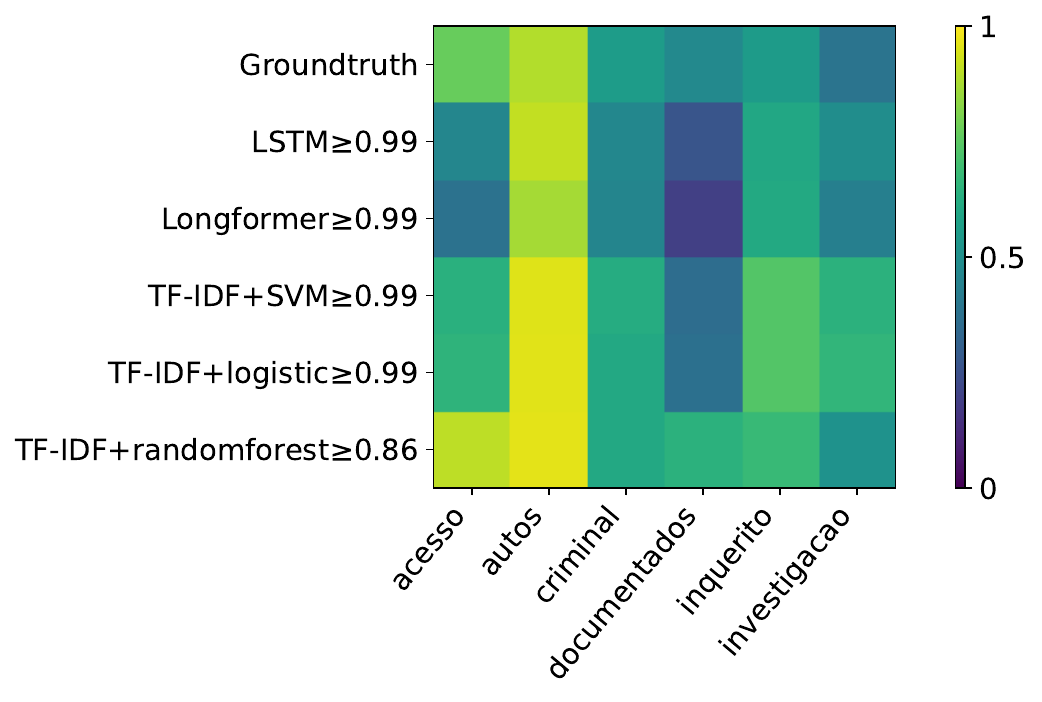}
\caption{Frequency of the selected relevant words, when restricted to documents citing BP~14 (``Groundtruth''), or documents predicted by one of our models.}\label{fig:heatmap_SV14}
\end{figure}

Interestingly, all the other models peaked in 2017/2018.
To investigate why they estimate many documents, we will not consider the correlations, as we did for BP~11 (in \Cref{fig:correlations_SV11}), but simply the frequencies of the most relevant words, represented in \Cref{fig:heatmap_SV14}.
Almost all groundtruth documents cite \textit{acesso aos autos} (access to records).
Similarly, 95.8\% of TF-IDF+random forest's predictions mention \textit{autos}, of which 92.1\% also mention \textit{acesso}.
Besides, among TF-IDF+SVM, TF-IDF+logistic, LSTM and Longformer's predictions, respectively, 95\%, 95.5\%, 90.8\%, and 86.4\% mention \textit{autos}, while only 64.6\%, 66.3\%, 48.7\%, and 41.7\% of them conjointly contain \textit{acesso}.
This suggests some underfitting of the models, which are unable to identify the two words in combination.

\begin{figure}[!ht]\centering
\footnotesize{Additional metadata for documents predicted by Longformer for BP~14}
\includegraphics[width=.99\linewidth]{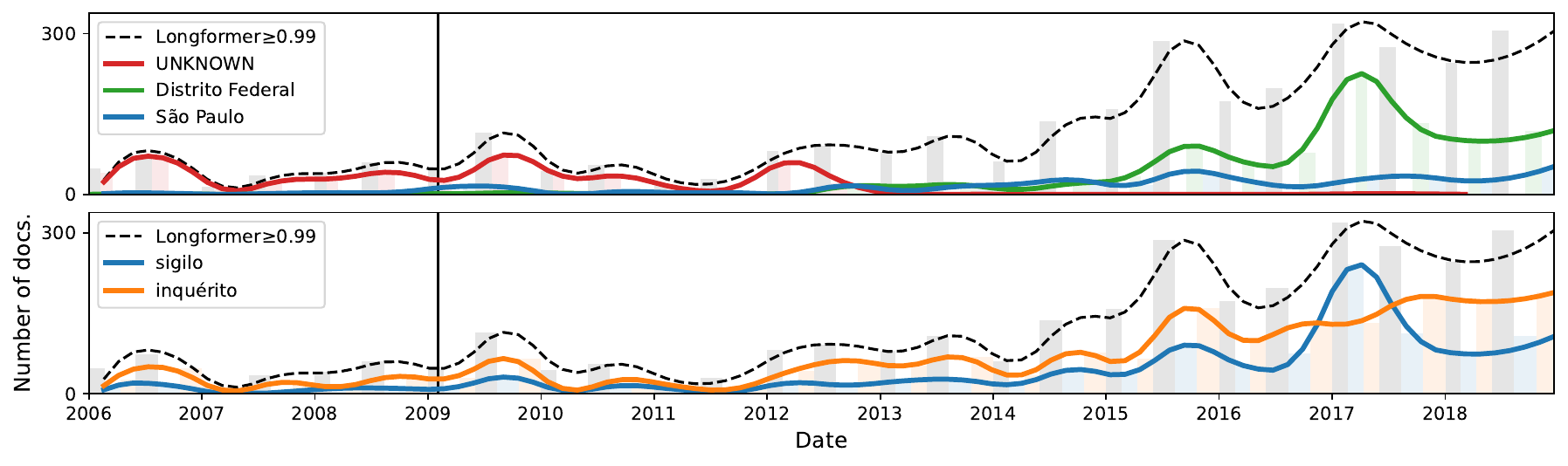}
\caption{\textbf{Top:} State to which the cases predicted by Longformer for BP~14 belong, with label ``UNKNOWN'' when the metadata is unavailable.
\textbf{Bottom:} Regex search for the presence of the words \textit{sigilo} (secrecy/confidentiality) and \textit{inquérito} (inquiry/investigation).
In both graphs, the bins have a length of 6 months.}
\label{fig:predictions_randomforest_SV14_type}
\end{figure}

At first sight, a manual inspection of a sample of the detected documents, excluding elements of the groundtruth class, raises questions about the qualities of these predictions. 
Although certain decisions\footnote{Such as Inq 4.260 \url{https://portal.stf.jus.br/processos/downloadPeca.asp?id=313917544&ext=.pdf} and RHC 133.298 \url{https://portal.stf.jus.br/processos/downloadPeca.asp?id=315014507&ext=.pdf}} are on topics regarding investigations (specifically, on the initiation of an official investigation and the transcription of some telephonic tapping record), neither of them concerns topics related to access of investigation material by the investigated part, which is the subject of the BP.
The addition of metadata, as in \Cref{fig:predictions_randomforest_SV14_type}, shows that this peak mainly consists of cases from \textit{Distrito Federal} that also mention \textit{sigilo} (secrecy/confidentiality).
This suggests that these cases are complaints filed by attorneys representing individuals investigated or condemned for white-collar crimes, most relevantly, by attorneys representing politicians or individuals of political interest. 
It is, therefore, unsurprising that this increase in predictions after 2012 partly matches the interval when \textit{Operação Lava Jato} took place, a joint cooperative investigation aiming at politicians accused of corruption-related crimes.
We continue this discussion, from a legal perspective, in \Cref{subsec:BP_14}.

\subsubsection{Binding Precedent 17}

\begin{figure}[!ht]\centering
\footnotesize{Models' predictions for BP~17}
\includegraphics[width=.99\linewidth]{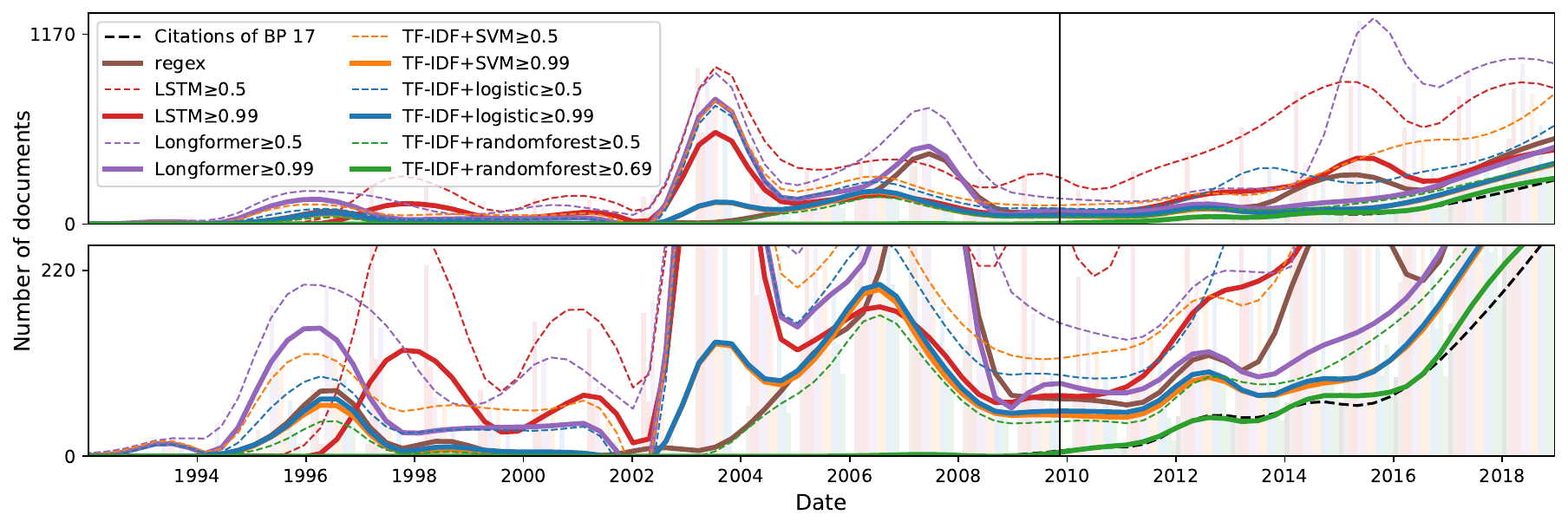}
\caption{Number of documents predicted by each model for BP~17 in Dataset~\#2, represented as a histogram (window length of one year) and interpolated via quadratic spline. 
We give the predictions for the thresholds adapted to Dataset~\#1 (dashed) and Dataset~\#2 (solid), as given in \Cref{tab:adjusted_probabilities}.
Two views are given, the second one zooming in on the ordinate axis.}
\label{fig:predictions_allmodels_SV17}
\end{figure}

\Cref{fig:predictions_allmodels_SV17} presents the predictions of the five models for BP~17, together with a regex search for \textit{precatório} and \textit{juros de mora} (court orders, late payment interest).
Some common trends stand out: a peak between 1995-1998 (by regex, TF-IDF+SVM, and logistic regression), one between 2002-2005 (by LSTM and Longformer, but also TF-IDF+SVM and logistic), one in 2007-2009 (by regex and Longformer), and a last peak in 2014-2016 (after the publication of the BP, by regex and LSTM), as well as an increasing tendency starting from 2011, predicted by all the models.
Besides, LSTM and Longformer individually detect two other peaks: in 1998 and 1996, respectively.

As with all the other binding precedents, BP~17 had random forest for the best model, in the sense that it predicts better the groundtruth curve, after the publication of the precedent.
However, in the context of BP~17, this model is not of interest: the many legal events that took place before the publication of the BP, and which will help us to analyze it in \Cref{subsec:BP_17}, are not captured by TF-IDF+random forest.

\begin{figure}[!ht]\centering
\footnotesize{Correlations between words among the models' predictions for BP~17}
\includegraphics[width=.9\linewidth]{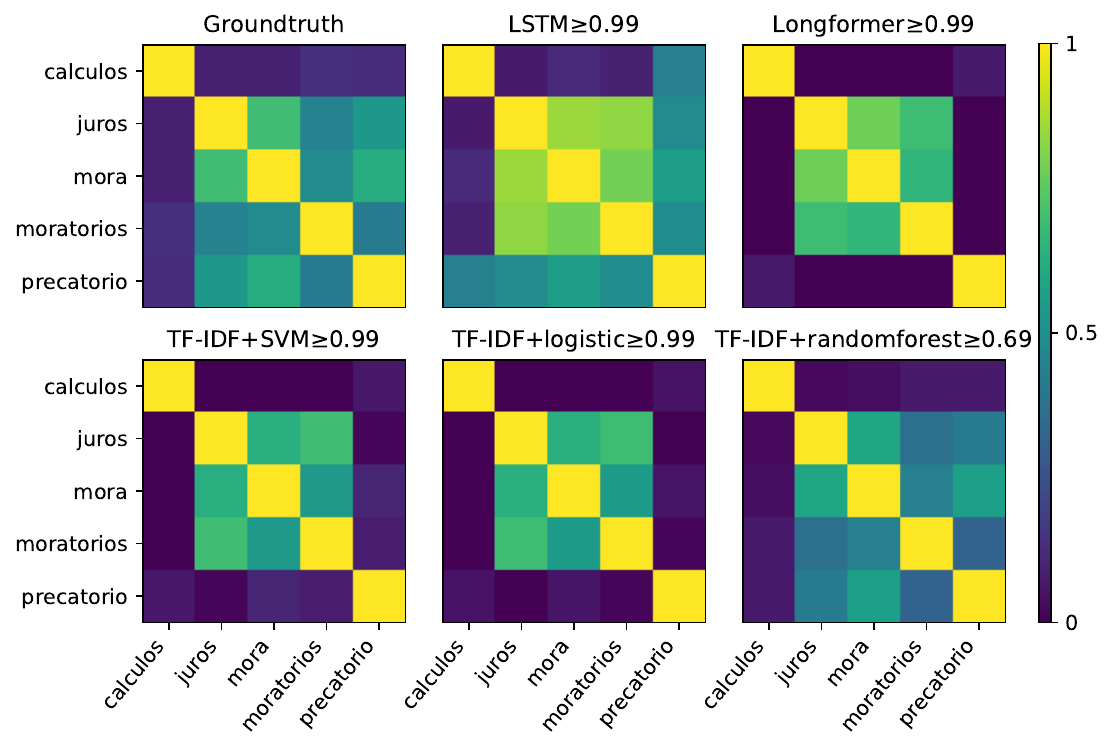}
\caption{Correlations between the selected relevant words, when restricted to documents citing BP~17 (``Groundtruth''), or documents predicted by one of our models.}\label{fig:correlations_SV17}
\end{figure}

The difference between the model's predictions is directly explained by the words correlation in \Cref{fig:correlations_SV17}: except TF-IDF+random forest (and also LSTM to a lesser extent), none of the models accurately captured the correlation of \textit{precatório} (court order payment) with the others words (\textit{mora}, late payment).
In particular, the most important peak, between 2002-2005, is principally composed of documents mentioning ``court order'' but not ``late payment''. 
This explains the absence of regex's predictions in this period, once more highlighting the limitations of a simple regex search.
This peak, even though not referring to cases directly linked to the theme of BP~17, contains, as we will analyze, relevant information.

Another interesting observation is the significant correlation, in most of the models, between the terms \textit{moratório} and \textit{juros}/\textit{mora}.
The term \textit{moratório} (moratorium) can, indeed, be used as a synonym for \textit{mora} (late).
However, it seems that the models have imposed that predicted documents, when they contain the latter, must contain the former.
This would explain why regex's peak in 2007, mainly composed of documents citing \textit{juros de mora} but not \textit{moratório}, has only been detected by Longformer.

As it turns out, Longformer's peak in 1996 consists solely of correct predictions, as confirmed by manual reading.
Their analysis (involving Article 33 of the Act
of Transitory Constitutional Dispositions) will be carried out in \Cref{subsec:BP_17} (\Cref{fig:citations_SV17_Longformer_type}).
Interestingly, although 120 of these documents contained both the terms \textit{precatório} and \textit{juros de mora}, being therefore also detected by regex, many others had only the word \textit{precatórios} and reference to Article 33. In this sense, the neural network model could identify cases strongly related to the BP, but for which our regex terms would be blind. 

LSTM's 1998 peak is more challenging to understand. All documents correspond to appeals promptly rejected by the STF based on procedural arguments, with little or no analysis of the materiality of the cases. 
In addition, of the 221 documents detected by LSTM between 1997 and 1999, only 10 contained the term \textit{precatório} and two contained \textit{mora}, although, surprisingly, 31 contained terms related to monetary corrections of inflation, which, in principle, has little to do with the BP. Because all these documents were particularly short (with an average of 91 tokens), it is possible to hypothesize that LSTM was, for some reason, biased towards short decisions for this particular BP, although we do not further explore this possibility here.

To conduct our legal analysis, we will examine in greater depth the predictions offered by Longformer and regex; the addition of metadata will help reveal the underlying legal context (see \Cref{fig:citations_SV17_Longformer_type,fig:citations_SV17_regex_type}, respectively).
These two models cover all the peaks observed in \Cref{fig:predictions_allmodels_SV17}, with the exception of LSTM in 1998, deemed irrelevant.
More specifically, we will study the peaks detected separately by Longformer and regex in 2002-2005 and 2014-2016, respectively, and those detected jointly in 1996 (the latter's predictions are a subset of the former) and 2007-2009.

\subsubsection{Binding Precedent 26}

\begin{figure}[!ht]\centering
\footnotesize{Models' predictions for BP~26}
\includegraphics[width=.99\linewidth]{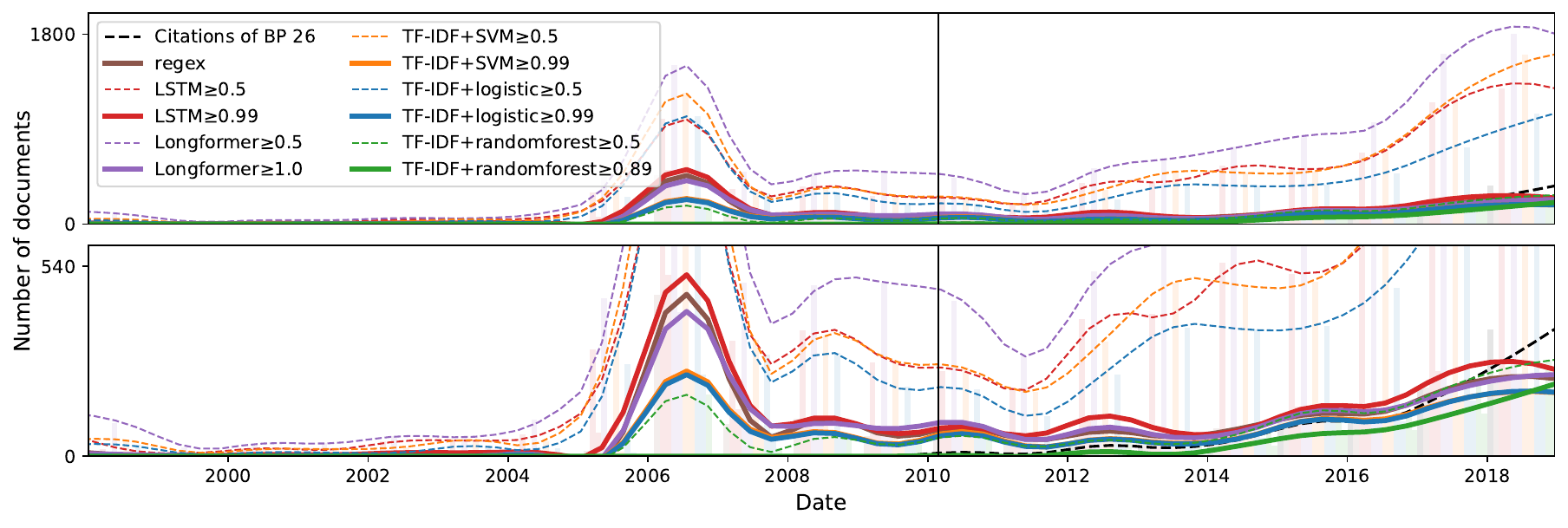}
\caption{Number of documents predicted by each model for BP~26 in Dataset~\#2, represented as a histogram (window length of one year) and interpolated via quadratic spline. 
We give the predictions for the thresholds adapted to Dataset~\#1 (dashed) and Dataset~\#2 (solid), as given in \Cref{tab:adjusted_probabilities}.
Two views are given, the second one zooming in on the ordinate axis.}
\label{fig:predictions_allmodels_SV26}
\end{figure}

\Cref{fig:predictions_allmodels_SV26} shows the predictions of the five models, as well as a regex search for the words \textit{exame criminológico} and \textit{progressão de regime} (criminological examination, regime progression).
Except for random forest, they all present a peak in 2006-2008. For all models, one also notes an increasing tendency starting from 2014.
Both observations correspond to legal events specific to BP~26 and will be discussed in \Cref{subsec:BP_26}.

As far as the 2006-2008 peak is concerned, the regex search finds 659 documents in this period.
A manual inspection of these documents shows that they are indeed related to the theme of BP~26, hence are considered correct predictions of the models.

To explain why TF-IDF+random forest did not detect this peak, let us consider two of its most important features (not presented in \Cref{tab:importance_features}): the words \textit{realização} (realization) and \textit{subjetivos} (subjective) are present in respectively 86.6\% and 78.1\% of its 529 predicted documents (between 2012-2019).
In opposition, they are only present in 60.4\% and 60.8\% of the 659 documents detected by regex between 2006-2008.
This suggests that this TF-IDF model has learned unnecessary restrictions, imposing the presence of words \textit{realização} and \textit{subjetivos}, resulting in too few predicted documents.
In our legal analysis, we will not consider the predictions of TF-IDF+random forest, the other models being deemed to contain more valuable information.

\subsubsection{Binding Precedent 37}

\begin{figure}[!ht]\centering
\footnotesize{Models' predictions for BP~37}
\includegraphics[width=.99\linewidth]{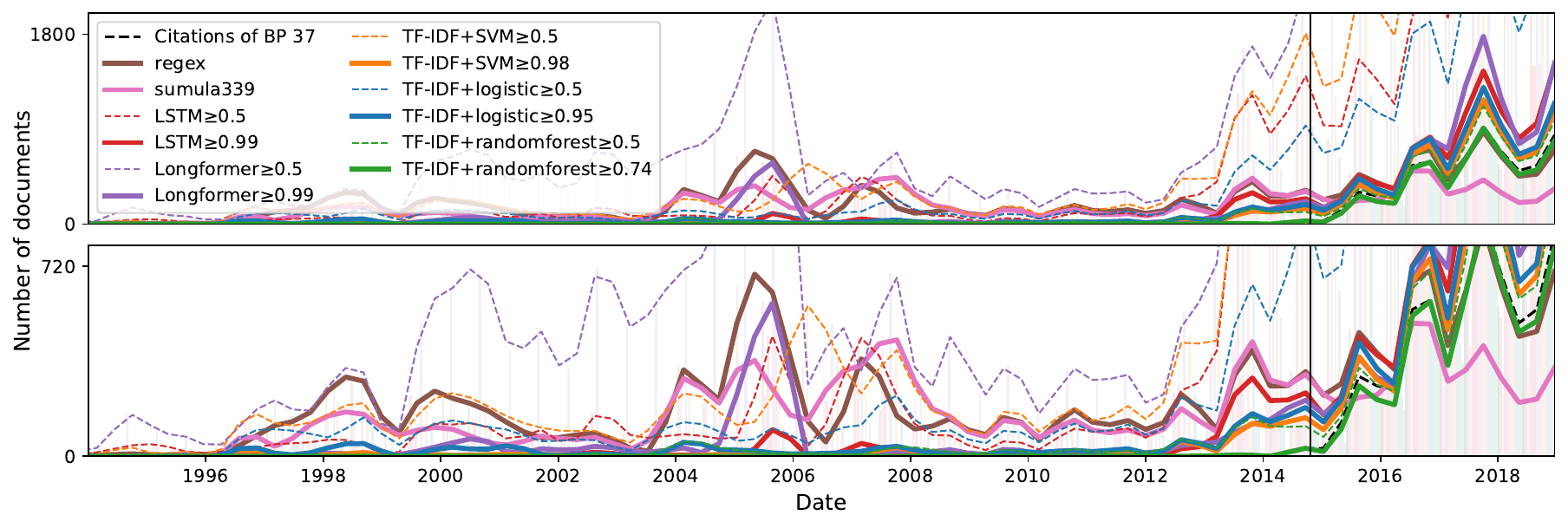}
\caption{Number of documents predicted by each model for BP~37 in Dataset~\#2, represented as a histogram and interpolated via quadratic spline. 
We give the predictions for the thresholds adapted to Dataset~\#1 (dashed) and Dataset~\#2 (solid), as given in \Cref{tab:adjusted_probabilities}.
Two views are given, the second one zooming in on the ordinate axis.
In addition, we represent a regex query for \textit{Súmula 339}.
Unlike the other BPs, the bins in this figure are six months long, to observe the peaks more precisely.}
\label{fig:predictions_allmodels_SV37}
\end{figure}

We show in \Cref{fig:predictions_allmodels_SV37} the predictions of the five models, together with a regex search for the words \textit{isonomia}, \textit{vencimentos} and \textit{servidores públicos}. 
We also add a search for \textit{Súmula 339} (Precedent~339), which, as we will see in \Cref{subsec:BP_37}, is responsible for the creation of BP~37, and can therefore be used as a groundtruth before the precedent's publication date.
As demonstrated in the bottom part of the figure, TF-IDF+random forest presented the best results, along with regex, since they closely follow the groundtruth curve after the BP's publication.

\begin{figure}[!ht]\centering
\footnotesize{Additional metadata for documents predicted by Longformer for BP~37}
\includegraphics[width=.99\linewidth]{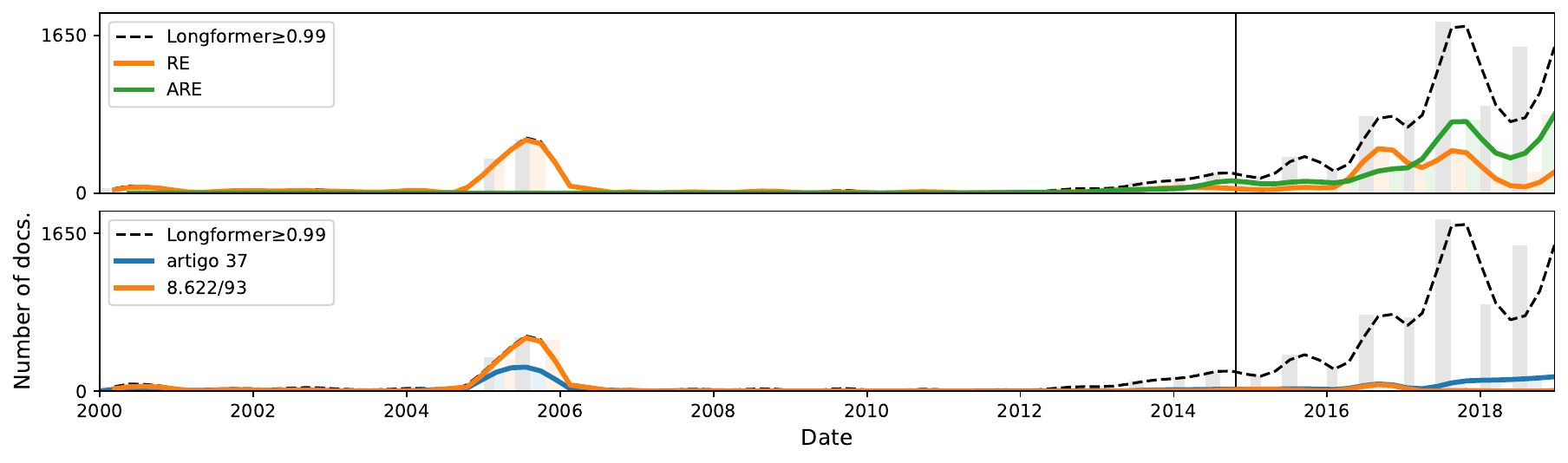}
\caption{\textbf{Top:} Type of legal process used in the documents predicted by the model Longformer for BP~37, among RE (\textit{Recurso Extraordinário}), Rcl (\textit{Reclamação}), and ARE (\textit{Recurso Extraordinário com Agravo}).
\textbf{Bottom:} Documents containing the words \textit{artigo 37} (Article 37) or \textit{8.622/93} (the law).
Bins are six months long.}
\label{fig:predictions_SV37_Longformer}
\end{figure}

One observes a few common trends in the figure.
Before the publication of the BP, the model Longformer, as well as regex, show a clear peak in 2005.
These documents indeed contain words that characterize BP~37.
More specifically, a manual inspection shows that the documents involve a request for salary readjustment, requested by military personnel, based on Law 8.622/1993\footnote{Law 8.622/1993 \url{https://www.planalto.gov.br/ccivil_03/LEIS/L8622.htm}}, which the STF addressed on the grounds of Article 37, Section X of the Constitution.
This observation is confirmed by \Cref{fig:predictions_SV37_Longformer}, which shows the presence of the words 8.622/93 and \textit{Artigo 37} (Article 37) in Longformer's predictions, found by regex.
This event is part of the history of Precedent~339, studied in the dedicated juridical section (see the discussion of \Cref{fig:citations_339_type}).

Besides, after the publication of BP~37, three peaks in the number of predictions can be observed (in late 2015, 2016, and 2017), reflected by all the models.
As it turns out, they can be mapped to specific juridical events, as studied further (see \Cref{fig:citations_SV37_type}).

Last, to understand the low number of predictions of the models before 2012 compared to the documents citing Precedent~339, one can refer to the important words of TF-IDF, collected in \Cref{tab:importance_features}. 
In particular, \textit{isonomia} has the greatest importance, hence we expect that a large portion of the predicted documents contain this word.
It is indeed the case, at a rate of 77.2\%, 76.3\%, and 89.6\% for SVM, logistic regression, and random forest.
Comparatively, only 65.2\% of the documents citing Precedent~339 mention this word.
This suggests that the models are biased towards estimating principally documents employing \textit{isonomia}, which is not representative of all the documents similar to BP~37.
A similar observation can be made for the word \textit{aumentar}, explaining the small number of documents predicted.

\section{Legal analysis}\label{sec:legal_analysis}

For the last step of our program, we will use the models, trained on Dataset~\#1 in \Cref{sec:methods}, and whose predictions on Dataset~\#2 have been validated in \Cref{sec:generalization}, to help uncover the juridical mechanisms behind the observed increase in cases citing a binding precedent (visualized in \Cref{fig:CitationCurve_AfterPublication_allSVs}).
We shall start with a general description of our methodology for empirical evaluation of BP (\Cref{subsec:methodology}), then give a detailed analysis of each BP (\Cref{subsec:BP_11,subsec:BP_14,subsec:BP_17,subsec:BP_26,subsec:BP_37}), and finally draw juridical conclusions (\Cref{subsec:legal_conclusions}).

\subsection{Description of our methodology}\label{subsec:methodology}

\subsubsection{Time series of similar cases}
To evaluate the effect of a binding precedent on jurisprudence, we argue that one should not perform a mere reading of the cases citing that precedent, but also compare them to similar cases published before the precedent.
As already presented in \Cref{subsec:evaluating_generalization}, this information can be visualized through the curve which associates, to each timestamp, the number of similar documents.
In our context, we expect such a curve to behave as follows: the existence of a peak of documents just before the publication of the BP, followed by a steady decrease, until reaching a stable state, hopefully consisting of few documents. 
This behavior would indicate that the BP served its main purpose, which is to ``settle'' an increasingly contentious legal issue, clarifying for interested parties and the lower courts what the final position of the STF is, thus reducing the likelihood of future conflicts involving the same issue.

If this behavior is not observed, as is the case for the five precedents studied in this article, then one can add metadata information to the curve, to reveal the reasons for this deviant phenomenon.
Namely, we will consider the type of process (e.g., habeas corpus or appeal), the state of provenance, the decision of the court (accepted or rejected), the category of juridical reasoning (based on the merits of the case or due to procedural flaws), and the presence of certain words (specific to the legal subject considered).

We point out that to obtain a thorough analysis of the impact of a BP, it would have been appropriate to detect similar documents issued not only by the STF but by lower courts as well.
This analysis, however, is beyond the scope of this article, given the current unavailability of such a dataset.

\subsubsection{Common hypotheses}
A variety of legal, administrative, and political factors may explain the increase in the number of cases concerning a given subject. 
We list below some of these hypotheses, which will be embodied by the binding precedents in the next sections.
We will conclude on the adequacy of these hypotheses in \Cref{subsec:legal_conclusions}.

\begin{enumerate}[style=multiline,leftmargin=1.9cm,labelwidth=3.5cm,itemsep=0.5em,topsep=0.5em]
\item[New avenue to STF or procedural issues] 
The adoption of a BP creates a new gateway for parties to bring their cases to the STF. Since the existing legislation restricts access to the STF, these cases might otherwise have received their final decision from a lower court. Conversely, losing litigants have an incentive to take advantage of a new opportunity to reach the STF in an attempt to change the lower court's decision. This also creates procedural disputes regarding the proper use of specific classes of appeals and what cases fall under the new BP.
\item[Resistance by a group of litigants or regional specificity] 
A group of recurrent litigants or judges deviates from the pacified interpretation of the BP.  Because of the current organization of the Brazilian Judiciary, it is not uncommon for regions and states to develop diverging positions on specific legal issues. When a new BP conflicts with such a position, many cases that reach the STF might stem from this disagreement, following an attempt from specific groups to influence the interpretation of the BP.
\item[Vague wording] 
An element of the precedent is open to interpretation, leading to cases that seek to clarify its concrete meaning. 
Vagueness might create, for instance, confusion related to which cases fall under the BP and which do not. It may also create disputes over how the content of the BP should be interpreted and applied in a specific case. In all these instances, many new cases might reach the STF as a byproduct of legal uncertainty.
\item[New theme] 
The BP introduced a new theme, resulting in a variety of cases discussing this previously unmentioned topic in the STF. 
It should be noted that, in every legal system, judicial decisions entail a creative element, particularly heightened when a court possesses the authority to establish general norms through precedents. Thus, a BP might also be wielded as an instrument of judicial activism, whenever the STF encounters a situation where current social practices are deemed contradictory to the Constitution.
\item[External conjecture not directly related to the precedent] 
A new development unrelated to the issuing of the BP has brought a specific legal issue into the spotlight, causing, as a side-effect, a significant number of decisions to refer to the precedent. 
In this context, the growth in the number of citations should not be seen as a direct effect of the precedent, but rather as a natural trend in the law, where subjects of interest are dynamic and dependent on political, economic, and social conjectures.
\end{enumerate}

\subsection{BP~11}\label{subsec:BP_11}
\subsubsection{Juridical context}
During the trial of HC (\textit{Habeas Corpus}) 91.952\footnote{HC 91.952 \url{https://redir.stf.jus.br/paginadorpub/paginador.jsp?docTP=AC&docID=570157}} (08/07/2008), the defense requested the removal of handcuffs from the defendant, due to concerns about the negative perception that the sight of handcuffs could convey to the jury. 
This case triggered a highly publicized debate, which then shifted to discussing the use of handcuffs for public exposure, and more generally the sensationalization of criminal prosecution in the media.
Based on the presumption of innocence, individual freedom, and the dignity of the human person, the Supreme Court voted a few days later on the following:
\vspace{.33cm}

\begin{flushright}
\begin{minipage}{.9\linewidth}
\textbf{Binding Precedent 11.}
“The use of handcuffs is only permitted in cases of resistance and a well-founded fear of escape or danger to the physical integrity of the prisoner or others, justified in writing, under penalty of disciplinary, civil, and criminal liability of the agent or authority and nullity of the arrest or the procedural act to which it refers, without prejudice to the civil liability of the State.” (STF, 08/2008)
\end{minipage}
\end{flushright}
\vspace{.33cm}

\begin{flushright}
\begin{minipage}{.9\linewidth}
\textbf{Súmula Vinculante 11.}
“Só é lícito o uso de algemas em casos de resistência e de fundado receio de fuga ou de perigo à integridade física própria ou alheia, por parte do preso ou de terceiros, justificada a excepcionalidade por escrito, sob pena de responsabilidade disciplinar, civil e penal do agente ou da autoridade e de nulidade da prisão ou do ato processual a que se refere, sem prejuízo da responsabilidade civil do Estado.” (STF, 08/2008)
\end{minipage}
\end{flushright}
\vspace{.33cm}

This precedent asserts that the use of handcuffs is allowed only when explicitly justified. 
In particular, their use is prohibited in the context of trials or media exposure, which was deemed as humiliating by part of the public discourse.
It has been considered a notable accomplishment in advancing the principles of the Democratic Rule of Law over those of a Police State \citep{dainfluencia,sarlet2016sumula}.

It is worth mentioning certain controversies surrounding the precedent.
First, the nature of the nullity (absolute or relative) is not explicitly specified.
Currently, the Supreme Court understands that the nature of the nullity is relative, as it depends on a demonstration of concrete harm to the defendant. However, \cite{sarlet2016sumula} note that lower courts across the country have been deciding similar cases under the assumption of absolute nullity.
Besides, as pointed out by \cite{sganzerla2012controle} and \cite{da2014analise}, a question remains regarding the exact modalities of \textit{justification} of the ``fear of escape or danger'' evoked in the text.

As already visualized in \Cref{fig:CitationCurve_AfterPublication_allSVs}, the number of documents citing BP~11 has been rising steadily since 2008, with a notable jump in 2016.
We aim, through our analysis, to uncover some of the reasons for this increase.

\subsubsection{Discussion} 
In order to analyze BP~11 quantitatively, we return to the prediction curves, already analyzed in \Cref{subsec:manual_evaluation} (see \Cref{fig:predictions_allmodels_SV11}). 
As we have seen, except regex after 2016 and LSTM, all the models faithfully follow the citation curve of the precedent after its publication, which means they are reliable for the retrieval of documents similar to those citing the binding precedent.
The case of BP~11, however, is unfortunate: no document is predicted by the models before its creation, hence no interesting behavior can be observed in \Cref{fig:predictions_allmodels_SV11}.
This restricts the application of our methodology described in \Cref{subsec:methodology}.
Nevertheless, two juridical insights can be drawn.
First, the regex search shows that the term ``handcuffs'' has come to be used in STF's cases recently, thanks to the edition of BP~11.
It should be noted that a regex search for \textit{uso de força} (use of force), a term commonly used as a synonym of handcuffs, only finds 121 documents, 87.7\% of which also cite the precedent.
On top of that, the fact that the TF-IDF models and Longformer did not detect any similar documents before 2008 suggests that BP~11 not only changed the way cases are formulated but really introduced this matter in the Supreme Court.
As a matter of fact, when reduced to branches “Criminal Procedure Law” and “Criminal Law”, Dataset~\#2 contains 20.3\% of its documents published before 2008, a non-negligible proportion.
This observation falls under the \textit{New theme} hypothesis, evoked in \Cref{subsec:methodology} as a way of explaining the inefficiency of the precedent in reducing the number of cases.

We see in this precedent that a BP does not solely serve to pacify jurisprudence, but it might also be used as a tool of judicial activism by the STF, to resolve a situation deemed contrary to the Constitution.
Among these situations is the widespread use of handcuffs, addressed in the precedent, which, despite being widely accepted by the public, the Justices deemed to be incorrect and to violate the basic rights of the defendants. In this case, BP~11 did not originate from a recurrent legal controversy, but from the realization, by the court, that common social practices were in sharp contradiction with the Law.

\begin{figure}[!ht]\centering
\footnotesize{Additional metadata for documents predicted by TF-IDF+SVM for BP~11}
\includegraphics[width=.99\linewidth]{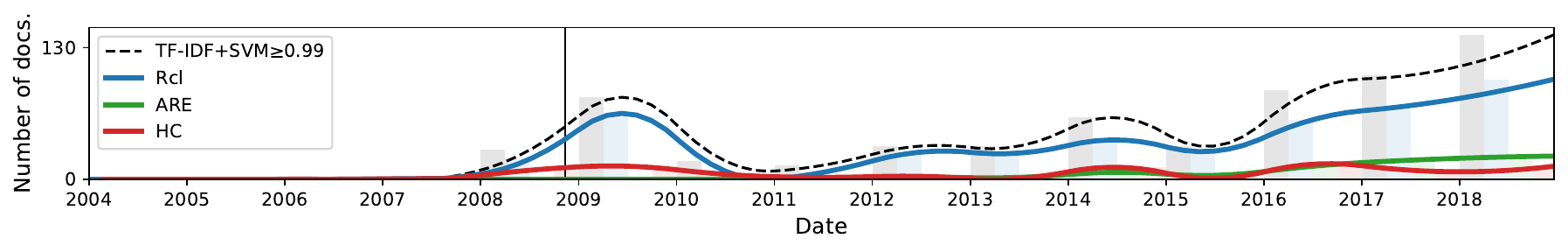}
\caption{Type of legal process used in the documents predicted by TF-IDF+SVM for BP~11 in Dataset~\#2, among Rcl (\textit{Reclamação}), ARE (\textit{Recurso Extraordinário com Agravo}), and HC (\textit{Habeas Corpus}).}\label{fig:citations_SV11_type}
\end{figure}

To provide a more detailed picture of BP~11's applications, and following our methodology, we represent in \Cref{fig:citations_SV11_type} the type of juridical processes used in conjunction with the precedent.
As observed, most of these cases are \textit{reclamações} (complaints/appeals, abbreviated Rcl), a specific process filed to contest a decision or address a grievance, typically related to matters falling within the jurisdiction of the STF.
To understand the purpose of these claims, we read some of the 830 documents citing BP~11.
It turns out that many decisions of the STF are merely ``procedural'', in the sense that they discuss the unsuitability of the \textit{reclamação} as the proper avenue for the case. 
All these cases are deferred, based on internal procedural regulation.
This seems to indicate that the \textit{reclamação} is used by lawyers as a substitute for an appeal, corresponding to our hypothesis \textit{New avenue to the court and procedural issues}.

Besides, we point out that, among the \textit{reclamações} that actually address the question of the use of handcuffs --- i.e., non-procedural cases --- the majority found their use adequately justified. However, these decisions often refrain from delving into a detailed discussion of the justification.
This issue is reinforced by the legal precedent of \textit{preclusão} (preclusion), which states that if there were no objections at the time of handcuff use, and if an authority figure was present, then the \textit{reclamação} cannot be used.
The often sparse written justification of handcuffs use, as well as the application of \textit{preclusão}, call for a clarification of BP~11's statement, as already mentioned in legal literature \citep{sganzerla2012controle,da2014analise}.

\subsection{BP~14}\label{subsec:BP_14}

\subsubsection{Juridical context}
Binding Precedent 14 derived directly from the Attorney Statute of 1994\footnote{Attorney Stat., Article XIV \url{https://www.oab.org.br/Content/pdf/LegislacaoOab/Lei-8906-94-site.pdf}}, which predicts the defense attorney's rights ``to examine, in any institution responsible for conducting investigations, even without authorization, records of flagrant crimes and investigations of any kind, completed or ongoing, even if already concluded for the authority, being able to copy documents and take notes, in physical or digital format''.

This right, however, is not free from dispute, especially in situations where classified investigations take place. The \textit{Habeas Corpus} 88.190\footnote{HC 88.190 \url{https://redir.stf.jus.br/paginadorpub/paginador.jsp?docTP=AC&docID=382091}} exemplifies these further complications: in 2005, the Brazilian newspaper \textit{O Globo} brought light to the existence of an investigation aiming at one of the owners of an important exportation company, a fact that was also new to the investigation's target, given its classified status. After consistent refusals of investigation disclosure,  lawyers decided to file a \textit{Habeas Corpus} to the Federal Court of the Second Region (\textit{Tribunal Regional Federal da 2ª Região}, TRF-2) against the Prosecutor's Office, asking for access to the investigation proceedings. After failing to obtain a favorable decision at TRF-2 and the Supreme Court of Justice\footnote{Different from the STF, this institution is the highest court of appeal in cases where no direct constitutional rights seem to be trespassed.} (\textit{Superior Tribunal de Justiça}), they filed another \textit{Habeas Corpus} to the STF pleading the recognition of unconstitutionality of the denial of access to the ongoing investigation documents, based on the guarantees of the right of sustaining contradictory positions through the whole criminal litigation.
After debates, the STF Justices decided to allow the responsible attorneys to access the classified materials regarding their clients.

Cases similar to the \textit{Habeas Corpus} described in the previous paragraph convinced Justices to approve Binding Precedent 14 in March 2009, which states the following:
\vspace{.33cm}

\begin{flushright}
\begin{minipage}{.9\linewidth}
\textbf{Binding Precedent 14.}
``It is the right of the defender, in the interest of the represented party, to have broad access to the evidence elements that are already documented in an investigative procedure conducted by an authority with jurisdiction in criminal investigation, and that pertain to the exercise of the right to defense."
(STF, 09/2009)
\end{minipage}
\end{flushright}
\vspace{.33cm}

\begin{flushright}
\begin{minipage}{.9\linewidth}
\textbf{Súmula Vinculante 14.}
“É direito do defensor, no interesse do representado, ter acesso amplo aos elementos de prova que, já documentados em procedimento investigatório realizado por órgão com competência de polícia judiciária, digam respeito ao exercício do direito de defesa.” (STF, 09/2009)
\end{minipage}
\end{flushright}
\vspace{.33cm}

The time series at the top of \Cref{fig:CitationCurve_AfterPublication_allSVs} indicates that, for the first few years after its approval, BP~14 was effective in reducing STF cases regarding its topic, although this is followed by a steep increase in references to the object.
This calls for a throughout study of the reasons behind this trend.

\subsubsection{Discussion}

Let us turn to the predictions curves for BP~14, already presented in \Cref{fig:predictions_allmodels_SV14}.
First, we shall analyze the predictions of the models TF-IDF+SVM, TF-IDF+logistic, LSTM, and Longformer, which, although not following the groundtruth curve, evoke a political event worth pointing out.
A reading of the documents shows that their common peak, visualized in \Cref{fig:predictions_randomforest_SV14_type}, can be dated to April 4th, 2017, the day which Justice Edson Fachin decided on 255 cases to proceed with investigations of criminal cases of corruption against holders of federal political positions, such as deputies and senators, based on denunciation of participation in the \textit{Lava Jato} scheme. 
These decisions were pushed by the public prosecutors' office's accusation, based on information retrieved from leniency deals involving those others investigated through a special criminal institution known as \textit{delação premiada} (turns state's evidence). 
However, in the law establishing this form of testimony (12.850/2013\footnote{Law 2.850/2013 \url{https://www.planalto.gov.br/ccivil_03/_ato2011-2014/2013/lei/l12850.htm}}), secrecy is imposed on the shared information and bargain terms, aiming to preserve both the collaborator's and the investigation's integrity. The cases of April 4th, on the other hand, had important public and media impact, together with the many distinct requests from the Public Prosecutors' Office involving the establishment and continuation of investigation on individuals, petitions for publicity of the content of delations, and the list of those investigated. Not only did Justice Fachin decide to make the information public, but justified so, in all 255 documents, using very similar language. It is likely that such a ``copy-and-paste" extract of text, in which the decider argues that the publicity of the investigation information does not detriment ``the defense's right, after collecting the accusatory piece, and with the means and
financial resources to the adversary, the possibility of appealing against a complaint" --- which \textit{is} the topic of BP~14, although the decision, as a whole, is only marginally related to this issue --- caused the models to detect some of these instances as potential applications of the BP. 

\begin{figure}[!ht]\centering
\footnotesize{Additional metadata for documents predicted by TF-IDF+random forest for BP~14}
\includegraphics[width=.99\linewidth]{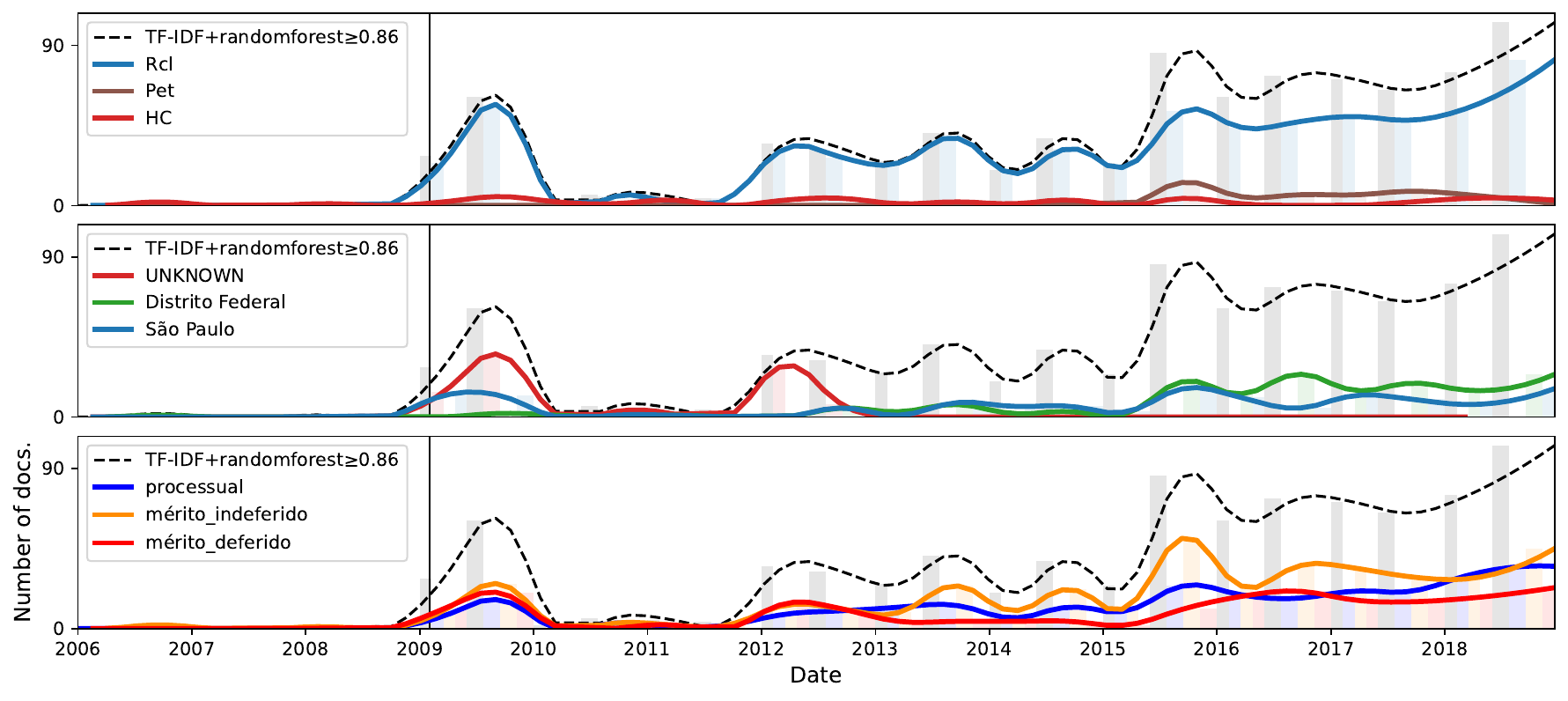}
\caption{\textbf{Top:} Type of legal process used in the documents predicted by TF-IDF+random forest for BP~14, among Rcl (\textit{Reclamação}), Pet (\textit{Petição}), and HC (\textit{Habeas Corpus}).
\textbf{Middle:} State to which these cases belong, with label ``UNKNOWN'' when the metadata is unavailable.
\textbf{Bottom:} Type of decision taken at trial: \textit{processual} if deferred for procedural matters, or \textit{deferido} (accepted) and \textit{indeferido} (rejected). 
In all three graphs, the bins have a length of six months.}
\label{fig:citations_SV14_type}
\end{figure}

We now turn to the predictions of regex and TF-IDF+random forest, which can be considered accurate, given their agreement with the groundtruth curve in \Cref{fig:predictions_allmodels_SV14}.
We gather additional metadata in \Cref{fig:citations_SV14_type}, showing that BP~14 was not enough to settle all debates about lawyers' access to investigation material, as already predicted right after its publication by \cite{sanches2009reflexos}. 
By manually inspecting the decisions citing BP~14, we identified several complaints in which the defense of individuals, although not directly investigated, requested access to classified investigation material, due to some potential relation of their clients with the investigated individuals. Although unaddressed by the statement of the BP, the occurrence of this sort of complaint is frequent enough that the STF itself considered it as a subtopic of the binding precedent\footnote{BP~14, see \textit{Jurisprudência selecionada} (Selected case law) \url{https://portal.stf.jus.br/jurisprudencia/sumariosumulas.asp?base=26&sumula=1230}.}, with decisions going in both directions.
The fact that BP~14 is cited in many cases not directly related to its content fits the hypothesis \textit{New avenue to STF or procedural issues}.

In addition, and as mentioned above, it seems reasonable to associate the increasing trend of uses of BP~14 with the \textit{lavajatism} process of the end of the 2010s, in line with our hypothesis \textit{External conjecture not directly related to the precedent}, defined in \Cref{subsec:methodology}.
As a matter of fact, most complaints concern the investigation of corruption cases aimed at politicians and individuals in power.
In this case, not only the access to the records, or the accords of \textit{delação premiada}, is the source of disagreement, but also the fact that the very status of the investigated individuals poses a threat to the security of informants and the process of evidence gathering\footnote{These recurrent situations were also perceived by the Tribunal, as again indicated in \url{https://portal.stf.jus.br/jurisprudencia/sumariosumulas.asp?base=26&sumula=1230}.}.
Further corroboration to the hypothesis that the cause of increasing uses of BP~14 is tied to potential white-collar crimes involving politics might be gathered from the comparison of the available time series of citations with data after 2019, the end of the period of analysis of this article, because of the ever-changing policies regarding the combat against corruption in Brazil.

This showcases how the realities of the Judiciary Power may be less predictable than most juridical theories on binding precedents seem to suggest. Although BPs are created to address specific legal problems, their subsequent use, in actuality, might be greatly influenced by new dynamics and unforeseeable trends in Case Law. Thus, as a greater number of BPs are created over time, the complexity of the legal system tends to increase and more opportunities arise for external factors to influence the usage of this legal instrument.

\subsection{BP~17}\label{subsec:BP_17}

\subsubsection{Juridical context}
In Brazil, the mechanism by which the public administration (the Union, states, municipalities, and autarchies) pays debts resulting from final judicial sentences is called the \textit{precatórios} system (court order payment). This is a privileged payment system designed to protect the public treasures, regulated by Article 100 of the Constitution\footnote{Article 100 \url{https://www.planalto.gov.br/ccivil\_03/constituicao/Constituicao.htm\#art100.}}. 
More precisely, \textit{precatórios} are payment requisitions issued by the Judiciary in favor of individuals or legal entities that have won lawsuits against the public administration. These debts can be of an alimentary nature (such as salaries, pensions, retirements, indemnities for death or disability) or non-alimentary (other debts, such as expropriations and contracts).

Article 100 establishes rules for the payment of \textit{precatórios}. Payments must follow the chronological order of presentation, with a distinction between alimentary and non-alimentary debts. Each indebted public entity maintains its own list of \textit{precatórios}. Alimentary \textit{precatórios} take precedence over non-alimentary ones. Additionally, elderly individuals (over 60 years old), those with serious illnesses, and people with disabilities are given priority, up to a stipulated limit.

Another important aspect of this system relates to the functioning of public budgets in Brazil. Public entities must forecast the necessary amounts for the payment of \textit{precatórios} in their annual budgets, under penalty of federal, state, or municipal intervention, as the case may be. Thus, Paragraph 5 of Article 100 requires public entities to include in their annual budgets the necessary amounts for the payment of judicial \textit{precatórios} presented by April 2 of each year\footnote{This provision was originally located in Paragraph 1 of Article 100 and the limit date for presenting \textit{precatórios} was July 1st. These changes were introduced by Constitutional Amendments 62 and 114, as is further discussed ahead.}. These \textit{precatórios} must be paid by the end of the following fiscal year. This rule ensures that debts arising from final judicial sentences are properly budgeted for and settled within a reasonable timeframe, preserving the real value owed to the creditor.

Both issues --- that of the order of payments and that of the processing of payments through the budget --- become relevant because the \textit{precatórios} system faces several challenges, such as delays in payments, especially in states and municipalities with a large volume of debts, and the constant need for constitutional amendments and adaptations to address default.
The Binding Precedent 17 was created in November 2009 to clarify in which circumstances the Brazilian Public Treasuries need to pay late payment interests (\textit{juros de mora}). Specifically, the issue was whether this type of interest should apply to the period designated in the current Paragraph 5 (Paragraph 1, at the time of the BP's creation) of Article 100 of the Constitution. It was debated whether late payment interest would apply to the period between the deadline for presenting the \textit{precatórios} and the actual date on which the payment is finally concluded by the Public Treasury. The BP states that, in this period, late payments are not applicable. 
\vspace{.33cm}

\begin{flushright}
\begin{minipage}{.85\linewidth}
    \textbf{Binding Precedent 17.}
        ``No late payment interest is charged on court orders paid during the period stipulated in Paragraph 1 of Article 100 of the Constitution'' (STF, 11/2009)
    \end{minipage}
\end{flushright}
\vspace{.33cm}

\begin{flushright}
    \begin{minipage}{.85\linewidth}
        \textbf{Súmula Vinculante 17.}
        ``Durante o período previsto no parágrafo 1º do artigo 100 da Constituição, não incidem juros de mora sobre os precatórios que nele sejam pagos.'' (STF, 11/2009)
    \end{minipage}
\end{flushright}
\vspace{.33cm}

The legal interpretation adopted by the STF in BP~17 was based on the fact that the Public Treasuries were forced to process the payments through the budget. Thus, according to the Court, the time between the provision of the payment and the actual payment could not be considered a ``delay'' and should not be subjected to late payment interests. 

We draw the reader's attention to the fact that only one month after the creation of the BP, Constitutional Amendment 62 included a Paragraph 12\footnote{Constitutional Amendment 62 (see Art.1º/§12) \url{https://www.planalto.gov.br/ccivil\_03/constituicao/Emendas/Emc/emc62.htm\#art1}} to Article 100, which establishes how late payment interests should be calculated, regardless of the period. Consequently, certain Justices understand this paragraph as overcoming the BP completely\footnote{Overview of the controversies surrounding BP~17 \url{https://www.youtube.com/watch?v=7Rig_iyM8mY}}. However, it does not seem to be the prevalent understanding of the Court, as expressed by Justice Rosa Weber in the \textit{Recurso Extraordinário} (Extraordinary Appeal) RE 577.465/2016\footnote{RE 577.465 \url{https://redir.stf.jus.br/paginadorpub/paginador.jsp?docTP=TP&docID=11795802}}. Many Justices still hold that late payment interests referred to in Paragraph 12 are only applicable if the payment does not occur by the end of the next fiscal year, after the provision, for there would be an actual ``delay'' in the payment, not attributable to normal processing of payments through the budget.

\subsubsection{Discussion}
The prediction curves studied in \Cref{subsec:manual_evaluation} (see \Cref{fig:predictions_allmodels_SV17}) show that the history of BP~17's similar documents should be studied from both the models' predictions --- that may use synonyms instead of exact words from the precedent --- and the regex-detected documents --- that do not suffer from the correlations learned by the models.

\begin{figure}[!ht]\centering
\footnotesize{Additional metadata for documents predicted by Longformer for BP~17}
\includegraphics[width=.99\linewidth]{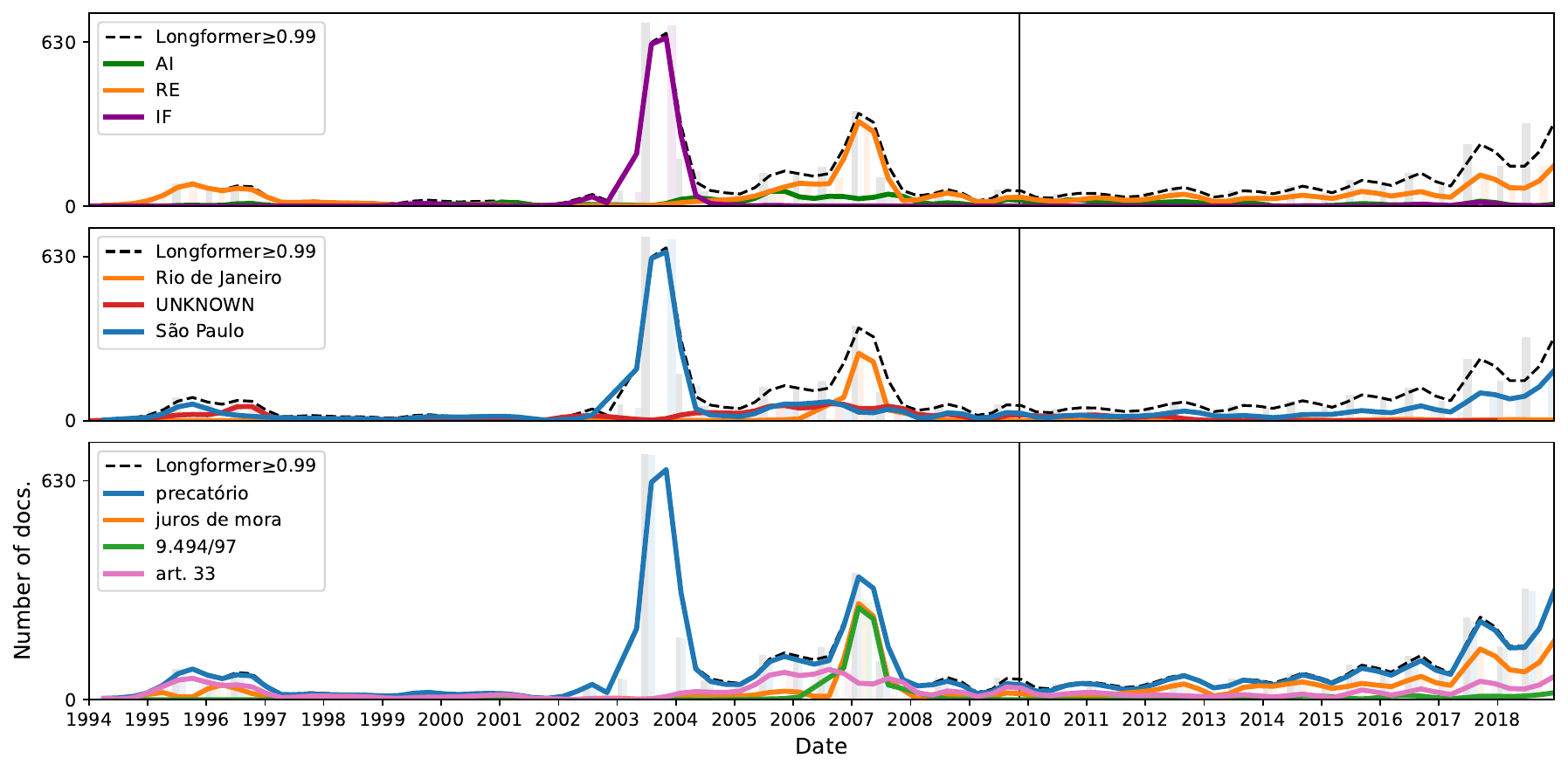}
\caption{\textbf{Top:} Type of legal process used in the documents predicted by the model Longformer for BP~17, among AI (\textit{Agravo de Instrumento}), RE (\textit{Recurso Extraordinário}), and IF (\textit{Intervenção Federal}).
\textbf{Middle:} State to which these cases belong, with label ``UNKNOWN'' when the metadata is unavailable.
\textbf{Bottom:} Presence of the words \textit{precatório} (court order payment), \textit{juros de mora} (late payment interests), \textit{9.494/97} (the law), and \textit{art. 33} (the article).
The bins are six months long.
}
\label{fig:citations_SV17_Longformer_type}
\end{figure}

Starting with the models' prediction, we give in \Cref{fig:citations_SV17_Longformer_type} some additional information regarding Longformer. 
In chronological order, the first peak is visible between 1995-1997.
Manual inspection indicates some abundance of complaints by public entities mostly in cases of land expropriation. For, after the promulgation of the newest Brazilian Constitution, in 1988, it was established by Article 33\footnote{Act of Transitory Constitutional Dispositions (see Art. 33) \url{https://www2.camara.leg.br/legin/fed/conadc/1988/constituicao.adct-1988-5-outubro-1988-322234-normaatualizada-pl.pdf}} of the Act of Transitory Constitutional Dispositions (\textit{Ato das Disposições Constitucionais Transitórias}) that, except for alimentary, \textit{precatórios} could be paid by equal installments through eight years. Because installments were said to be equal, no delay interests were predicted. Judges from a lower court of appeals in the State of São Paulo, however, decided that compensations due to land expropriations were not contemplated by the text of Article 33 and, therefore, late interests should be applicable. This generated a series of appeals to the STF, where judges decided that the exception to the non-applicability of Article 33 was restricted to alimentary \textit{precatórios}, which did not include cases of expropriation (see the search for ``art. 33'' at the bottom of the figure). Article 33's effect naturally died as the period of eight years predicted in its text ended. It is noteworthy that a less significant effect from the same period, but that was also detected by both Longformer, regex, and TF-IDF+SVM and logistic regression, corresponds to appeals, again mostly by public entities, related to the applicability of late payment interests, this time to alimentary \textit{precatórios}. 
However, we could not identify a single event causing this effect, although the appeals come, in their majority, from the States of São Paulo and Paraná.

We now turn to the main peak, between 2002-2005.
As it stands out, it is almost exclusively composed of cases with origin in the State of São Paulo and of type \textit{Intervenção Federal} (Federal Intervention).
This is an exceptional measure of interference by the Union in the States, temporarily suppressing the autonomy of these entities.
This event corresponds to the well-known refusal of São Paulo to pay its debts to alimentary/food creditors\footnote{Report on the request for Federal Intervention in SP \url{https://www.sandovalfilho.com.br/ha-20-anos-pedido-de-intervencao-federal-no-estado-de-sao-paulo-comecou-a-mudar-a-historia-dos-precatorios-judiciais/}}, prioritizing non-alimentary debts instead, in the early 2000s.
A significant change occurred after a pivotal Supreme Court trial deciding on a Federal Intervention request in 2002, led by Attorney Antônio Roberto Sandoval Filho, and brought to trial by Justice Marco Aurélio. 
Though at the time there were questions about the claims brought before the court in this case --- which, in fact, would not be successful ---, this decision marked a turning point, drawing attention to the unsustainable nature of the court-ordered debt situation in Brazil.
Even though we still observe a high value on the curve from 2016 (\Cref{fig:citations_SV17_Longformer_type}), the payment of debts in São Paulo was considered maximum priority by the Court of Justice of the State of São Paulo\footnote{TJSP's \textit{precatórios} campaign \url{https://www.tjsp.jus.br/Imprensa/Campanhas/Precatorios}}, with more than 19 billion reais (4 billion dollars) being released for court-ordered payments in 2023\footnote{TJSP's disclosure \url{https://www.tjsp.jus.br/Noticias/Noticia?codigoNoticia=95996&pagina=1}}.

Regarding the last peak, in 2007-2008, the analyzed documents predominantly address legal disputes involving military personnel and the application of late payment interest in cases of delayed compensation, often resulting in \textit{precatórios}. 
Central to these cases is Article 1º-F of Law 9.494/1997\footnote{Law 9.494/1997, Article 1º-F \url{https://www.planalto.gov.br/ccivil_03/leis/l9494.htm}}, which establishes a reduced annual interest rate of 6\% for public debts, instead of 12\% (see ``9.494/97'' at the bottom of \Cref{fig:citations_SV17_Longformer_type}).
More precisely, the article has been introduced by Provisional Measure 2.180-35\footnote{MPV 2.180-35 \url{https://www.planalto.gov.br/ccivil_03/LEIS/L9494.htm\#art1f.}} in 2001 (later rectified in 2009).
Military personnel challenged this reduced rate, arguing it violates the constitutional principle of isonomy (equality before the law), particularly when compared to civilian public servants or private-sector norms.
While many rulings uphold the lower rate as consistent with fiscal policy, others question its fairness, especially given the unique service conditions of military personnel. These cases highlight a systemic conflict between fiscal governance, the equitable treatment of public servants, and the adequacy of \textit{precatório}-related compensation mechanisms.

\begin{figure}[!ht]\centering
\footnotesize{Additional metadata for documents predicted by regex for BP~17}
\includegraphics[width=.99\linewidth]{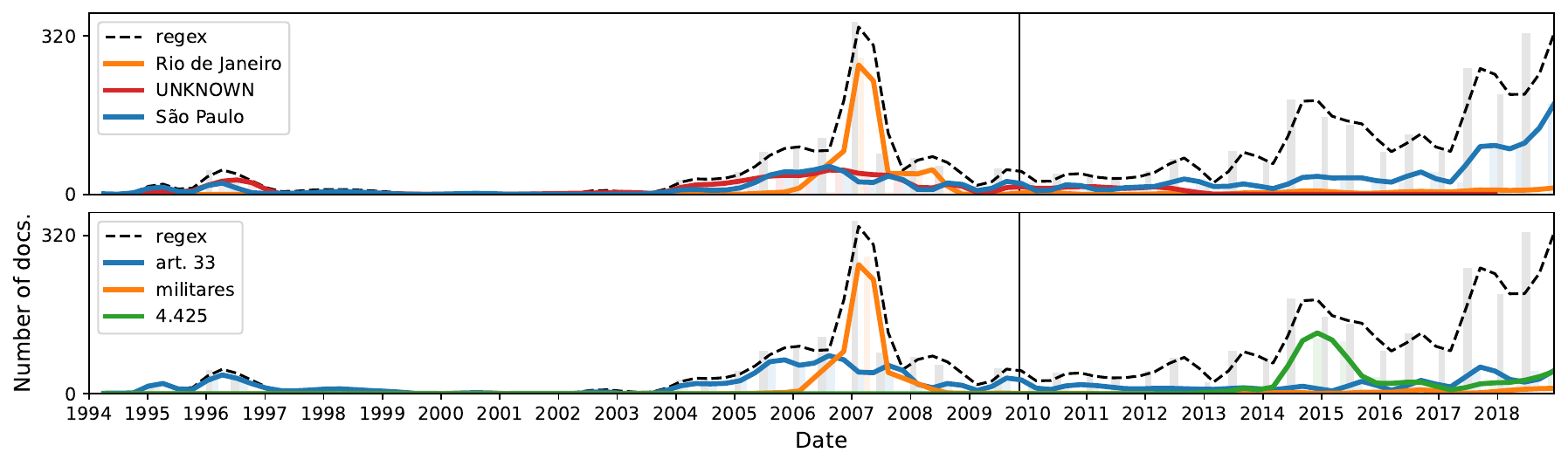}
\caption{
\textbf{Top:} State to which the cases satisfying the regex query belong, with the label ``UNKNOWN'' when the metadata is unavailable.
\textbf{Bottom:} Presence of the words \textit{art.~33} (Article 33), \textit{militares} (military personnel) and 4.357 (the law).
The bins have a length of six months.}
\label{fig:citations_SV17_regex_type}
\end{figure}

We now turn to regex's predictions, with \Cref{fig:citations_SV17_regex_type} representing additional metadata.
The peaks in 1995-1997 and 2007-2008 have already been identified in Longformer's predictions (also visualized in \Cref{fig:citations_SV17_regex_type}), so we skip ahead to 2015.
Between 2014 and 2017, the number of documents satisfying the regex query derive from ADIs 4.357\footnote{ADI 4.357 \url{https://portal.stf.jus.br/processos/detalhe.asp?incidente=3813700}} and 4.425\footnote{ADI 4.425 \url{https://redir.stf.jus.br/paginadorpub/paginador.jsp?docTP=TP&docID=5067184}} (\textit{Ação Direta de Inconstitucionalidade}, Direct Action for Unconstitutionality). This relates to a very specific, technical legal discussion, regarding a transition rule introduced by Constitutional Amendment 62 (already mentioned). The Court found this amendment to be partially unconstitutional, due to some of the rules it introduced for the order of payments --- especially some of the provisions concerning preference due to old age --- and, as a result, there was uncertainty regarding how the interest rate for late payments should be calculated during the transition period.

The data suggests that BP~17 was partially ineffective in reducing the number of cases brought to the STF. The BP was created after the major peak of cases concerning its main subject had already subsided. Soon after its creation, major changes to the underlying constitutional text were introduced by Constitutional Amendments 62 and 114, raising questions if the BP was still standing. Furthermore, by the end of the period of analysis, we may see again a tendency towards an increase in the number of cases concerning the subject of the BP.
The only peak identified by regex after the creation of the BP, in 2015, can be attributed to an external event --- matters related to the effects of the declarations of unconstitutionality in ADIs 4.357 and 4.425. This corresponds to our
hypothesis \textit{External conjecture not directly related to the precedent}. 

But the main lesson that can be drawn from this case seems to be that the dynamic nature of the legal system sometimes will bring changes that render a BP ineffective, or that raise questions about its validity soon after it is issued. It might be argued that, in a Civil Law system, the preferred solution to a problem of legal interpretation or implementation comes from a change in the underlying legal texts --- and not through Case Law or a BP. 
Nevertheless, it is frequently necessary to rely on judicial solutions as they provide a faster solution to urgent problems and address concrete problems more directly. 
In the case of BP~17, however, we see how the dynamics of the legal system, and how changes in the legal text and disputes over related topics can limit the effectiveness of this judicial tool to address legal problems and reduce litigation.

\subsection{BP~26}\label{subsec:BP_26}

\subsubsection{Juridical context}
In December 2009, Binding Precedent 26 was introduced to address two key legal issues, regarding individuals convicted of heinous crimes (\textit{crimes hediondos e equiparados}), a category that encompasses offenses like torture, homicide, drug dealing, and rape.
As specified in the 1990's Law 8.072\footnote{Law 8.072 \url{https://www.planalto.gov.br/ccivil_03/leis/l8072.htm}}, more severe treatment was imposed on these prisoners, in particular prohibiting progress through the three prison regimes, closed, semi-open, and open (Article 2, Paragraph 1). In HC 82.959/SP\footnote{HC 82.959/SP \url{https://redir.stf.jus.br/paginadorpub/paginador.jsp?docTP=ac&docID=79206&pgI=156&pgF=160}}, the court found this provision to be unconstitutional. This sparked two subsequent discussions related to the determination of regime progression that would be answered in BP~26.
Firstly, BP~26 clarified whether the law 11.464/07\footnote{Law 11.464 \url{https://www.planalto.gov.br/ccivil_03/_Ato2007-2010/2007/Lei/L11464.htm}}, allowing sentence progression for heinous crimes, applied retroactively. 
Secondly, it determined whether judges could require a criminological examination (\textit{exame criminológico}) for sentence progression. 
The following wording was established, which would address both issues raised:
\vspace{.33cm}

\begin{flushright}
\begin{minipage}{.9\linewidth}
\textbf{Binding Precedent 26.}
``For the purpose of sentence progression in the case of the imprisonment for a heinous crime or an equivalent offense, the execution judge shall consider the unconstitutionality of art.~2 of Law 8.072, dated July 25, 1990, without prejudice to assessing whether the convicted individual meets the objective and subjective requirements for the benefit, and may, for this purpose, order, with reasoned justification, the performance of a criminological examination."
(STF, 12/2009)
\end{minipage}
\end{flushright}
\vspace{.33cm}

\begin{flushright}
\begin{minipage}{.9\linewidth}
\textbf{Súmula Vinculante 26.}
“Para efeito de progressão de regime no cumprimento de pena por crime hediondo,
ou equiparado, o juízo da execução observará a inconstitucionalidade do art. 2º da Lei 8.072, de 25 de julho de 1990, sem prejuízo de avaliar se o condenado preenche, ou não, os requisitos objetivos e subjetivos do benefício, podendo determinar, para tal fim, de modo fundamentado, a realização de exame criminológico.” (STF, 12/2009)
\end{minipage}
\end{flushright}
\vspace{.33cm}

However, as shown in \Cref{fig:CitationCurve_AfterPublication_allSVs}, the implementation of BP~26 has been followed by certain growth of cases related to it.
As reported by \cite{amaral2016habeas}, these cases are mainly appeals (\textit{reclamação}) aimed to challenge decisions from lower courts that had not fully embraced the new legal theories.
We expect to reveal, via our methodology, additional explanations for this issue.

It is worth mentioning that the debate that led to the voting of BP~26 reveals a high level of criticism among public defenders.
The dispute, especially led by the Public Defender's Office of the State of São Paulo, represented by Dr. Rafael Ramia Muneratti, focused on the second part of the binding precedent, regarding the criminological examination.
It was argued that this examination, extensively rejected by psychologists and other specialists\footnote{The contradiction between the conduct of criminological examination and the Psychologists' Code of Ethics is discussed in \url{https://www.crpsp.org/noticia/view/1846/nota-tecnica-sobre-a-suspensao-da-resolucao-cfp-0122011---atuacao-dao-psicologao-no-ambito-do-sistema-prisional}.}, would contradict the \textit{princípio da individualização da pena} (principle of individualization of punishment).
The following year, this disagreement was explicitly formalized by the Public Defender's Office under the name ``Thesis 53''\footnote{Thesis 53 of the Public Defender's Office of the State of São Paulo  \url{https://www2.defensoria.sp.def.br/dpesp/Conteudos/Materia/MateriaMostra.aspx?idItem=61257&idModulo=9706}.}.

\subsubsection{Discussion}

The trends of the prediction curves, in \Cref{fig:predictions_allmodels_SV26}, can be mapped to exact juridical events.
First, the peak around 02/23/2006 corresponds to the declaration of unconstitutionality in HC 82.959/SP, the case that launched the discussion leading up to BP~26.
The almost immediate reaction of the Brazilian Congress to address the unconstitutionality declared by this judgment, with the presentation of a law project as early as 03/23/2006, would explain the subsequent decline.

\begin{figure}[!ht]\centering
\footnotesize{Additional metadata for documents predicted by Longformer for BP~26}
\includegraphics[width=.99\linewidth]{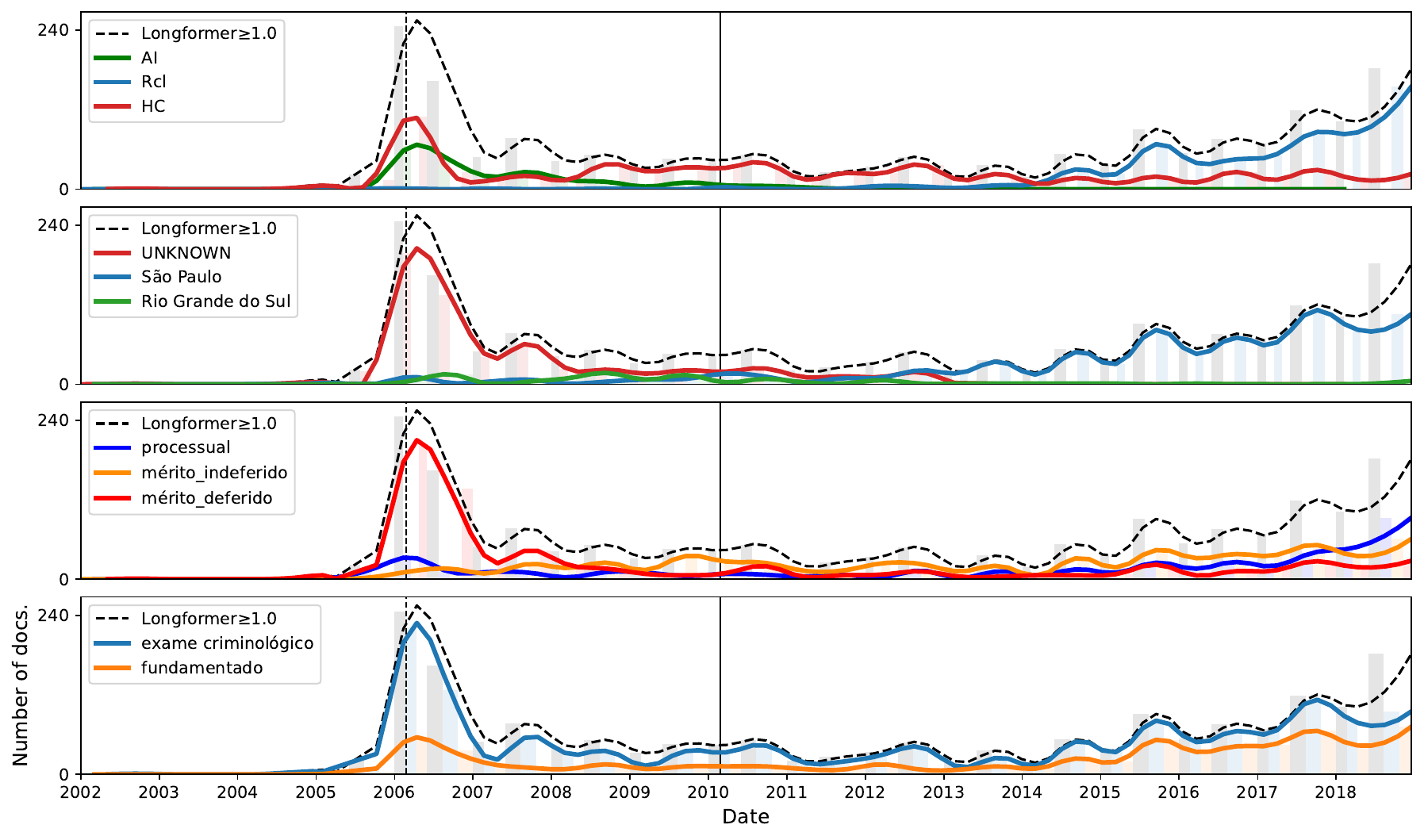}
\caption{\textbf{First:} Type of legal process used in the documents predicted by the model Longformer for BP~26, among AI (\textit{Agravo de Instrumento}), Rcl (\textit{Reclamação}), and HC (\textit{Habeas Corpus}).
\textbf{Second:} State to which these cases belong, with label ``UNKNOWN'' when the metadata is unavailable.
\textbf{Third:} Type of decision taken at trial: \textit{processual} if deferred for procedural matters, or \textit{deferido} (accepted) and \textit{indeferido} (rejected). 
\textbf{Fourth:} Presence of the words \textit{exame criminológico} (criminal examination) and \textit{fundamentado} (grounded/founded).
In all graphs, the bins have a length of six months. 
The solid vertical line represents the date of publication of BP~26, and the dashed one the trial of HC 82.959/SP (declaration of inconstitutionality).}
\label{fig:citations_SV26_type}
\end{figure}

Besides, the trend of increased citations, evident after the second half of 2015, can be understood further with \Cref{fig:citations_SV26_type}: most of the processes are appeals, emitted in the State of São Paulo.
A reading of the texts suggests that from this date, a large number of complaints regarding the rationale behind the \textit{request} for criminological examinations, which, according to the wording of the precedent, should be ordered ``with reasoned justification'' (\textit{de modo fundamentado}).
As evoked earlier, this matter had already been raised by the Public Defender's Office of São Paulo at the time of BP~26's debate, arguing that a precise justification for the request for criminological examination is essential.
It is reasonable to hypothesize that São Paulo's defenders follow this thesis and express themselves through appeals to the STF.
This phenomenon falls under the hypothesis \textit{Resistance by a group of litigants or regional specificity} defined in \Cref{subsec:methodology}.

\Cref{fig:citations_SV26_type} reveals other important information: starting from 2017, the cases tend to be deferred by the Supreme Court based on procedural matters.
More precisely, we have observed that it is common for cases to arrive in front of the judges having already lost their purpose (for instance because the convicted, in the meantime, obtained regime progression through other means).
But more frequently, cases are deferred because the court considered that \textit{reclamação} was not the correct procedural instrument.
Even though the chance of an appeal being accepted might be low, we could reasonably think that defenders continue submitting them, gambling on this chance.
This constitutes a typical case of the \textit{New avenue to the court and procedural issues} hypothesis.

In conclusion, our analysis shows that the expected effect of unifying the jurisprudence surrounding BP~26 --- regime progression for convicted of heinous crimes --- was not entirely achieved. The \textit{possibility} of the progression is taken for granted, but it has simultaneously provoked a new wave of discussions, centered around the use of criminological examination, and more precisely the justification of its request.
In this respect, the vague wording of BP~26 is to blame. 
However, we should note that vague wording is a common trait of legal provisions. Under certain conditions, it may be viewed not as a problem inherent to the production of Law, but as a quality of it. Vagueness in legal text allows for flexibility of interpretation and may be intentionally used when relevant context-specific dimensions of a legal problem are not known \textit{ex ante}. Thus, when a general or vague term is used in Law, aspects of the solution of the problem are, in practice, deferred to be solved at a posterior time, when more information about the concrete situation is available to the parties involved. Vagueness in the legal text is both unavoidable, to a certain extent, and desirable, as precision can generate arbitrary solutions that are contradictory to the main purpose of a norm under specific or unforeseen circumstances. As Endicott \cite[p.~14]{endicott-2011} puts it: ``Far from being repugnant to the idea of making a norm, vagueness is of central importance to law-makers (...) It is a central technique of normative texts.''

When elaborating a binding precedent, the STF is in the unique position of being both the creator of a general normative text and the ultimate interpreter of that same text. This induces legal tensions that our study of BP~26 has brought to light: if the vague wording of the precedent has allowed for more flexibility in the determination of possible uses of criminological examination, it has also caused many cases to be brought to the STF for clarification, rendering one of the main purposes of a BP --- that of reducing legal disputes and legal uncertainty --- ineffective.

\subsection{BP~37}\label{subsec:BP_37}

\subsubsection{Juridical context}
Many legal cases awaiting judgment by the Justices of STF have been recognized as having general repercussions. In August 2014, the Justices ruled on one of these cases (the Extraordinary Appeal RE 592.317\footnote{RE 592.317 \url{https://redir.stf.jus.br/paginadorpub/paginador.jsp?docTP=TP&docID=7181942}}), which would release 1100 ordinary cases for judgment in lower judiciary branches. In this appeal, the STF revisited and applied one of its precedents to overrule a decision from a Brazilian State court. This (non-binding) precedent, \textit{Súmula 339}\footnote{Precedent 339 \url{https://portal.stf.jus.br/jurisprudencia/sumariosumulas.asp?base=30&sumula=1484}} (Precedent~339), establishes that the judiciary cannot increase the salaries of public servants on the grounds of isonomy. 
During the debates on the mentioned appeal, Justice Gilmar Mendes suggested converting the Precedent into a binding precedent, given its consistent application in the STF's decisions. This way, lower judiciary branches would have to rule according to what was established by the Supreme Court, which, hopefully, would reduce the number of cases related to it that reach the STF. In October 2014, BP~37 was created with the same wording as Precedent~339:
\vspace{.33cm}

\begin{flushright}
\begin{minipage}{.9\linewidth}
\textbf{Binding Precedent 37.}
“It is not up to the Judiciary, which has no legislative function, to increase the salaries of public servants on the grounds of isonomy.” (STF, 10/2014)
\end{minipage}
\end{flushright}
\vspace{.33cm}

\begin{flushright}
\begin{minipage}{.9\linewidth}
\textbf{Súmula Vinculante 37.}
“Não cabe ao Poder Judiciário, que não tem função legislativa, aumentar vencimentos de servidores públicos sob o fundamento de isonomia.” (STF, 10/2014)
\end{minipage}
\end{flushright}
\vspace{.33cm}

Long before the creation of this Binding Precedent, a law from 2003 (Law 10.698\footnote{Law 10.698/2003 \url{https://www.planalto.gov.br/ccivil_03/leis/2003/l10.698.htm}}) was enacted to grant an increase of R\$ 59.87 to all federal public servants. This resulted in 30,000 legal cases over 15 years arguing that the adjustment should not be a fixed amount but rather proportional to each servant’s salary\footnote{Report on Law 10.698 \url{https://www.conjur.com.br/2017-out-23/lei-deu-aumento-59-servidores-produziu-30-mil-processos}} (specifically, 13.23\%). Different judges issued different rulings, so legal uncertainty took place. Many cases reached the STF, and the Justices, opposed to the adjustment, ended up citing Precedent~339 and later Binding Precedent 37 in their decisions.
While BP~37 and repeated rulings from the STF demonstrated the unconstitutionality of such an adjustment, the controversy surrounding the issue continued to create some degree of legal uncertainty. As a result, the number of decisions citing this BP throughout time remained high.

\subsubsection{Discussion}
As a consequence of the analysis of the predicted documents in \Cref{subsec:manual_evaluation} (see \Cref{fig:predictions_allmodels_SV37}), the models detect documents of interest pre- and post-publication of BP~37.
In addition, similar documents before its publication can be studied through citations of Precedent~339. 
We will therefore divide our discussion into two parts, first studying the predictions of TF-IDF+random forest. We note that it would have been equivalent, as far as post-publication documents are concerned, to consider the groundtruth curve (actual citations of BP~37), or that of regex.

\begin{figure}[!ht]\centering
\footnotesize{Additional metadata for documents predicted by TF-IDF+random forest for BP~37}
\includegraphics[width=.99\linewidth]{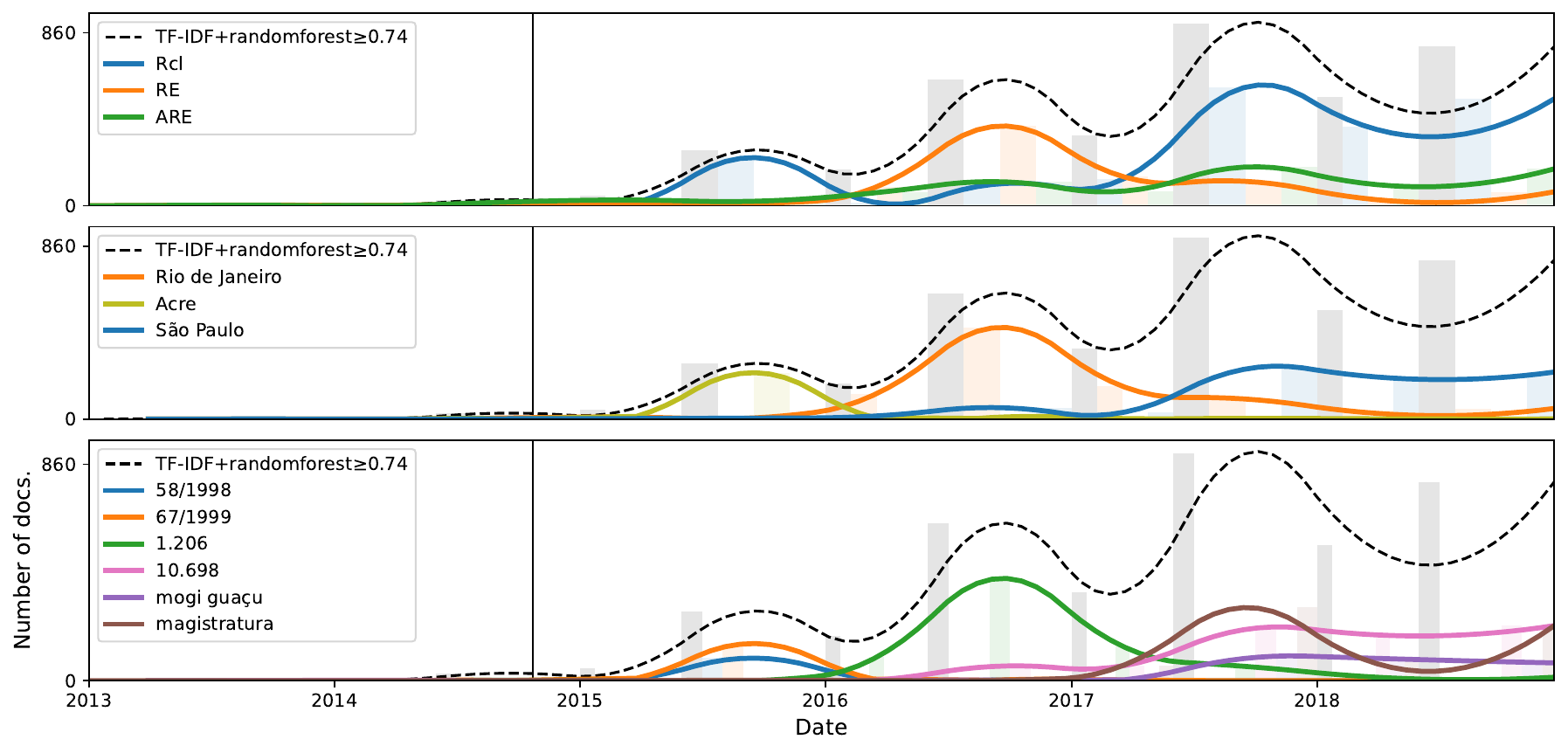}
\caption{\textbf{Top:} Type of legal process used in the documents predicted by the model TF-IDF+random forest for BP~37, among Rcl (\textit{Reclamação}), RE (\textit{Recurso Extraordinário}), and ARE (\textit{Recurso Extraordinário com Agravo}).
\textbf{Middle:} State to which these cases belong, with label ``UNKNOWN'' when the metadata is unavailable.
\textbf{Bottom:} Documents containing the words \textit{58/1998}, \textit{67/1999}, \textit{1.206}, \textit{10.698}, \textit{Mogi Guaçu}, or \textit{magistratura} (Judiciary members).
The bins have a length of six months.}
\label{fig:citations_SV37_type}
\end{figure}

\Cref{fig:citations_SV37_type} gathers relevant metadata regarding TF-IDF+random forest's predictions: the type of process, the state of emission, and the mention of particular words, which we will explain throughout the text. 
The first peak of the curve, in 2015, concerns complaints/appeals (Rcl) originating from the State of Acre (\Cref{fig:citations_SV37_type} (Top, Middle)).
A manual inspection of some of these documents shows the application of BP~37 in the context of the Acre Complementary State Laws 58/1998\footnote{Acre Complementary State Law 58/1998 \url{https://www.al.ac.leg.br/leis/?p=11843}} and 67/1999\footnote{Acre Complementary State Law 67/1999 \url{https://www.al.ac.leg.br/leis/?p=11843}} (see the respective curves in the figure). While temporary teachers from that State are subject to the first law, the latter applies to tenured professionals. 
The main reason these two laws and BP~37 appear together in many documents is that a lower court in Acre granted 45 days of vacation to temporary teachers (the same duration granted to tenured teachers), basing its decision not on the laws but on the principle of isonomy. The State subsequently filed complaints to the STF to overturn the Court’s decision, arguing that BP~37 had not been respected. However, the STF rejected the complaints, stating that the BP pertains to salary increases, which was not relevant in this case.

The second peak of the curve, in 2016, can be related to real-world juridical events as follows. 
As depicted in the figure, most predicted documents in that year were Extraordinary Appeals (RE) that came from the State of Rio de Janeiro (RJ).
A manual inspection of some of these documents shows the application of BP~37 in the context of the RJ State Law 1.206/1987\footnote{RJ State Law 1.206/1987 \url{https://leisestaduais.com.br/rj/lei-ordinaria-n-1206-1987-rio-de-janeiro-dispoe-sobre-o-reajuste-de-vencimentos-e-proventos-do-funcionalismo-estadual-e-da-outras-providencias?origin=instituicao}}.
This is reflected by the citation curve for ``1.206''.
Given that this law was enacted to increase the salaries of public servants from the Executive and Legislative branches, servants from the Judiciary questioned their exclusion by taking legal action. 
After citing BP~37 in several decisions, the STF issued a final ruling dismissing the servants' request in October 2016\footnote{General repercussion thesis ARE 909.437 \url{https://redir.stf.jus.br/paginadorpub/paginador.jsp?docTP=TP&docID=11828219}}.

The third peak of decisions, from 2017 onwards, can be understood as a combination of several initiatives.
The first reason is Law 13.317/2016\footnote{Law 13.317/2016 \url{https://www.planalto.gov.br/ccivil_03/_ato2015-2018/2016/lei/l13317.htm}} which altered the career and salary structure of public servants.
This led to a significant increase in cases claiming a 13.23\% salary adjustment, based on Law 10.698/2003 (already mentioned), as shown by the citation curve in \Cref{fig:citations_SV37_type} (see citations of ``10.698'').
As a response, Justice Gilmar Mendes proposed in 2017 the creation of a new and specific binding precedent that would explicitly prohibit the granting of the mentioned adjustment\footnote{Proposal for a BP by Justice Gilmar Mendes \url{https://www.conjur.com.br/dl/psv-128-reajuste-1323.pdf}}. 
After further analysis of the proposition, which considered the Prosecutor General’s manifestation\footnote{Proposal for BP 128 \url{https://portal.stf.jus.br/processos/downloadPeca.asp?id=312747008&ext=.pdf}} and others, the creation of this new BP was rejected.

In addition, the many mentions of ``Mogi Guaçu'', a municipality in São Paulo, can be attributed to a collective agreement signed between the municipality and workers' syndicate (\textit{Lei Complementar Municipal} 1.121/2011\footnote{Mogi Guaçu Complementary State Law 1.121/2011 \url{https://sistema.camaramogiguacu.sp.gov.br/consultas/norma_juridica/norma_juridica_mostrar_proc?cod_norma=5779}}), determining the incorporation of bonuses into the salaries of all municipal employees and civil servants.
A somewhat similar movement can be observed through the documents citing \textit{magistratura} (members of the judiciary, such as judges), which relates to a controversy concerning the \textit{licença prêmio} (long-service leave), a bonus awarded every five years to public servants (federal, municipal, or state).
Certain members of the judiciary invoked the principle of isonomy to obtain this bonus, which was rejected, in the terms of BP~37\footnote{Statement on general repercussion, Justice Luiz Fux, the 2017-07-26 \url{https://portal.stf.jus.br/jurisprudenciaRepercussao/verPronunciamento.asp?pronunciamento=7188706}}.

In conclusion, the five trends observed on the citations of BP~37 --- vacation for temporary teachers in the State of Acre, salary increases for public servants from Law 1.206/1987 or Law 13.317/2016 in the State of Rio de Janeiro, the Mogi-Guaçu syndicate, and \textit{licença prêmio} for judiciary --- are protests led by specific groups, as formulated by our hypothesis \textit{Resistance by a group of litigants or regional specificity}.

\begin{figure}[!ht]\centering
\footnotesize{Additional metadata for documents citing Precedent~339}
\includegraphics[width=.99\linewidth]{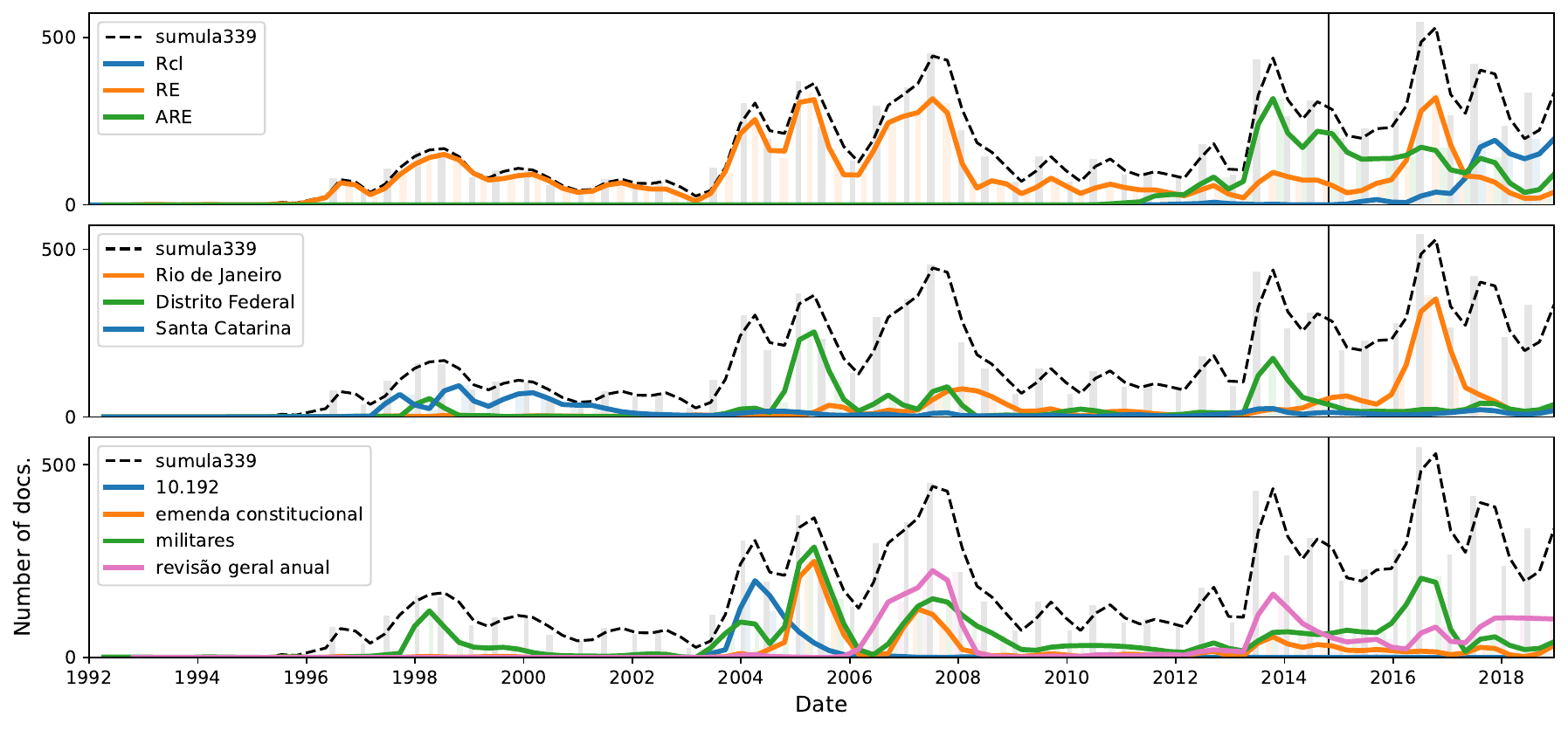}
\caption{
\textbf{Top:} Type of legal process used in the documents citing Precedent~339, among Rcl (\textit{Reclamação}), RE (\textit{Recurso Extraordinário}), and ARE (\textit{Recurso Extraordinário com Agravo}).
\textbf{Middle:} State to which these cases belong.
\textbf{Bottom:} Search for the presence of the words \textit{10.192}, \textit{emenda consitucional} (constitutional amendment), \textit{militares} (military personnel), or \textit{revisão geral anual} (annual general revision).
In all three graphs, the bins are six months long.
}
\label{fig:citations_339_type}
\end{figure}

Next, we turn to the analysis of BP~37's similar documents before its publication, through those that mention Precedent~339.
We remind the reader that the latter is a direct descendant of the former, hence they share similar documents.
\Cref{fig:citations_339_type} contains metadata relative to these documents.

A first trend can be observed between 1997 and 2002: most of the documents come from the State of Santa Catarina, as visualized on the blue curve in the middle graph.
A manual inspection revealed that most documents refer to two general cases that discuss permanent increases in the salaries of state servants of Santa Catarina who have temporarily held commissioned positions (Complementary State Law 43/1992 and State Law 9.847/1995).
Thus, citing Precedent~339 in its arguments, the State of Santa Catarina filed extraordinary appeals to the STF against lower court decisions that guaranteed to the servants (i) the right to the permanent increase, or (ii) equal salary adjustment to those servants receiving such an increase due to past commissioned positions and those who are currently in equivalent positions.

To continue, one observes a significant rise in citations between 2003 and 2010, made of three peaks. 
They are all linked to specific events, represented by the presence of the words \textit{10.192}, \textit{emenda consitucional} (constitutional amendment) or \textit{revisão geral anual} (annual general review).
Regarding the first one, a manual inspection revealed they refer to the application of Precedent~339 in the context of Law 10.192/2001, a law that increased the salaries of non-public servants (10,87\%) and which was used by public servants to require a similar increase.

As far as the 2005 peak is concerned, it must be traced back to 1993.
Indeed, that was the year the Laws 8.622/1993 (already mentioned in \Cref{fig:predictions_SV37_Longformer}) and 8.627/1993 came into effect, adjusting the salaries of higher ranks in the armed forces by 28.86\%. STF Justices then recognized the same right to lower-ranking military personnel, understanding it was a case of general revision of salaries, therefore not applying Precedent~339 but rather Article 37, Section X of the Constitution.
The Union filed extraordinary appeals\footnote{Such as AI 551.744 \url{https://portal.stf.jus.br/processos/detalhe.asp?incidente=2308620}} (RE), and one of the arguments was that the Constitution allowed for different adjustments for different ranks (Article 142), but the Justices upheld their decision based on Article 37. 
Article 37 was amended by Constitutional Amendment 19/1998\footnote{Constitutional Amendment 19/1998 \url{https://www.planalto.gov.br/ccivil_03/constituicao/Emendas/Emc/emc19.htm\#art3}}, which established the necessity of a specific law for any form of salary setting or change, thus exempting the requirement for adjustments to consistently follow the same percentage. In light of this, the Federal Government requested, and the STF accepted, that the 28.86\% adjustment should be valid only until June 1998, when the Constitutional Amendment came into effect.

We now turn to the last peak.
In addition to the need for a specific law for any form of setting or change in the salary of public servants, Constitutional Amendment 19/1998 also established that the general revision of salaries must occur annually and based on annual laws proposed by the Head of the Executive branch (President of the Republic or state governors). A manual inspection of documents between 2006 and 2008 found extraordinary appeals\footnote{Such as RE 524.145 \url{https://portal.stf.jus.br/processos/detalhe.asp?incidente=2461438}} requesting compensation for public servants due to omissions in issuing such a law. The STF denied these appeals based on Precedent~339. After many years of judgments on this matter, the STF stated in 2019 that not submitting a law's proposition regarding the annual revision of public servants' salaries does not create a subjective right to compensation; however, the Executive branch must provide justification for why the revision was not proposed (STF \textit{Tema} 19\footnote{Topic 19 \url{https://portal.stf.jus.br/jurisprudenciaRepercussao/tema.asp?num=19}})~\citep{sobrinho2021direito}.  

In conclusion, our analysis shows that the tumultuous history of Precedent~339, marked by various controversies led by public servants (seen in \Cref{fig:citations_339_type}), continues in a similar way through its new avatar, BP~37 (as seen in \Cref{fig:citations_SV37_type}), corresponding to the hypothesis \textit{Resistance by a group of litigants or regional specificity}. 

It has also created, due to the persistence of this resistance, an influx of cases to the higher courts, not only to STF but also to STJ (\textit{Superior Tribunal de Justiça}), corresponding to the hypothesis \textit{New avenue to the court}. As the controversy around this issue ensues, new cases continue to be brought before the STF, as new attempts to adjust the salary of public servants occur in the absence of formal laws. The contradiction of this never-ending administrative practice to the content of the BP, in itself, creates a clear avenue for the issue to be brought up time and time again before the court. In this case, however, it is interesting to note that this new avenue is not a consequence of lawyers trying to bend formal restrictions or take advantage of indirect legal connections to gain access to the court. Rather, this avenue is a result of direct confrontation between persevering administrative practices and what has been determined as the correct legal interpretation of the Law by the STF.

\subsection{Final legal considerations}\label{subsec:legal_conclusions}

\subsubsection{Empirical insights into the legal debate on binding precedents}
As discussed in \Cref{subsec:legal_literature}, a BP is understood as a legal tool for standardizing judicial decisions \citep{book-tavares-jurisconst}. The previously mentioned requirements for the creation of BPs, as set forth in Paragraph 1 of Article 103-A of the Federal Constitution, reinforce the understanding that this instrument aims to unify jurisprudence, addressing recurring controversies that cause uncertainty or legal insecurity, thus reducing the volume of litigation.
Indeed, it is expected that the BP, by granting binding effects to the consolidated understanding of the court on controversial issues, will produce desirable effects for the legal system \citep{book-mancuso-divergencia}. Among such effects, the following can be highlighted: (i) the reduction of litigation arising from doubts about the correct application of the law, thereby accelerating the resolution of conflicts and increasing the efficiency of judicial activity; (ii) clearer guidance for social agents on how to conduct their legal relations, which in turn would discourage behaviors incompatible with the prevailing legal order; and (iii) the promotion of equity and formal equality, by restricting the decisions rendered by lower courts (which, in turn, reduces the likelihood of similar legal problems receiving divergent judicial solutions) \citep{book-tavares-paradigmas}.

Moreover, the BP can also enhance the effectiveness of constitutional norms and strengthen the role of the STF as the court responsible for constitutional interpretation \citep{book-tavares-justconst}. In this regard, the instrument represents a natural course of evolution in Brazil, given the increasing importance acquired by Constitutional Law since 1988, both for the interpretation and application of infraconstitutional norms across various branches of law and for the realization of fundamental rights provided for in the Federal Constitution itself \citep{book-agra-legitimidade,book-tavares-justconst}.

Much of the academic and doctrinal discussion that accompanied the introduction of the BP in Brazil has focused on understanding how this instrument affects the role of the STF --- and the judiciary more broadly --- in the country \citep{book-streck-sumulas,book-dallari-poder,article-leal-futuro,book-agra-legitimidade}. There was debate over whether the BP, a legal instrument typically associated with the common law system, would be compatible with Brazilian law, which is rooted in a Civil Law tradition. Additionally, questions were raised about the extent to which the BP differs from the legislative function and how the introduction of such a prerogative for the STF would impact the Brazilian separation of powers \citep{inbook-rothenburg-dialetica,book-cunha-efeito,book-cappelletti-legisladores}. It should be noted that these are essentially normative issues, aimed at assessing whether the BP is a desirable instrument and determining what its contours, implementation limits, and main characteristics should be.

However, the main justifications for the introduction of the BP in the Brazilian constitutional system are predicated on empirical claims. After all, the instrument results from an effort to standardize constitutional jurisprudence and contain the demands that reach the STF, given the high volume of litigation the court must handle every year. This effort, in fact, not only led to the introduction of the BP but also to other instruments aimed at reducing the number of cases judged by the STF. Despite this fact, few empirical studies seek to verify whether the BPs produce the expected effects --- particularly studies that aim to ascertain whether there is indeed a standardization of jurisprudence and a reduction in litigation concerning the BP's subject. The reasons for this are partly methodological, as discussed earlier in this text.

Thus, this research, by proposing a methodology that allows measuring the volume of litigation pertaining to the subject of a BP from before its creation, aimed only to confirm the results typically predicted by legal doctrine --- that is, to empirically demonstrate that the introduction of a binding precedent indeed reduces the number of cases judged by the STF on the subject in question. However, the research results revealed a counter-intuitive scenario that has not received appropriate attention in the legal field. The investigated cases showed an increase in the number of decisions addressing the theme or issue covered by the BP after its introduction.
We proposed some hypotheses to explain this result, whose legal significance points to interesting issues in constitutional theory that still deserve to be the subject of applied studies in the future.

The first and most frequent explanatory hypothesis for the increase in litigation following the creation of a BP is what we refer to as the creation of a \textit{New avenue to STF or procedural issues}. This hypothesis is relevant for explaining almost all the cases studied\footnote{It is less present only in BP~17, but was explicitly identified as a relevant cause of litigation in BPs 11, 14, 26, and 37.}. This arises from the fact that the judicial system involves actors who act strategically. It is often a mistake to assume, when seeking a social outcome through a legal change, that the recipients of that change will continue to behave the same way they did before its introduction. Generally, it is necessary to understand how a change in the law will alter the incentives of the actors involved. The claim that a judicial decision is not in accordance with a BP allows the case to reach the STF, creating an interest among litigants to address this type of issue precisely to gain access to the court. In other words, by establishing a BP on a particular subject, the STF increases the constitutional relevance of the issue, signaling to litigants that it will hear cases involving violations of that BP.

Therefore, the establishment of a BP always creates incentives for parties to invoke the issue of the BP to access the court, even if their cases' connection to the subject of the BP is weak or indirect. The same process, in fact, can be observed with the creation of legal rules through the usual legislative process: it is a well-known legal phenomenon that the multiplication of normative acts establishing specific norms tends to increase the complexity of the existing legal system, generating more legal uncertainty and litigation, instead of providing clarity to the law applicable to each specific case. The language of the BP itself can introduce new complexities if the way the BP addresses its subject contains ambiguities or vague terms that give rise to new judicial discussions.

In our analysis, we found that the hypothesis we called \textit{Vague wording} was clearly related only to BP~26, but this case highlighted some peculiarities regarding the problem of vagueness in BPs. As previously noted, the use of vague terms plays an important role in the production of legal norms, allowing different social situations to be addressed by the norm, even when the author of the normative text does not know the specifics of each particular case to which the norm applies. However, in the case of the STF, there is a rather peculiar situation: the body that creates the vague normative text will also be the one that has the final word on how the text should be interpreted and applied. The contemporary state has other bodies with normative and judicial functions, but the decisions issued by these bodies can always be reviewed judicially. In the case of BPs, it will always be up to the STF, ultimately, to resolve controversies regarding its own binding precedents.

The semantic openness of normative statements is also what allows groups of litigants to resist the application of a BP or for divergent interpretations to arise in specific lower courts, creating regional specificities in how the BP is addressed. As a rule, these points of resistance need to mobilize some ambiguity or legal gap left by the BP. The hypothesis we called \textit{Resistance by a group of litigants or regional specificity} proved relevant in two cases: BP~26 and BP~37. In these cases, we see that the creation of a BP will not always have the effect of pacifying social relations and reducing conflicts; it can also incite more litigation and opposition from those affected, especially when the content of the BP does not seem to adequately address a particular case and the agents seek to differentiate it from the original scope of application of the BP (the well-known \textit{distinguishing} strategy).

External circumstances to the judicial dynamics can also create problems for the application of the BP. An unpredictable social or economic event that stimulates new conflicts or affects the conditions for the application of the BP --- or even changes to the constitutional text itself --- can lead to new peaks of litigation on the BP's subject. Two cases were also identified where the hypothesis we called \textit{External conjecture not directly related to the precedent} proved relevant. These are BP~14 and BP~17. In these cases, we observe how the influx of litigation can be related to factors beyond the control of the STF. The BP is intended to govern a complex, dynamic, and somewhat unpredictable socio-economic reality \citep{book-tavares-jurisconst}. When deciding to create a new BP, the STF relies on the cases that have reached the court up to that moment, which are also the result of changing social dynamics. Therefore, the court's perspective on BP issues is subject to biases.

Finally, there was a case (BP~11) where the issue had not been recurrently debated before the introduction of the BP, a hypothesis we call \textit{New theme}. In this case, the STF determined that a practice, which had not been questioned before the court, constituted a significant constitutional violation. The BP was used, therefore, as a way to intervene in social reality to ensure compliance with the Constitution. This is a particularly relevant use of the BP because it relates to a broader debate about the increasing role of the judiciary. The methodology developed by this research allows for measuring the cases in which the use of the BP does not directly result from recurrent societal demands, but rather from a more assertive action by the court to ensure the effectiveness of the Constitution.

All five hypotheses listed in this study suggest that the establishment of a BP can, in certain cases --- or possibly in most cases --- result in an increase in the number of litigations, raising new controversies and legal issues that demand judicial solutions. This problem has already been discussed theoretically by legal doctrine, but it has not yet been measured, despite being based on claims that are, as stated, essentially empirical. We hope that the methodology proposed here will provide important contributions to a better understanding of the impacts of BPs and the possible causes of increased litigation after the issuance of a BP.

\subsubsection{Limitations and potential improvements}

It is important to note that our entire analysis, which claims to uncover the mechanisms underlying the repetitive use of a binding precedent, is significantly biased by the fact that we perform Case Classification only on the documents brought up to trial at the Supreme Court.
In particular, our method is vulnerable to a survivorship bias, unaware of themes that are confined to the lower courts, or not taken to court at all (e.g., theses from advocacy groups, reports from various agencies, political events).

This problem can be mitigated in at least three ways. The first, already mentioned above, would be to acquire data from courts below the STF.
In the case of STJ (Superior Court of Justice), data is available on the official website\footnote{STJ search engine \url{https://www.stj.jus.br/sites/portalp/Processos/Consulta-Processual}}.
This way, the authors of \cite{Nunes2022} extracted 129,602 cases, enabling an analysis of corporate law precedents.
For other courts, however, a search engine is not available, and we are not aware of any public database --- the data acquisition must be done by asking each court.

Another important avenue would be a field survey, involving interviews with legal professionals (e.g., lawyers, public defenders, prosecutors, or judges). 
In this regard, we had the opportunity to speak with members of the Public Defender’s Office and Public Prosecutor's Office of the State of São Paulo, who kindly presented to us, from their point of view, the important issues at stake in the creation of the BPs (more specifically BP~26).
However, the development of a sociological analysis, which would take into account the various political and legal positions with regard to Supreme Court precedents, would require the collection of many other points of view.

Finally, our analysis would benefit from the incorporation of expert annotations in the training set, in order to isolate key elements of the documents. Indeed, our models are trained in such a way that all parts of the text are considered equally; this potentially introduces an important bias, by encouraging the integration of elements that are too specific to certain cases, or by giving equal importance to cases judged solely on procedural grounds. 
Such an approach is expected to refine our models.
In this regard, and given the considerable number of documents studied, it would be interesting to produce automatic annotations using Large Language Models.

\section{Conclusion}\label{sec:conclusion}

The Brazilian Federal Supreme Court routinely deals with an overwhelming number of cases with repetitive demands. 
While there is a juridical instrument specially created to reduce the number of repeated demands through the establishment of a normative jurisprudence --- the binding precedent ---, this instrument often fails in this regard. 

In this article, we have performed a series of mathematical and juridical analyses that allowed us to explain why BPs 11, 14, 17, 26, and 37 (some of the most cited in the STF’s decisions) fail in reducing the Court's workload. First, we have applied and compared different Case Classification methods (TF-IDF-based models, LSTM, Longformer, and regular expressions) on a database composed of STF’s decisions. By studying time series of similar cases, our methodology enabled us to assess the impact of laws on jurisprudence and to perform an empirical study of the juridical mechanisms behind these BPs' inefficiency. We finally described five main hypotheses that explain the large number of cases reaching STF.

From the mathematical point of view, the TF-IDF models used in our analyses performed slightly better than LSTM and Longformer, when compared through the $F_1$ score calculated from the labeled dataset (Dataset~\#1).
Nevertheless, this is counterbalanced by the fact that, in the larger dataset (Dataset~\#2), the deep learning models were able to detect certain important legal events that TF-IDF missed.
To enable a more refined analysis of the predictions, it would be interesting to consider models that not only detect cases falling under the BP's field of application but also the role it could play in the case at hand. 
As we have seen throughout the article, binding precedents can be used for a wide variety of purposes, whether for supporting or countering an argument, providing context, or introducing a legal doctrine, but they are also used to judge cases based on their merits or procedural grounds, an important piece of information for characterizing the applications of the BP.
In addition, we intend to investigate whether the incorporation of expert-annotated key parts would lead to better results or new insights.

From a legal perspective, our study uncovers a counter-intuitive effect that has been overlooked in purely doctrinal works: the very nature of the binding precedent brings with it reasons for increasing repetitive demands.
This includes the fact that, when a BP is published, it is declared to be a new topic of relevance that the Supreme Court is committed to dealing with; that it can lead to the formation of diverse groups of litigants; that it introduces new complexities and potentially ambiguities; that it exposes and legitimizes a new discussion topic; and that it is naturally subject, in an indirect way, to other economic, political or social conjectures.
To refine our study, which is based solely on the analysis of STF decisions, we plan to integrate somehow lower court decisions, as well as conduct a field survey to gather the views of various legal professionals.

\bibliography{main}
\end{document}